\documentclass{article}

\usepackage{arxiv}

\usepackage[utf8]{inputenc} 
\usepackage[T1]{fontenc}    
\usepackage{hyperref}       
\usepackage{url}            
\usepackage{booktabs}       
\usepackage{amsfonts}       
\usepackage{nicefrac}       
\usepackage{microtype}      
\usepackage{lipsum}
\usepackage{subcaption}
\usepackage{amsmath}
\usepackage{graphicx}
\usepackage{booktabs}
\usepackage{caption}
\usepackage{subcaption}
\usepackage{float}
\usepackage{xcolor}
\usepackage{tikz}
\usepackage{hyperref}
\usepackage{cleveref}
\usepackage{siunitx}
\usepackage{enumitem}
\usepackage{microtype}
\usepackage{listings}
\usepackage{multirow}

\newcommand{\promptcaption}[1]{%
  \par\smallskip\noindent\textit{#1}\par\smallskip
}

\usepackage{graphicx}
\usepackage{subcaption}
\usepackage{stackengine}

\newcommand{\panel}[4][3pt]{%
  \stackinset{l}{#1}{t}{#2}{\textbf{(#3)}}{#4}%
}

\usepackage[numbers]{natbib}

\title{LLM sequential decision making under uncertainty in biochemical domains}

\author{
  Mattias Akke \footnotemark[2] \\
  Lund University \\
  \And
  Soojung Yang \thanks{Co-correspondence: \texttt{soojungy@mit.edu}, \texttt{rafagb@mit.edu}} \\
  Stanford \\
  \And
    Jur\'{g}is Ru\v{z}a\\
  MIT \\
  \And
  Sathya Edamadaka\\
  MIT \\
  \And
  Rafael G\'{o}mez-Bombarelli \footnotemark[1] \\
  MIT \\
}

\begin{document}
\maketitle
\renewcommand{\thefootnote}{\fnsymbol{footnote}}
\footnotetext[2]{Work done at MIT}
\renewcommand{\thefootnote}{\arabic{footnote}}

\begin{abstract} 
Large language models (LLMs) are increasingly used to drive scientific discovery. Understanding how LLMs make decisions from new data and memory of the literature is vital before trusting them to design experiments under tight experimental budgets. However, their decision strategies are invisible in the current performance scores used to evaluate research agents. Here, we benchmark five frontier LLMs in a Bayesian Optimization setting against published statistical baselines on seven combinatorial datasets spanning protein engineering, reaction optimization, molecular design, peptide self-assembly, and catalysis. Performance is paired with direct measurements of model beliefs and actions, enabling highly resolved behavior analysis. A prompt ablation that progressively strips context separates memorization from chemical reasoning and from bare categorical optimization. Prior chemical knowledge helps in expectation, but with high variance and occasionally even harms performance. No configuration tested decisively beats a mean statistical baseline across domains. Belief-movement and Martingale diagnostics, corrected here for a measurement-noise bias that mislabels rational agents as irrational, show that models overreact to incoming data rather than entrenching on their priors in the contexts studied here. Interestingly, while LLM actions are exploitative, models sincerely intend to explore and consistently act on that intent. This failure is a competence gap arising from context-stickiness. Removing in-context history restores exploration, indicating that priors and data must be decoupled to achieve effective LLM-driven discovery.
\end{abstract}

\section{Introduction}
\label{sec:introduction}
A central feature of scientific research is that the search space of experimental conditions is potentially enormous, yet physical experiments are constrained by cost and time. Discovery then unfolds as a series of decisions under partial knowledge, where new information often contradicts prior belief. Large language models (LLMs) are increasingly deployed in this loop to propose, test, and revise hypotheses in response to data --- off-the-shelf, fine-tuned on domain data~\citep{schwaller2024chemistryllm, sun_synllama_2025, chaves_tx-llm_2024, thomas_test-time_2025}, or embedded in agentic frameworks~\citep{sumers_cognitive_2024, fei2026agentsfailautoresearchendtoend} that act through tools to run experiments. Successful examples are reported in protein~\citep{ghafarollahi_sparks_2025} and alloy~\citep{ghafarollahi_automating_2025} design, studies of molecular degradation ~\citep{ghareeb_robin_2025}, and more~\cite{rankovic2025largelanguagemodelsuncertaintycalibrated}. Off-the-shelf models already encode priors over the search space that exceed those of models fit only to a task-specific training distribution. However, it remains unclear how those priors update against incoming evidence, and therefore whether LLMs can be trusted to spend real experimental budget in discovery.   

\begin{figure*}[t]
  \centering
  \begin{subfigure}[t]{0pt}\phantomsubcaption\label{fig:setup-loopdata}\end{subfigure}%
  \begin{subfigure}[t]{0pt}\phantomsubcaption\label{fig:setup-loop}\end{subfigure}%
  \begin{subfigure}[t]{0pt}\phantomsubcaption\label{fig:setup-configs}\end{subfigure}%
  \begin{subfigure}[t]{0pt}\phantomsubcaption\label{fig:setup-prompts}\end{subfigure}%
  \begin{subfigure}[t]{0pt}\phantomsubcaption\label{fig:setup-scoring}\end{subfigure}%
  \begin{subfigure}[t]{0pt}\phantomsubcaption\label{fig:setup-behaviour}\end{subfigure}%
  \begin{tikzpicture}
    \node[anchor=south west,inner sep=0] (fig)
      {\includegraphics[width=\textwidth]{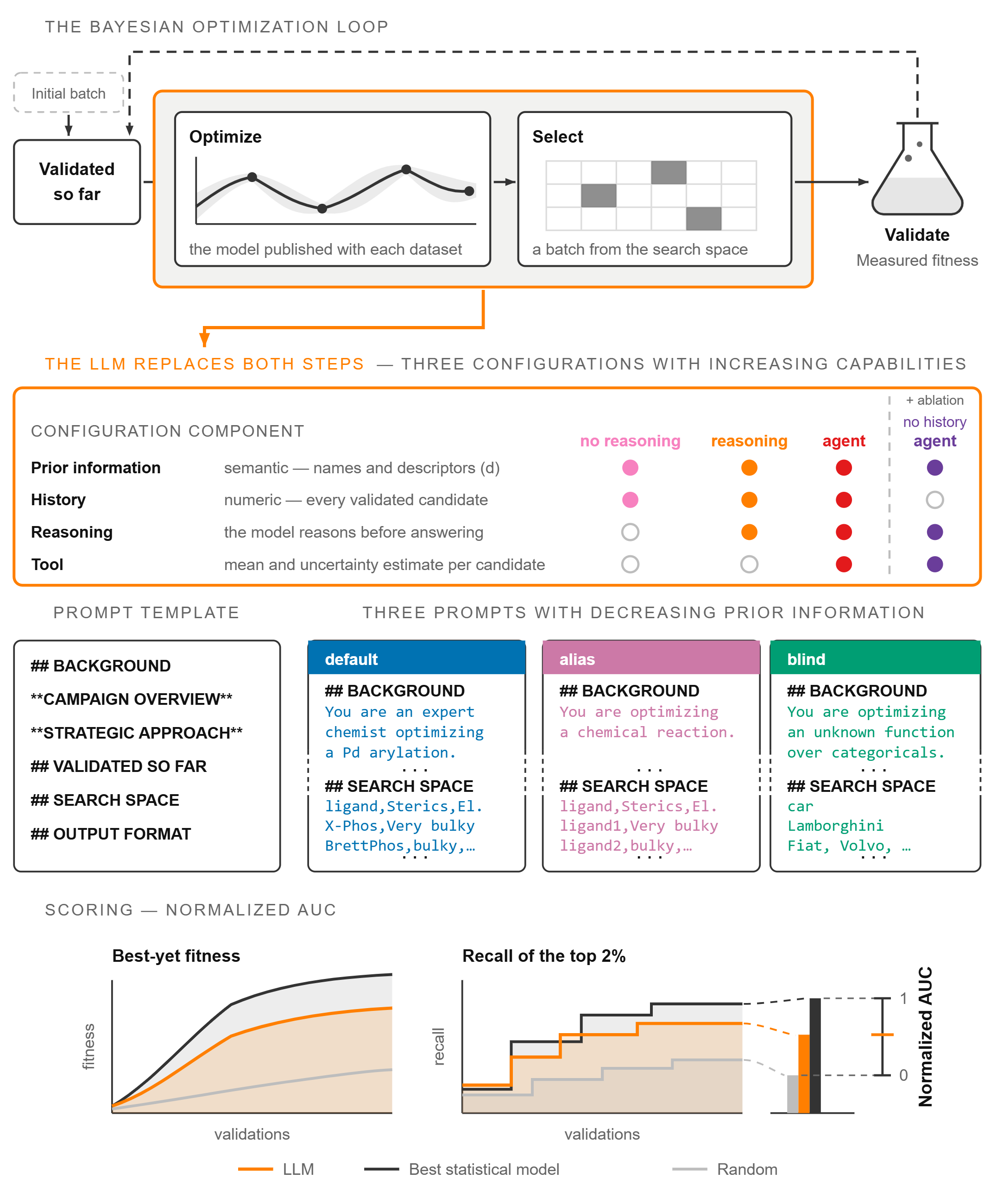}};
    \begin{scope}[x={(fig.south east)},y={(fig.north west)},
                  every node/.style={anchor=base west,inner sep=0,font=\sffamily}]
      \node at (0.008,{0.9654-0.0001}) {(\subref{fig:setup-loopdata})};
      \node at (0.008,{0.6900-0.0001}) {(\subref{fig:setup-loop})};
      \node at (0.008,{0.4850-0.0001}) {(\subref{fig:setup-configs})};
      \node at (0.008,{0.2040-0.0001}) {(\subref{fig:setup-prompts})};
    \end{scope}
  \end{tikzpicture}
  \caption{Overview of each experimental setup. 
  \textbf{(a)} The standard Bayesian optimization loop. An initial batch of validated data trains a model to predict fitness over the search space of candidates. An acquisition function then selects a batch of next experiments. Experiments (here simulated using a lookup table) validates selected candidates. \textbf{(b)} The LLM replaces the optimization and selection steps. Three different LLM configurations with increasing capabilities are tested. The \textbf{no-reasoning} prompt includes prior information about the task and a history of validated candidates each turn. It provides the next batch of candidates directly. The \textbf{reasoning} models are asked to strategize with reasoning turned on. The \textbf{agent} has a tool wrapping the statistical model in (a), offering an independent mean and uncertainty estimate about the candidates. Additionally, \textbf{no history agent} probes the behavioral impact of the in-context history (Section~\ref{sec:results-in-context}). \textbf{(c)} All experiments use a single prompt template. Three different prompt modes with decreasing prior information about the task at hand separate memorization, domain knowledge, and optimization ability. \textbf{(d)} Performance is scored using normalized AUC by two different metrics to enable comparisons between datasets. 
  }
  \label{fig:setup}
\end{figure*}
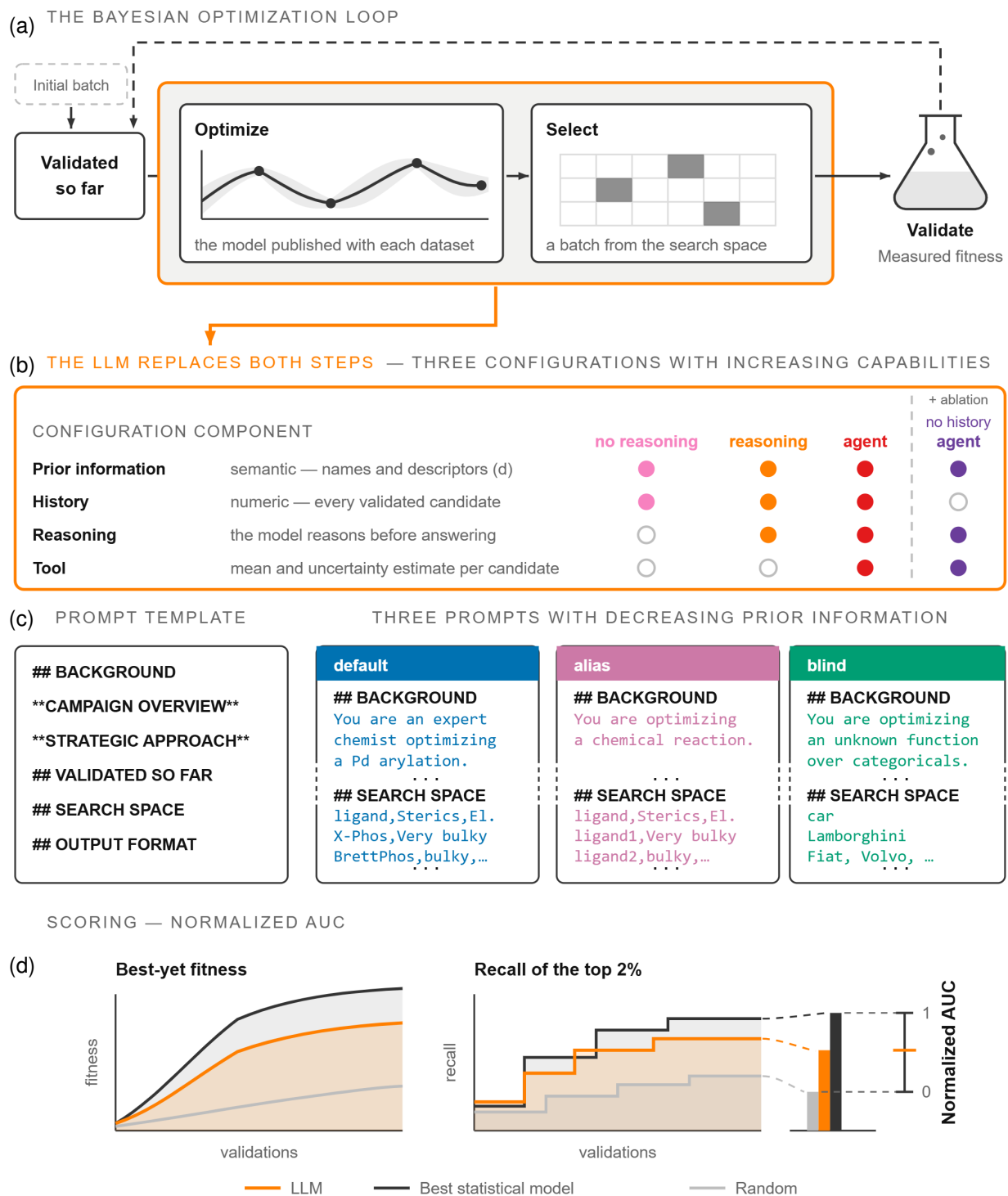

This discovery loop can be formalized as Bayesian Optimization (BO), where a prior over the search space is updated as data accumulate and each new query is drawn from the resulting posterior (Figure~\ref{fig:setup-loop}). BO has become a standard tool for biochemical and materials discovery, valued for sample efficiency in costly evaluation settings where the acquisition function makes the exploration-exploitation tradeoff explicit~\citep{mcdonald_bayesian_2025, dang_preferential_2025, dave_autonomous_2022, jenewein_navigating_2024, tran_active_2018, ghorbani_active_2024, khan_toward_2023, stanton_accelerating_2022}. To leverage both statistical efficiency and the rich prior knowledge of LLMs, a growing body of work casts language models as BO optimizers~\citep{yang_large_2024}. The literature reports mixed signals. In the biochemical and materials domains, off-the-shelf models have been reported to match or even beat statistical baselines~\citep{liu_large_2024,lu_generative_2025}. Reinhart \textit{et al.} (2024)\citep{reinhart_large_2024} have shown that Claude 3.5 \citep{noauthor_introducing_nodate} outperforms active-learning and genetic-algorithm pipelines on macromolecule design, suggesting LLMs implicitly balance exploration and exploitation. However, others report poor performance of language models as optimizers~\citep{wang_molecular_2024} unless fine-tuned on domain data~\citep{kristiadi_sober_2024}. Gupta \textit{et al.} (2025)~\citep{gupta_llms_2025} show that statistical models outperform open-source LLMs across genetic perturbation and molecular property optimization tasks, and even report that language models are insensitive to experimental feedback, performing no worse when true outcomes are replaced with randomly permuted labels. Our own earlier work landed in between, with LLMs overfixating on irrelevant context such that withholding information actually improved performance~\citep{akke2025bayesian}. Every side of this debate argues over a performance number, but none inspects the decision policy that produces it.  

Tools for understanding the decision-making and behavior of language model exist, but have not been applied to scientific discovery and LLM-driven BO. 
Belief-movement and Martingale diagnostics from cognition and economics respectively~\citep{AugenblickRabin2021BeliefMovement, Augenblick2023more, gershman2018humanAlgorithms} formalize how far a rational agent's beliefs should move as new evidence arrives, and have recently been applied to LLMs~\citep{falck2024incontextlearninglargelanguage}. 
These frameworks diagnose two opposing pathologies in two different settings. In semantically rich domains, where the model already has prior knowledge about the real-world meaning of available options, such as in forecasting real events or reviewing academic papers, He et al. (2025)~\citep{he_martingale_2025} find that, across iterative reasoning, LLMs entrench; their beliefs drift back toward the prior rather than toward the evidence. 
In abstract, semantically inert domains, where the agent has no useful prior and must rely on the incoming signal alone, the opposite is observed. In controlled card-drawing inference tasks with no domain content, humans move their beliefs too much in response to weak signals while underreacting to strong ones~\citep{Augenblick2023more}, and LLMs reproduce this pattern, overreacting when the signal is weak~\citep{bini_behavioral_2026}. 
Critically, entrenchment has been observed only where priors are rich and data absent, and overreaction only where priors are absent and data weak. 
Scientific discovery is both at once --- semantically rich, with only a few noisy points in a vast search space. Here, the two regimes make opposing predictions, and no existing work resolves which one governs LLM behavior in the scientific domains such as biochemical and materials discovery.    

A scientist deciding whether to hand an LLM an experimental budget needs to know whether the agent is acting on the data in front of it or on its memory of the literature. As we show, that distinction is invisible in performance scores alone. We resolve it by studying belief evolution and behavior directly, varying the semantic prior available to the agent while holding the data flow constant in a sequential decision-making campaign. Specifically, we control how much information the model receives through three prompting modes that progressively strip away the chemical context. At each cycle, the model chooses which regions of the combinatorial space to retrieve experimental data in the next batch, and we compare its behavior against a Bayesian reference. BO is useful as a behavioral reference because its surrogate model can provide an explicit posterior over the search space (i.e., a prediction and uncertainty estimate) to score LLM implicit beliefs and actions against. Its acquisition function is a separate, swappable policy with interpretable explore–exploit balancing, letting us separate what an agent believes from how it acts (Figure~\ref{fig:setup-behaviour}). We treat the Bayesian reference as a diagnostic, not a claim of optimality. We detail how we use BO as a behavioral reference and practical baseline, and its limits, in the Methods section.

We find that prior knowledge gives LLMs a real but high-variance edge that is occasionally even harmful to performance, and no model beats the mean statistical baseline on average. 
A common failure mode is that LLMs are too exploitative rather than exploring properly, and we show that what manifests as an exploitative policy is not a preference but a competence gap. LLMs sincerely intend to explore and act on that intent, yet realize little information gain because their selections avoid the high-uncertainty regions of the space.
An explicit uncertainty estimate, supplied as a tool, only partly repairs this. 
The underlying mechanism, common to every model tested, is context-stickiness --- the models ground their decisions in the data already in the context window and cannot reason beyond it --- and in this rich-prior, weak-signal regime, it manifests as overreaction to incoming data rather than entrenchment on priors.  

This paper makes three types of contributions. 
Methodologically, belief- and uncertainty-dynamics analysis, extended to combinatorial search spaces with marginalized belief updates, provides much finer resolution than prior LLM-BO work and uncover new insights into LLM decision making.
In addition, a noise correction removes a bias in widely used Martingale-based belief diagnostics that otherwise flags even a perfectly rational agent as overreacting when its beliefs are measured with noise. The corrected metric shows LLMs overreact to new data.
Empirically, a prior-ablation design --- default, alias, and blind prompting modes that progressively strip chemical information from the prompt --- identifies and separates failures related to the ability of an LLM to memorize chemical knowledge, reason over chemical descriptors, and optimize over bare categorical spaces. 
Scientifically, the failure modes of LLM discovery agents involve overreacting to weak signals rather than entrenching on priors, and their under-exploration is a competence gap rather than a preference. These findings yield concrete guidance for interpreting and designing LLM-based BO pipelines for scientific discovery. 
\clearpage
\section{Results}
\label{sec:results}

We benchmark five frontier LLMs against published statistical models across 7 datasets spanning 5 biochemical and materials domains (protein optimization, chemical reaction optimization, reaction synthesis, peptide self-assembly, and metal catalyst optimization; Figure~\ref{fig:setup-loopdata}). Each LLM receives a prompt containing prior background on the task, the optimization campaign, and a history of all data points tested in earlier steps, then is asked to sequentially propose batches of experiments (Figure~\ref{fig:setup-configs}). Performance is the area under the best-yet-fitness and top-2\%-recall trajectories, normalized per dataset so that random selection scores 0 and the best statistical model scores 1 (Figure~\ref{fig:setup-scoring}). To separate dataset memorization, general chemical knowledge, and optimization strategy, we vary how much prior information the prompt carries (Figure~\ref{fig:setup-prompts}): \textbf{default} (full biochemical names and descriptors), \textbf{alias} (descriptors retained, names replaced with generic labels), and \textbf{blind} (a domain-agnostic one-hot search space). The series progressively strips prior knowledge without changing the optimization function, letting us ask whether an LLM prior is entangled with its optimization strategy or merely a layer on top. The LLMs are run in three configurations (Figure~\ref{fig:setup-configs}): with reasoning off (\textbf{No reasoning}) and on (\textbf{Reasoning}), and as an agent equipped with a tool that wraps the same exact statistical models the LLM is benchmarked against, providing an independent uncertainty estimate beside the in-context history and prior information (\textbf{Agent}). 

\subsection{Prior knowledge helps on average but is unreliable}
\label{sec:results-benchmark}

\begin{figure}[tp]
  \centering
  \begin{subfigure}[b]{\linewidth}
    \centering
    \panel{3pt}{a}{\includegraphics[width=\linewidth]{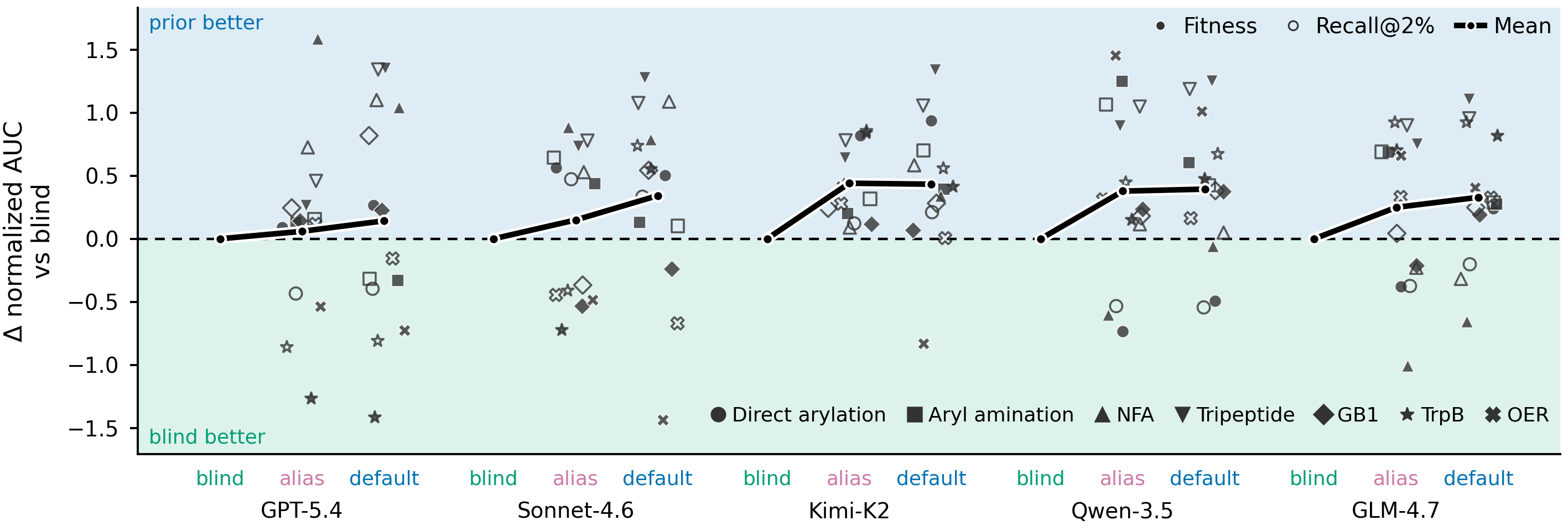}}
    \phantomsubcaption
    \label{fig:benchmark-prior}
  \end{subfigure}\\[0pt]
  \vspace{0.0226\linewidth}
  \begin{subfigure}[b]{0.4202\linewidth}
    \centering
    \panel{3pt}{b}{\includegraphics[width=\linewidth]{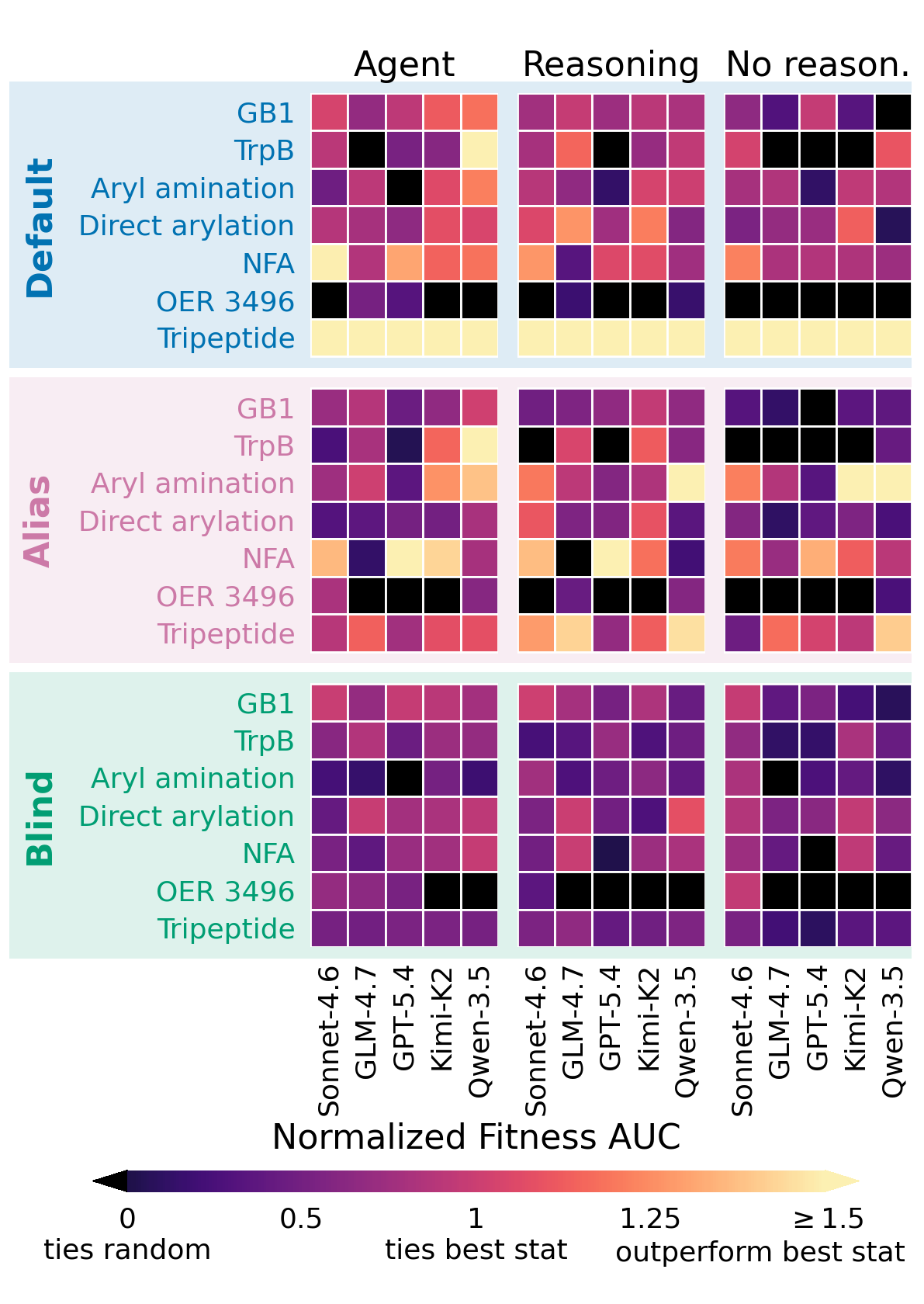}}
    \phantomsubcaption
    \label{fig:benchmark-heatmap}
  \end{subfigure}%
  \begin{subfigure}[b]{0.5798\linewidth}
    \centering
    \panel{3pt}{c}{\includegraphics[width=\linewidth]{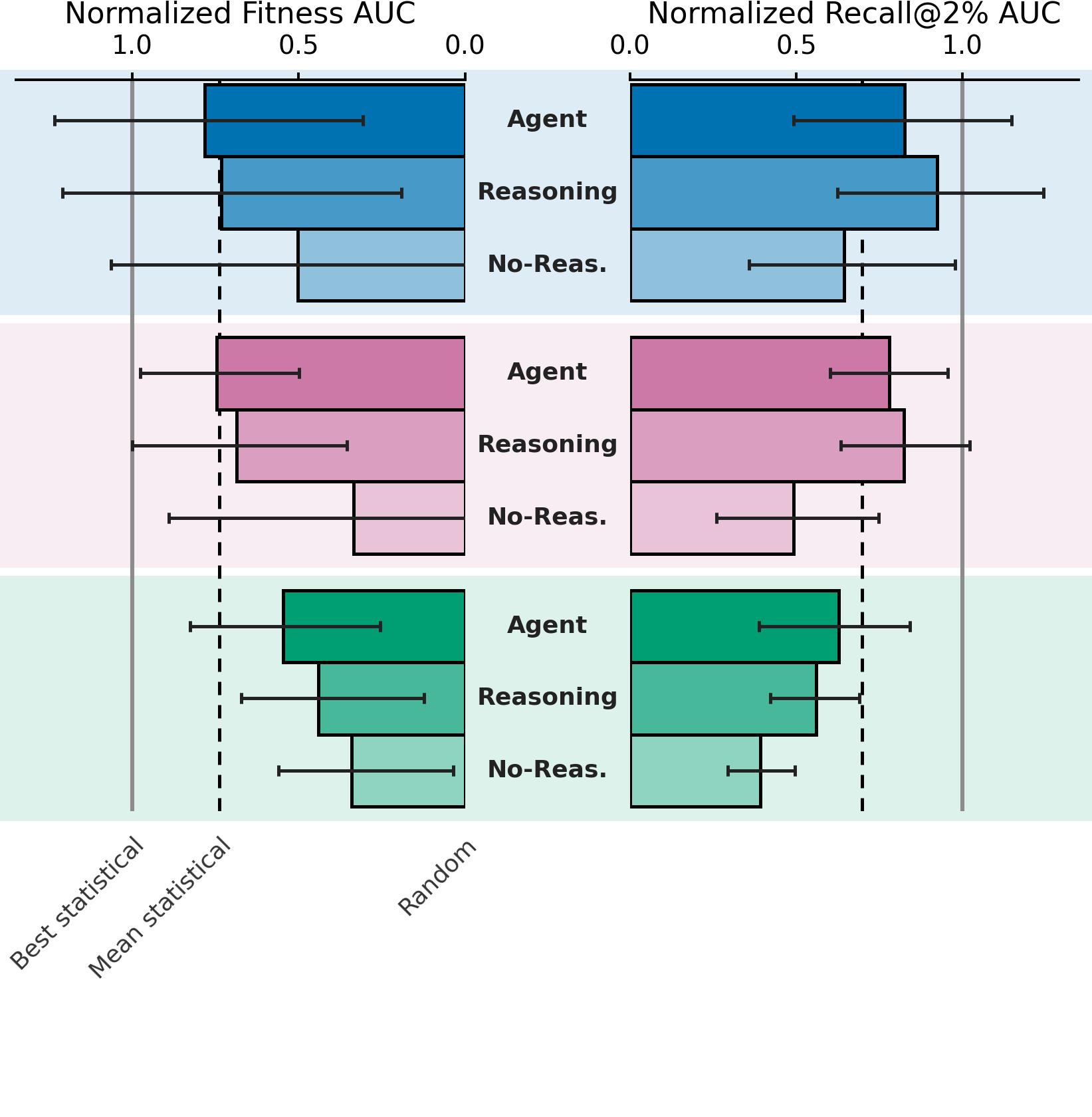}}
    \phantomsubcaption
    \label{fig:benchmark-overall}
  \end{subfigure}
  \caption{Prior knowledge helps on average but is unreliable. \textbf{(a)} Change in normalized AUC relative to blind mode, one point per dataset and metric, ordered by increasing prior information in the prompt (blind, alias, default). Filled points mark fitness; open points mark recall@2\%. Black line marks the mean per model. \textbf{(b)} Normalized fitness AUC per dataset, model, and capability (agent, reasoning, no reasoning), in each prompt mode. 1 (yellow) ties the best statistical model, 0 ties random, black cells underperform random. \textbf{(c)} Normalized AUC pooled over datasets and models, fitness left and recall right. No configuration reaches the best statistical model in expectation; reasoning and agents perform within each other's confidence intervals. Error bars are 95\% intervals from a nested bootstrap (datasets, then models, the seeds). The 7 datasets set the effective sample size}
  \label{fig:benchmark}
\end{figure}

The default and alias modes often yield similar performance across datasets and models (Figure \ref{fig:benchmark-prior}), indicating that the LLMs use the chemical descriptors given to them, rather than drawing decisions directly from memory invoked by domain-specific terms. The prior helps on average, and there is a weak positive correlation between increased prior information and performance across the two metrics used (Spearmann-$\rho=+0.20$, one-sided $p=0.03$, hierarchical bootstrap, Figure \ref{fig:benchmark-prior}). The actual value of the prior relative blind is high variance, worth about 1.0 batch (CI$_{95\%} [-0.21, 2.58]$) of data points. Adding a tool to the LLM (Agent) does not reliably improve performance in any prompt mode (Figure \ref{fig:benchmark-overall}). The high variance across models and datasets, as well as differences between fitness and recall scores, highlights the limitations of using absolute performance to elucidate behavior or capability (Figure \ref{fig:benchmark-heatmap}). 

Per dataset, the picture is highly heterogeneous (Figure \ref{fig:benchmark-heatmap}). LLMs beat the best statistical model in default mode on non-fullerene-acceptor (NFA) ($p_{BH}<0.05$) and tripeptide self-assembly ($p_{BH} <0.001$, correction across 7 respective datasets and 2 metrics). At the same time, they fall short elsewhere, occasionally harming performance relative to blind mode (Figure~\ref{fig:benchmark-prior}). Models also vary substantially, but no one model is consistently better. For full per-domain and per-model breakdowns, and convergence analysis, we refer to~\ref{si:per-dataset}. There, we also document three distinct failure modes that highlight peculiar dataset-specific biases. Briefly, we identify that (i) GPT-5.4 and Qwen3.5 consistently avoid the optimal ligand \texttt{CgMe-PPh} with and without reasoning or tools ($p_{BH} < 0.001$, 750 selections total); (ii) the name of a protein,``TrpB'' is associated with over-selection of aromatics in GPT-5.4 ($p_{BH}<0.001$, 5 models), although aromatics are worse on average; and (iii) that LLMs prefer to start with simple experiments. It is not trivial to determine a policy for where which model will work well (Section~\ref{si:benchmark}). 

\subsection{Belief dynamics rule out entrenchment towards priors}
\label{sec:belief-dynamics}
Overall performance is a poor surrogate for understanding LLM strategies in these settings as it is opaque to which strategies are used, and sensitive to test settings such as batch size, campaign length, and metric of choice. The weak performance gained from adding prior information could be explained by either entrenchment to the prior--models overfixate on semantically rich information and ignore data--or overreaction--models overfixate on the incoming data stream and ignore the prior. We therefore estimate the beliefs reasoning LLMs hold during optimization directly in two ways. An LLM judge reads each reasoning trace and assigns, per dimension in the combinatorial search space, a marginal probability that a given label belongs to the best candidate (stated belief). Separately, we derive a belief directly from the actions (action belief), as a smoothed marginal over the proposed batch. Neither belief can resolve cross-dimensional relationships such as "temperature 100 is good given ligand X". Marginalization also hinders analysis on the OER dataset, whose search space is a reduced simplex where beliefs cannot be easily assigned to each component.

\begin{figure}[tp]
  \centering
  \begin{subfigure}[b]{0.3598\linewidth}
    \centering
    \panel{6pt}{a}{\includegraphics[width=\linewidth]{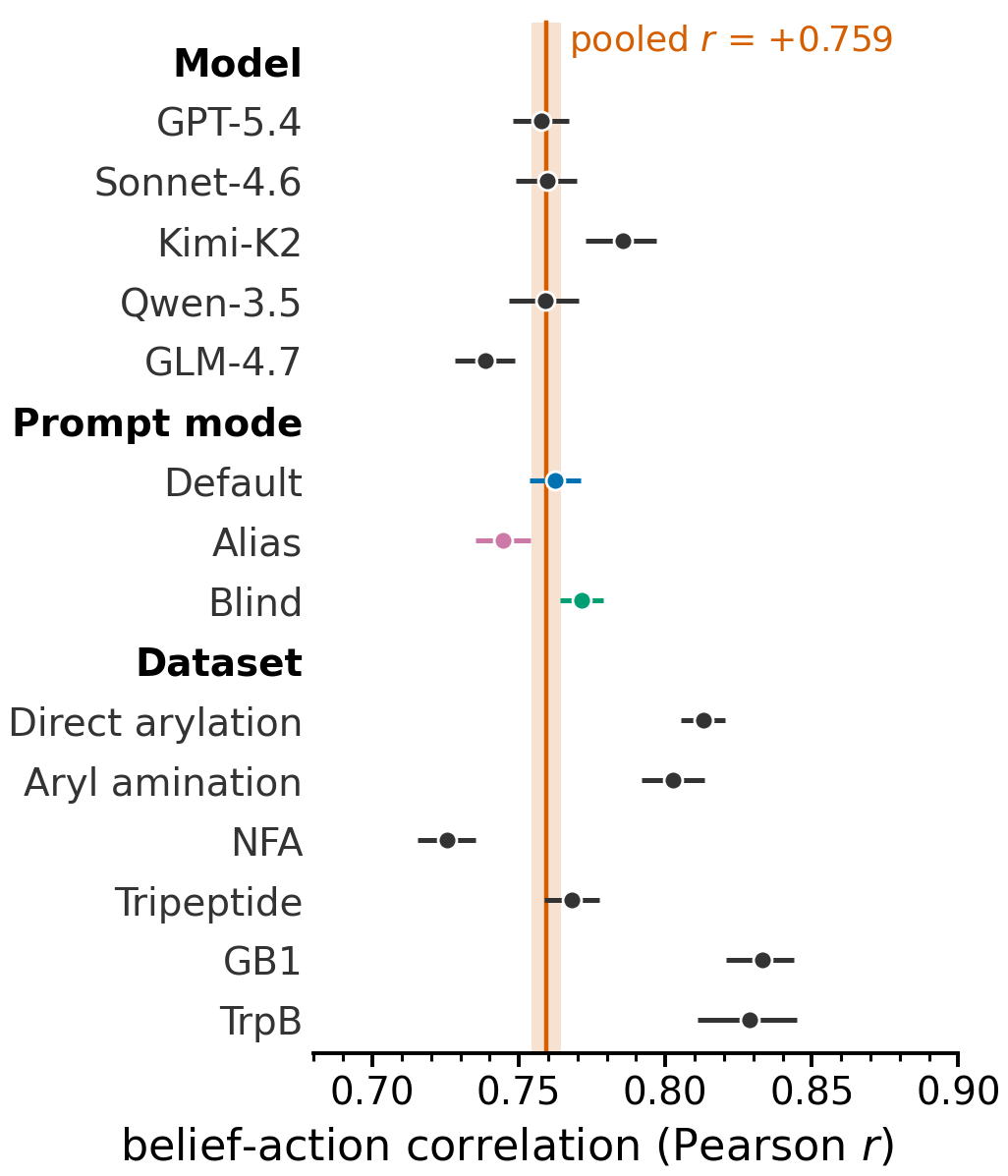}}
    \phantomsubcaption
    \label{fig:beliefs-correlation}
  \end{subfigure}%
  \begin{subfigure}[b]{0.6401\linewidth}
    \centering
    \panel[15pt]{0pt}{b}{\includegraphics[width=\linewidth]{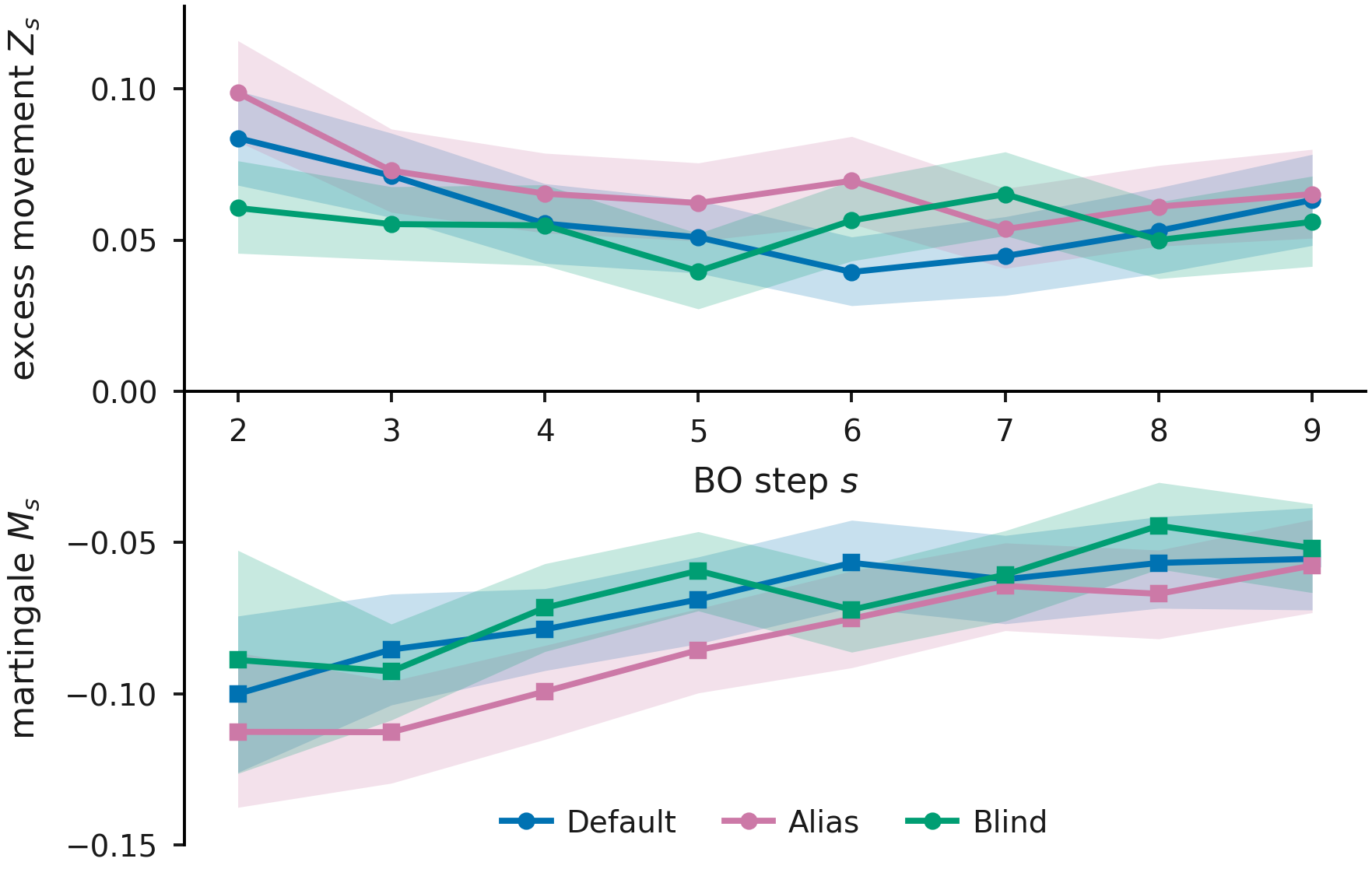}}
    \phantomsubcaption
    \label{fig:beliefs-martingale}
  \end{subfigure}\\[0pt]
  \begin{subfigure}[b]{\linewidth}
    \centering
    \panel{3pt}{c}{\includegraphics[width=\linewidth]{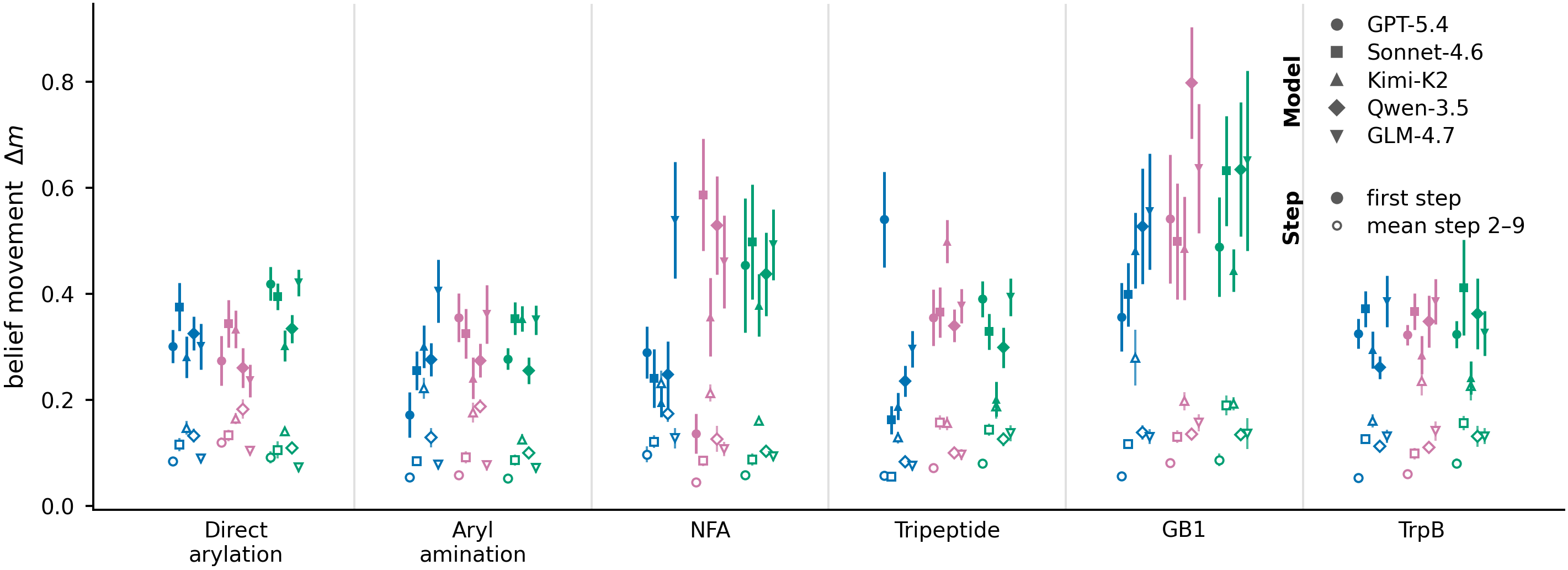}}
    \phantomsubcaption
    \label{fig:beliefs-firststep}
  \end{subfigure}
  \caption{Belief dynamics rule out entrenchment towards priors.\textbf{(a)} Agreement between stated beliefs, read from reasoning traces by an LLM judge, and action beliefs derived from the proposed batches. The pooled Pearson $\rho=0.759$ (CI$_{95\%}[+0.754,+0.764]$, 673{,}702 beliefs from 13{,}484 reasoning traces over 1{,}350 campaigns) holds across every model, prompt mode, and dataset. \textbf{(b)} Excess movement $Z_s$ and Martingale score $M_s$ by BO step, pooled by prompt mode, computed from the stated beliefs. A rational Bayesian agent would sit at 0 for both martingale score and excess movement. Every mode is positive in $Z_s$ and negative in $M_s$ at every step: the models over-react to incoming data rather than entrenching around their prior. Shading marks 95\% confidence intervals. \textbf{(c)} Per-dataset belief movement $\Delta m$, at the first step (filled) compared with the mean of steps 2--9 (open). Beliefs swing roughly $3\times$ more on the first update as the model moves away from a purely prior-driven belief, then settle into more calibrated updates thereafter. Error bars are 95\% confidence intervals across seeds.}
  \label{fig:beliefs}
\end{figure}

Across models, prompts, and datasets, the stated beliefs are well correlated with the actual selections (Figure \ref{fig:beliefs-correlation}). Section~\ref{sec:belief-validation} clarifies that this strong agreement between belief estimates is not driven mainly by the judge being able to deduce the final extracted candidates. The judge routinely places substantial probability mass on unselected candidates. 

Beliefs are then used to evaluate how the LLMs respond to new data. Belief updates of a rational agent should be unpredictable to the agent itself and from its prior beliefs. This does not mean that the magnitude should be unpredictable; However, if most of the search space has already been observed, the next update is probably small, which is foreseeable. The direction must be unpredictable. Suppose the agent's current best guess of the quantity it cares about is expected to rise. If it truly believed that, a rational agent would already have revised the guess upward before seeing any new data. So it cannot know in advance whether the next observation will push that guess up or down. 
We measure deviations from this property using two diagnostics from cognition and economics: (1) the excess-movement statistic, $Z = \bar{m} - \bar{r}$, comparing realized belief movement $\Delta m$ to realized uncertainty reduction $r$, and (2) the Martingale score $M$, the slope in the regression of belief updates on prior beliefs. For a rational Bayesian agent, both are zero; positive movement and negative slope indicate over-reaction to new data, while the reverse indicates entrenchment. We correct for an error-in-variables bias in the naive slope estimate from the original Martingale Score paper~\citep{he_martingale_2025}, and a similar bias in the original excess movement paper~\citep{AugenblickRabin2021BeliefMovement} (Methods-\ref{sec:methods-correction}). 

All models, on all datasets, overreact similarly to incoming data in all prompt modes (Figure \ref{fig:beliefs-martingale}). Beliefs move the most in the first step (Figure \ref{fig:beliefs-firststep}) as the model adjusts from being grounded solely in its prior to basing its beliefs on data. Later updates are more calibrated, suggesting that most of the prior belief is overwritten in the first step. Stated and action beliefs agree well, confirming that measured over-reaction is not caused by over-emphatic reasoning. The belief dynamics rule out entrenchment. Rather than ignoring new data, the models overreact to it, drawing more information from the new data stream than a rational agent would.

\subsection{LLMs are context-sticky optimizers}
\label{sec:results-context-sticky-optimizers}

Performance and belief dynamics can explain \emph{where} the LLMs look for evidence. However, it does not answer \emph{how}. We therefore decompose every selection made by an LLM or a statistical model into a value feature (V) and an uncertainty feature (U). We derive the features by fitting a conditional logit model to the actions taken, using a single shared Gaussian-process surrogate (Methods). High V denotes a value-seeking choice; high U, an uncertainty-seeking one. Six statistical acquisition strategies make up the behavioral baselines: upper confidence bound (UCB), expected improvement (EI), Thompson sampling (TS), $\epsilon$-greedy, greedy, and directed evolution (DE, i.e., randomly changing one dimension around the top performer). The decomposition separates the statistical policies as expected, placing DE and greedy acquisition in the uncertainty-agnostic regime, and UCB and EI sampling in the roaming, uncertainty-aware regime (Figure~\ref{fig:mechanism-vu}).

As expected, reasoning and non-reasoning LLMs' actions are not uncertainty-seeking~\citep{englander_agents_2026, krishnamurthy_can_2024, zhang_comparing_2025}. They select more low-uncertainty candidates than any other statistical model tested. Non-reasoning LLMs match or are more uncertainty-avoiding than even directed evolution (Figure~\ref{fig:mechanism-decomp}). The reasoning LLMs achieve even less coverage of the search space than DE (Figure~\ref{fig:mechanism-explore}).

We propose this because LLMs are ``context-sticky" -- clustering selections around candidates already in the context window -- unlike any other samplers tested. They therefore appear to avoid high-uncertainty candidates, which are generally rare or not included in the history of past selected candidates. Indeed, measuring the mean Hamming distance from each proposal to the history confirms that the reasoning LLM overwhelmingly selects candidates close to those already in context (Figure~\ref{fig:mechanism-novelty}), more so than any non-DE model. As we show in Section~\ref{sec:belief-dynamics}, the clustered selections are data-driven rather than an artifact of belief entrenchment, which would also cluster selections. Further, reasoning does not alleviate bias relative to non-reasoning, but enforces it (Figure~\ref{fig:mechanism-novelty}). The bias is structural and not caused by lack of effort or strategy. Adding tools to inform the search improves uncertainty awareness, but the agent is still notably bound by the bias. Novelty in selections matches that of a pure greedy acquisition function over the statistical models. 

\subsection{And yet, LLMs \emph{intend} to explore}
\label{sec:results-exploration}

\begin{figure}[tp]
  \centering
  \begin{subfigure}[b]{0.4889\linewidth}
    \centering
    \panel{5pt}{a}{\includegraphics[width=0.95\linewidth]{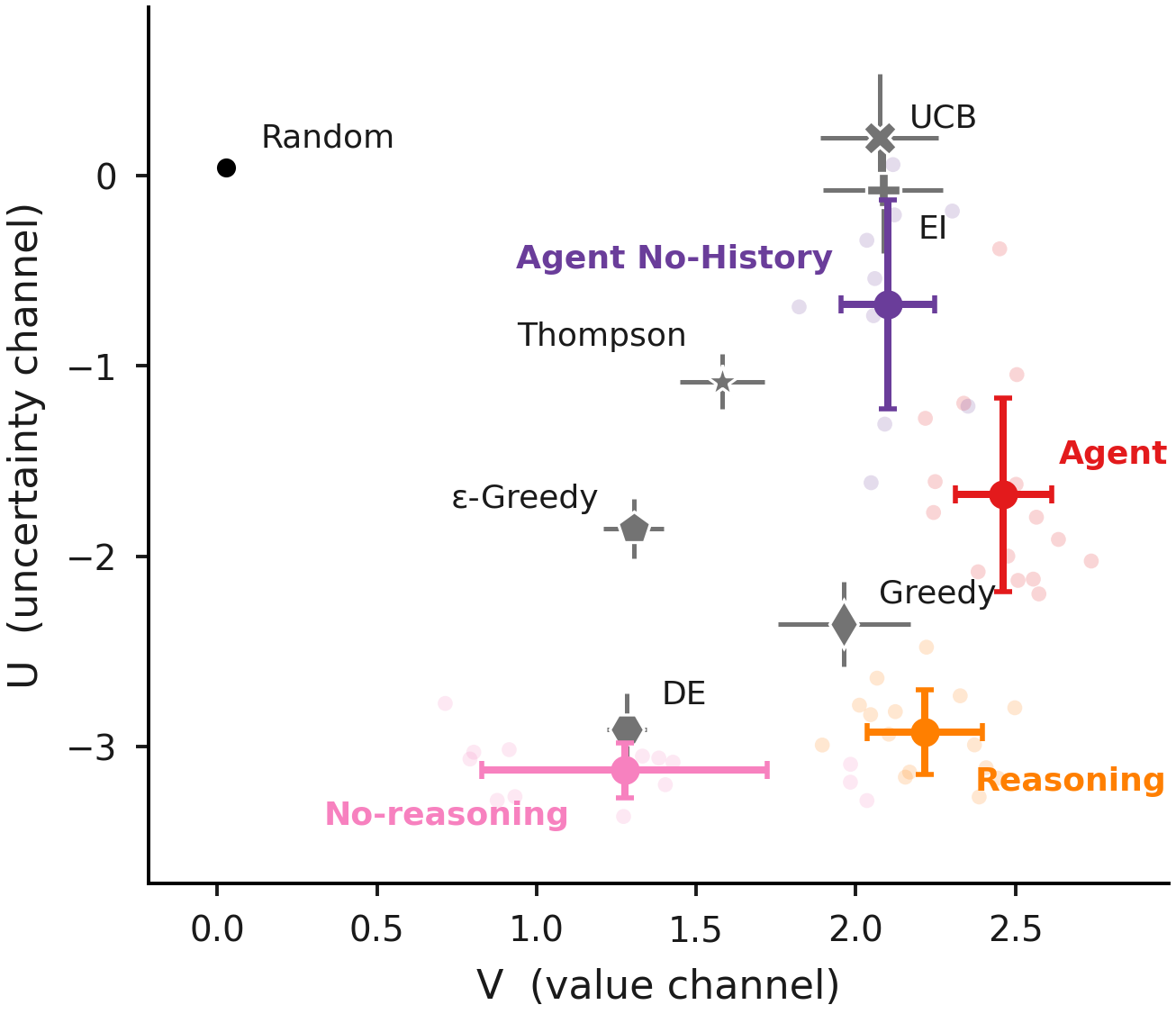}}
    \phantomsubcaption
    \label{fig:mechanism-vu}
  \end{subfigure}\hfill
  \begin{subfigure}[b]{0.4889\linewidth}
    \centering
    \panel{3pt}{b}{\includegraphics[width=0.95\linewidth]{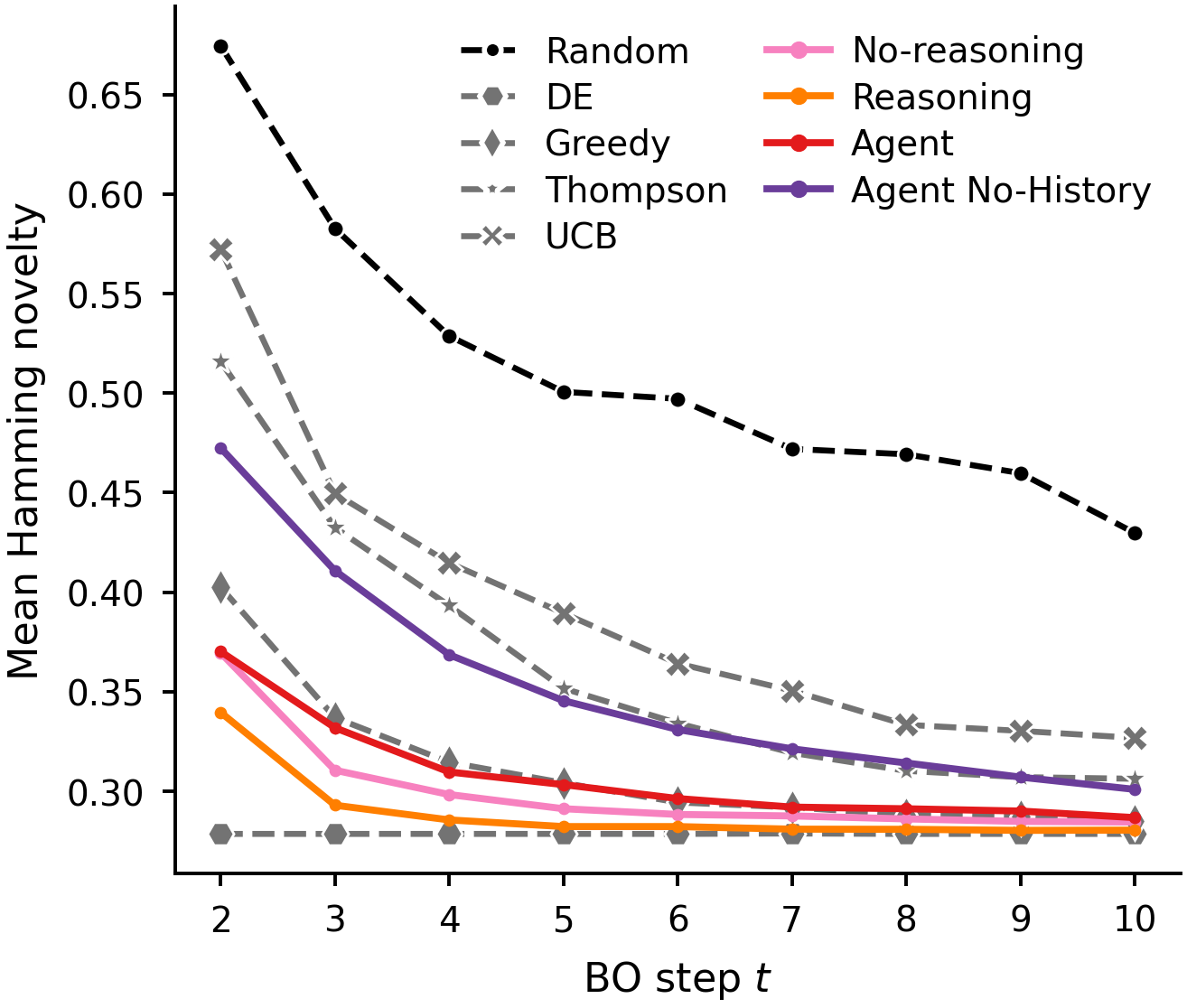}}
    \phantomsubcaption
    \label{fig:mechanism-novelty}
  \end{subfigure}\\[0pt]
  \begin{subfigure}[b]{\linewidth}
    \centering
    \panel[-3pt]{3pt}{c}{\includegraphics[width=0.95\linewidth]{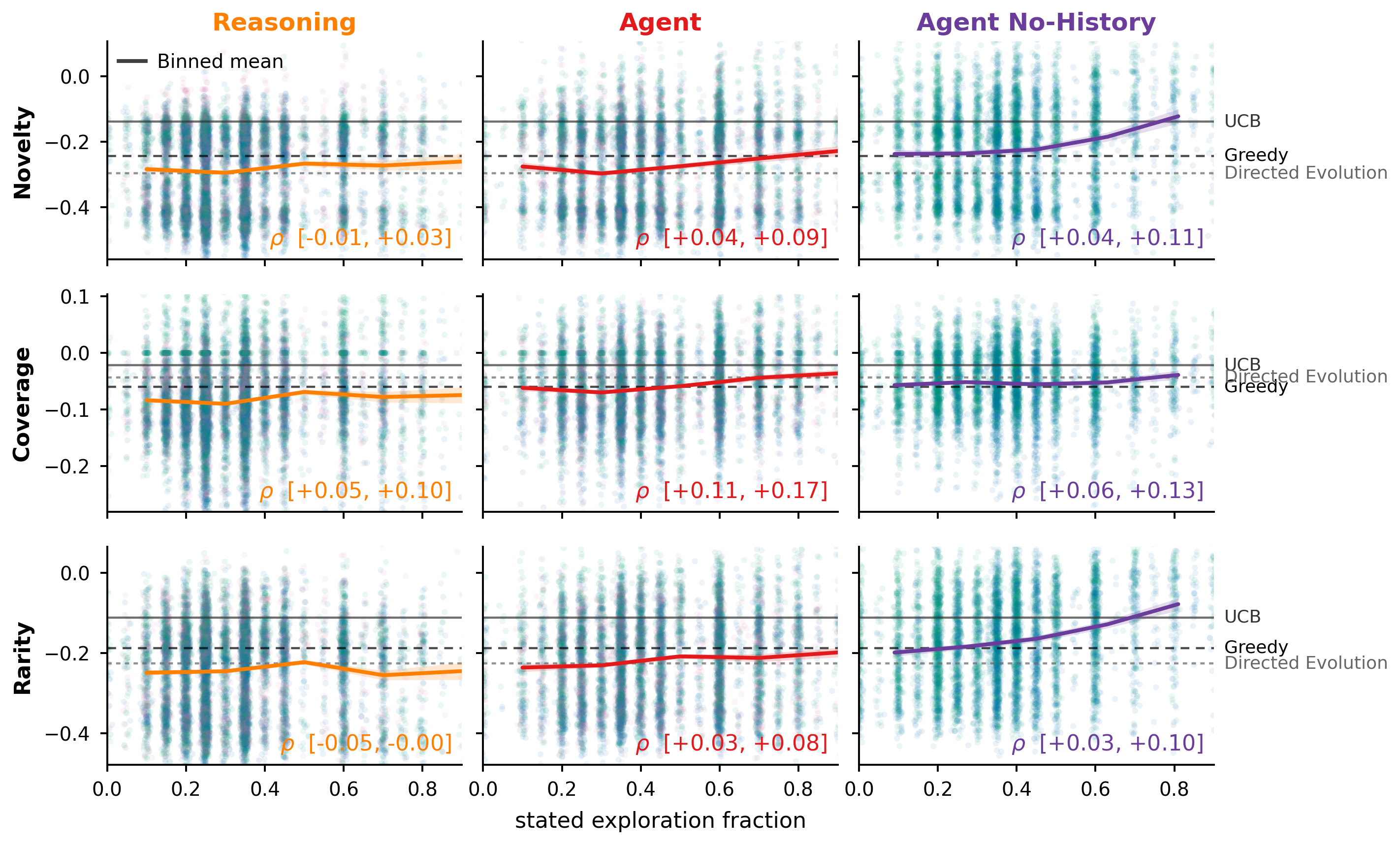}}
    \phantomsubcaption
    \label{fig:mechanism-explore}
  \end{subfigure}
  \caption{LLMs act as local, uncertainty-avoiding optimizers, yet intend to explore. \textbf{(a)} Every selection decomposed into a value feature $V$ and an uncertainty feature $U$ through a shared GP surrogate. High $V$ is value-seeking; high $U$ is uncertainty-seeking. The six statistical policies span the uncertainty axis as expected, placing greedy and directed evolution (DE) at the bottom and UCB and EI at the top. Reasoning LLMs sit below every predictive policy tested; non-reasoning models sit at or below DE. The agent recovers roughly half the distance in the $U$-axis to the uncertainty-based statistical policies, and the agent with the history removed reaches the $U$-range of a Thompson sampler (Section~\ref{sec:results-in-context}). \textbf{(b)} Mean Hamming distance from each proposal to the candidates already in history, per campaign step. Higher is more novel. The reasoning LLM stays closer to its own history than any policy except DE, and allowing reasoning enforces the bias against exploration rather than relieving it. \textbf{(c)} Cycle-corrected correlation between the exploration intent extracted from each reasoning trace and three realized exploration proxies (novelty, coverage gain, rarity), for reasoning LLMs, agents, and agents with the history removed. Each point is one campaign step. 13{,}550 traces, first step of each campaign omitted. For the reasoning LLM, the correlation is indistinguishable from zero on novelty and rarity; giving it a tool that provides an uncertainty estimate makes all three positive. The intent to explore is honest and enacted. The failure is an intention-ability gap, not an intention-action gap. Coloured lines are binned means with 95\% intervals. Grey reference levels mark UCB, greedy, and DE.}
  \label{fig:mechanism-decomp}
\end{figure}

The reasoning LLMs nonetheless invoke exploration in 92\% of all traces and reliably act on their intentions to explore (Section~\ref{sec:hypothesis}). To understand this seeming paradox, we extract an explore/exploit intent in [0,1] from each reasoning trace with an LLM judge, and correlate the intent to explore against three simple exploration proxies derived from the actions actually taken: \textbf{rarity}, \textbf{Hamming novelty}, and \textbf{coverage gain} (Figure~\ref{fig:mechanism-explore}). All metrics and intent are adjusted for a cycle trend in which exploration is artificially easier and more common early on in the campaign (Methods~\ref{sec:methods-exploration-intent}).

For reasoning models, the correlation between stated intent and realized exploration is zero for rarity and novelty and only weakly positive for coverage gain (Figure~\ref{fig:mechanism-explore}). A reasoning model's stated intention to explore does not track its realized information gain. When the model is given the tool (agent), the correlation becomes positive, indicating that injected statistical uncertainty metrics can help alleviate locality bias, as previously suggested~\citep{krishnamurthy_can_2024}. Crucially, a substantial part of the failure is transferred to the agents, relative to their own tools (UCB, EI). The intent to explore is honest and enacted and the failure is not in a intention-action gap.

\subsection{In-context data drive stickiness}
\label{sec:results-in-context}

To test that LLMs are context-sticky causally, we remove the history of validated candidates from the agent's prompt. The agent retains full access to the statistical tool and can still learn from the data, but only through the tool, never from raw data points in context (Figure~\ref{fig:setup-configs}, \textbf{no history agent}). The tools recommend candidates using chosen strategies, but the model still selects the final batch to submit and can still impose its priors.

Removing the history has mixed effects on performance. Comparing agents with and without in-context data on the benchmark, we expect that, without history in default prompt mode, net performance should be lower on average across the datasets, as the agent can no longer update its prior effectively. Any such difference in score can not be statistically determined here (See \ref{si:icl-filter}). In blind mode for example, differences in score between agents with and without history varies greatly between datasets. On datasets that reward local search, the standard agent's stickiness may help, whereas datasets that require wider exploration may benefit from the agent not hugging the history points. However, while there is a positive correlation between the improvement of removing the history, and UCB (uncertainty aware optimizer) - DE (local optimizer) difference (Figure \ref{fig:si-DE-agent}) the error is wide on 7 datasets ($\rho=0.6,\ $CI$_{95\%}[-0.25, +0.96]$ by normalized fitness and recall AUC) and more datasets would be needed to say this conclusively. 

In the $(V,U)$ plane, the effect is clear. Reasoning models sit in the uncertainty-avoiding region, the agent halfway towards the statistical policies, and the agent without history (no-history agent) sits at roughly the $U$-range of a Thompson sampler (Figure~\ref{fig:mechanism-vu}). Removing history further strengthens the correlation between the agent’s stated intent to explore and its realized information gain (Figure~\ref{fig:mechanism-novelty}), suggesting the effect is behavioral rather than purely mechanical as we restrict the model’s options.

\clearpage

\section{Discussion}
\label{sec:discussion}

While LLMs benefit in expectation from the priors, the benefit is variable and occasionally harmful. The greatest limitation to LLM performance in these settings is not strategy or domain knowledge, but a mechanical inability to identify high-uncertainty, information-rich areas of the search space. Across all models, stated intention to explore has little to no correlation with the actual information gain of the LLMs' actions. Providing LLMs with uncertainty-informed tools alleviates the issue but doesn't fully mitigate it. The cause is what we call "context-stickiness," whereby LLMs ground their reasoning and decision-making in the history data points — a tendency that manifests as overreaction to new data. 

Taken together, these results give a comprehensive diagnostic of LLM behavior in sequential decision making in prior- and data-heavy domains, complementing reports that LLMs entrench without data signals~\citep{he_martingale_2025}. The findings also add nuance to the claim that ``LLM-based agents show no sensitivity to experimental feed back'' \cite{gupta_llms_2025}, a conclusion drawn from shuffling the labels in the history and observing little change in performance. Another possible explanation is that the LLMs \emph{are} sensitive to feedback but are context-sticky. Shuffling the labels leaves the set of points in-context intact and may therefore not sufficiently alter the model's search strategy. However, removing the history data entirely aligns the intention of LLMs to explore with realized information gain. The context-stickiness is also orthogonal to the domain prior. It is structural and present regardless of whether the prior is correct or how strong the prior is. However, it can be difficult to separate the two. 

The failure is a competence gap, not a limitation in the LLM's intention to explore. The distinction matters for intervention design. Previously identified gaps in strategy and intention respond well to increased prompt instructions and to reward shaping, which is why RL fine-tuning narrows the knowing-doing gap on bandits~\citep{schmied_llms_2025, nie_evolve_2025}. A competence gap does not, which is why an explicit exploration instruction leaves every exploration metric unchanged (Section~\ref{si:nudge}), while supplying an external uncertainty estimate partly improves behavior.

In-context trajectory data induces context stickiness that keeps LLM decision-making suboptimal, and removing that context removes the failure mode. The results are consistent with previous reports on bandit problems and earlier models (Gemma 2, GPT4)~\citep{krishnamurthy_can_2024, schmied_llms_2025}). 

Many recent successful LLM-BO models~\cite{han2026chembomas, rankovic2025largelanguagemodelsuncertaintycalibrated} have solved for this issue: they have removed this confound between context and prior, either by converting the LLM prior to a statistical score to use with a statistical model, or by converting numerical data and uncertainty proxies into semantic text. These interventions allows the priors to act on the search without introducing the bias.

To support these behavioral findings, we contribute a measurement-noise correction for the Martingale and excess-movement diagnostics, reusable by anyone diagnosing belief dynamics. We also introduce a reconstructed dense NFA dataset. 

With current frontier models, it is important to decouple LLM priors from LLM decision-making under uncertainty and directly inform the construction of LLM-informed optimizers and general LLM agents in the biochemical domain. Future work should aim at designing methods that allow LLMs to reason rationally across data and semantic priors simultaneously. This study is limited in scope to settings with combinatorial search spaces and small budgets, but given similar exploration failures already reported in bandit and forecasting settings, and context-stickiness would likely occur in domains other than chemistry and materials science. The conclusions also provide nuance to an ongoing discussion about the role of single, large-context agents vs multi-agent systems in scaling autonomous research~\cite{kim2026sciencescalingagentsystems}. More information in larger context windows does not simply scale capability; it may change model behavior, and our results show the change is not always positive.
\section{Methods}
\label{sec:methods}

\subsection{Optimization domains and statistical surrogates}
\label{sec:methods-domains}

We benchmarked LLMs on 7 datasets spanning 5 domains of biochemistry: reaction optimization (EDBO)~\citep{shields2021edbo}, protein-motif optimization (ALDE)~\citep{yang2025alde}, peptide self-aggregation~\citep{Teijlingen2021tripeptides}, reaction catalysis optimization (OER)~\citep{rohr2020oer}, and synthesis of non-fullerene acceptors (NFA)~\citep{lopez2017nfa}. Organic–inorganic perovskites~\citep{hase2021gryffin} were initially included but were removed from the main benchmark result, as the LLMs showed strong evidence of memorization.

We used the direct-arylation (4 bases, 12 ligands, 4 solvents, 3 concentrations, 3 temperatures, 1,728 data points total, 100\% coverage) and aryl-amination-2b (3 aryl halides, 22 additives, 3 bases, 4 ligands, 792 data points in total, 100\% coverage) datasets from EDBO~\citep{shields2021edbo}, as these show the greatest difference between the greedy and EI acquisition functions. They are also less studied than the Suzuki reaction, which the LLMs appeared to know well.

For ALDE~\citep{yang2025alde}, we used the TrpB (4 mutation sites with 20 amino acids, 159,129 data points in total, 99.5\% coverage) and GB1 (4 mutation sites with 20 amino acids, 149,361 data points in total, 93.4\% coverage) datasets. 

For OER, we used the full set of 2,121 data points on the 3496 plate. The optimization landscape is a simplex where the task is to select a mix of up to four metals, with ratios in 0.1 increments.

We used the dropped perovskites dataset (16 organics, 4 anions, 3 cations; 192 data points in total; 100\% coverage) exclusively in an ablation study to illustrate how our suite could detect memorization.

The NFA dataset of $\sim$4200 data points, referenced in the Gryffin paper~\citep{hase2021gryffin}, has not been shared. We therefore constructed a new dense subset of NFAs from the original study~\cite{lopez2017nfa} (12 terminals, 8 spacers, 4 cores, 301 data points in total, 78.4\% coverage). The original study contains $\sim$51k NFA constructs made from a set of available molecular fragments. We reconstructed most fragment attachment points from the original paper, and we reconstructed an additional 4 by matching substructures in the larger dataset. No mapping from fragments to constructs exists. We matched the fragments using \texttt{rdkit 2026.3.2} against the large dataset by generating all possible constructs from the fragments in the pattern terminal-spacer-core-spacer-terminal, including all E/Z isomers and fragment orientations. The final subset of 301 compounds was selected to optimize coverage and size.

The LLMs knew the general settings of the 7 selected datasets. Still, they did not appear to know the actual optimal candidates or any specifics that would significantly contaminate the tests (except for perovskites).

We fetched each dataset from a paper proposing a statistical model for BO, optimized and validated on the respective dataset. We pulled these exact models in their entirety, upgraded them only as needed to \texttt{Python 3.13.5}, and used them directly as baselines. The model from the tripeptides paper~\cite{Teijlingen2021tripeptides} is a stack of two models operating on two different encodings (Judred and Mordred). While it likely performed well on its specific task of finding the top 20 candidates as quickly as possible, we could not achieve strong performance on our recall and fitness tasks. The optimized GP used in the EDBO domain, operating on the Judred encodings, performed significantly better than the published model and was used instead. We obtained similar results on the perovskites and NFA datasets used in the Gryffin paper~\cite{hase2021gryffin}. The EDBO GP significantly beat the distributed Python package \texttt{Gryffin 1.0.0} on both datasets. We therefore also used the EDBO GP model, operating on one-hot encodings, on the Gryffin datasets. Where these were not already available, we added expected improvement (EI), Thompson sampler (TS), and upper confidence bound (UCB) to the set of acquisition strategies to enable cross-dataset comparisons. We also implemented $\epsilon$-greedy (greedy acquisition with an $\epsilon=5\%$ probability of selecting a candidate at random) and directed evolution (DE, i.e., randomly selecting a candidate at Hamming distance 1 from the best-performing candidate) as behavioral baselines.

UCB sampled the top candidates from the score $s=\mu +\beta^{1/2}\sigma$, using $\beta=4$. Expected improvement sampled the candidates that maximizes $\mathbb{E}[\max(0,f(x)-f^\star)] = (\mu - f^\star),\Phi(Z) + \sigma,\phi(Z), \quad Z = \frac{\mu - f^\star}{\sigma}$, where $f^\star$ is the largest candidate found so far, $\Phi$ and $\phi$ are the CDF and PDF respectively. One dataset needed a custom Thompson sampler (TS). For OER we directly used the joint Gaussian-process posterior $f(X)\sim\mathcal{N}\big(\boldsymbol{\mu},\boldsymbol{\Sigma}\big),
\boldsymbol{\mu}\in\mathbb{R}^{n},
\boldsymbol{\Sigma}\in\mathbb{R}^{n\times n}$, with full covariance $\Sigma$. For a batch of size $b$, we pick the argmax candidate of $b$ independent functions as
$$
X_i = \arg\max_{X ,\notin, {X_1,\dots,X_{j-1}}}  \mathbf{f}^{(j)}(X), \quad \mathbf{f}^{j}\sim\mathcal{N}(\boldsymbol{\mu},\boldsymbol{\Sigma}),
\qquad j=1,\dots,b
$$

Table \ref{tab:domains} shows the exact settings used for each campaign.

\begin{table}[t]
\centering
\footnotesize
\setlength{\tabcolsep}{4pt}
\renewcommand{\arraystretch}{1.25}
\caption{Per-benchmark configurations. LLM campaigns use $n_{\mathrm{init}}{=}0$; statistical runs draw the first batch at random. New acquisitions are manually implemented over the surrogate posterior.}
\begin{tabular}{llrlcccll}
\toprule
Study & Dataset & $|\mathcal{X}|$ & $N_{dim}$ & Batch & Budget & Cycles & Surrogate & New Acquisitions \\
\midrule
\multirow{2}{*}{ALDE}
& TrpB & 159,129 & 4 & 10 & 100 & 10 &  ALDE DNN & EI \\
& GB1  & 149,361 & 4 & 10 & 100 & 10 & ALDE DNN & EI \\
\midrule
\multirow{2}{*}{EDBO}
& direct\_arylation   & 1,728 & 5 & 5 & 50 & 10  & EDBO GP & \\
& aryl\_amination\_2b & 792   & 4 & 5 & 50 & 10 & EDBO GP  & \\
\midrule
Tripeptide
& tripeptides & 8,000 & 3 & 5 & 50 & 10 &  EDBO GP &  \\
\midrule
\multirow{2}{*}{Gryffin}
& NFA         & 301 & 3 & 3 & 30 & 10 &  EDBO GP &  \\
& perovskites & 192 & 3 & 3 & 30 & 10 &  EDBO GP & \\
\midrule
OER
& 3496 & 2,121 & 10-binned & 5 & 50 & 10 &  OER GP & UCB, TS \\
& & &  6-simplex & & & & & \\
\bottomrule
\end{tabular}
\label{tab:domains}
\end{table}

\paragraph{Scope of the Bayesian reference}
We acknowledge the limits of using Bayesian behavior as a rational behavior reference. We do not claim that Bayesian policies are the most optimal policy for the scientific tasks, and that optimal policy remains an open question. We used the Bayesian reference in two separate roles: an information gain efficiency diagnostic and a performance diagnostic. The martingale property that the former tests holds independently of whether Bayesian updating is the best policy for the task. An agent can satisfy it and still optimize poorly, and it is possible to optimize the task well while violating it. This behavioral diagnosis is valuable because it evaluates an agent's belief trajectory against itself rather than against the unknown optimum, so that it can flag a pathologically updating agent and provide failure mechanism from its behavior alone.  

\subsection{Language models}
\label{sec:methods-models}

We ran 5 frontier LLM models as described in \ref{tab:llm_config}. All Anthropic runs cost \$515,26; OpenRouter (Qwen) cost \$211; OpenRouter (Kimi) cost \$192; Z.AI cost \$241,27; and OpenAI cost \$1256,57. The entire test suite contained $\sim$100k API calls and yielded $\sim$40k reasoning traces for analysis.

\begin{table}[t]
\centering
\footnotesize
\setlength{\tabcolsep}{4pt}
\renewcommand{\arraystretch}{1.25}
\caption{LLM configuration (reasoning-enabled). Temperatures and topP are all provider defaults. }
\begin{tabular}{lllllll}
\toprule
Model & Model ID & Provider & Temp. & topP & Thinking & Max tokens out \\
\midrule
GPT-5.4 & \texttt{gpt-5.4-2026-03-05} \cite{openai2026gpt54}& OpenAI & & 1.0 & \texttt{effort=medium} & 25,000 \\
Sonnet 4.6 & \texttt{claude-sonnet-4-6}\cite{anthropic2026sonnet46} & Anthropic & 1.0 & & thinking budget 25k & 32,000 \\
Kimi 2.5 & \texttt{moonshotai/kimi-k2.5}\cite{moonshot2026kimi25} & OpenRouter & 1.0 & 1.0 & thinking budget 25k & 25,000 \\
Qwen3.5 & \texttt{qwen/qwen3.5-397b-a17b} \cite{qwen2026qwen35} & OpenRouter & 1.0 & 1.0 & thinking budget 25k & 25,000 \\                                                                  GLM-4.7 & \texttt{glm-4.7} \cite{zhipu2026glm47} & Z.AI & 1.0 & 0.95 & thinking & 25,000 \\
\bottomrule                                                                      \end{tabular}
\label{tab:llm_config}
\end{table}

\subsection{Prompting modes}
\label{sec:methods-prompting}

All campaign prompts followed the same schema (Prompt \ref{prompt:schema}). We ablated the LLM’s access to domain-relevant prior knowledge through three prompting modes:

\begin{description}
\item[Default.] The LLM received the full reaction description (e.g., ``optimize the Pd-catalyzed direct arylation of imidazoles"), the identity of every search-space option (e.g., \texttt{XPhos}, \texttt{CsOPiv}, \texttt{p-Xylene}), along with generic but relevant chemical descriptors (charge, size, pKa), and may freely use prior knowledge of similar systems. This mode admits both legitimate transfer of chemical priors and inadvertent memorization-driven cheating.   
\item[Alias.] The reaction class is preserved (``optimize a chemical reaction") but specific component names in the search space were replaced with generic labels (\texttt{ligand\_1, ligand\_2, base\_1, ...}). Chemical descriptors remained. This isolated the LLM's general chemistry-strategy priors from compound-specific memorization. The descriptors were general enough that reconstruction was rare.
\item[Blind.] The optimization context was removed entirely. The LLM saw a one-hot combinatorial search space, described in domain-agnostic terms (e.g., \texttt{car}, \texttt{animal}, \texttt{tree}), with options such as \texttt{Lamborghini}, \texttt{birch}, \texttt{bird}. The LLM was tasked with optimizing an unknown function without knowledge of the underlying domain.
\end{description}

Amino acids were easily reconstructed from any descriptors. We therefore left amino acids without any descriptors in both default and alias mode (ALDE and tripeptides). The amino acids were not replaced with an aliased “amino-acid1” label. When possible, we extracted chemical descriptors from the respective datasets (EDBO, OER, perovskites). We  derived descriptors for NFA using \texttt{rdkit 2026.3.2} (mass, R$_g$) and \texttt{pySCF 2.13.0} (HOMO, LUMO, dipole). Final prompt examples are provided in the appendix (Section~\ref{sec:prompt-appendix}).

To avoid position and name bias in the search space table, the order and alias/ blind new labels were shuffled between seeds. 

\subsection{Reasoning and agent modes}
\label{sec:methods-reasoning}

We ablated the LLM’s access to various capabilities informing its reasoning:

\begin{description}
\item[No-reasoning.] The LLM had reasoning turned off and was prompted to one-shot the proposed candidates. (Prompt \ref{prompt:none})
\item[Reasoning.] Standard reasoning model, medium reasoning when that mode was possible.
\item[Agent.] The model was equipped with a single domain-agnostic tool harness that wrapped the domain-specific statistical model the LLMs were benchmarked against. The \texttt{filter\_candidates} tool trained the domain-specific prediction model, and the LLM selected an acquisition function (EI, UCB, or greedy) and relevant hyperparameters (ei-jitter, ucb-beta) for candidate filtering. A second filter across candidate parameters also allowed including or excluding specific candidates or patterns simultaneously.
\end{description}

\subsection{Bayesian optimization campaigns}
\label{sec:methods-campaigns}
Each (dataset, model, prompt-mode, reasoning-mode) condition was repeated over 15 independent seeds, after which the mean and variance converged. Statistical model baselines used 75 seeds to improve precision.

\subsection{Performance metrics}
\label{sec:methods-metrics}

Two trajectory-level metrics are computed at every step $t$:
\begin{description}
\item[Best-yet fitness.] $f^t = \max_{i \le t} f(x_i)$, the best observed objective value up to step $t$.
\item[Top-2\% recall.] The fraction of the dataset’s top-$2\%$ points discovered up to step $t$.
\end{description}

For each metric, we computed the area under the trajectory curve (AUC). To reduce the variance from dataset difficulty, we normalized AUC against two baselines:

\begin{equation}
\widetilde{\mathrm{AUC}} = \frac{\mathrm{AUC} - \mathrm{AUC}_{\mathrm{random}}}{\mathrm{AUC}_{\mathrm{best\text{-}stat}} - \mathrm{AUC}_{\mathrm{random}}},
\label{eq:methods-norm-auc}
\end{equation}

so that random selection scored 0 and the best statistical model scored 1. We defined the best statistical model as the one with the highest fitness AUC and, in case of ties, the highest recall AUC. Different acquisition functions performed best on different datasets. Values $>1$ were admissible and indicated that the LLM outperformed the best statistical baseline on that dataset. We then averaged the normalized AUC across models within each dataset, and then across datasets. We could also average across LLM models or modes. We propagated variance at each step by bootstrapping on seeds (N=10,000).

\subsection{Hypothesis extraction}
\label{sec:methods-hypotheses}

To investigate the validity and development of hypotheses stated by the LLMs at different steps along the trajectories, we extracted hypotheses and grouped them into exploitative, exploratory, and avoiding. We used GPT-5-mini to extract a list of hypotheses in a structured schema, including which search-space labels each hypothesis targeted (Prompt~\ref{prompt:extraction}). We limited extracted hypotheses to individual search-space labels rather than individual candidate combinations. For example, we considered the semantics of “ion=F” and “temperature=120”, but never the semantics of “ion=F and temperature=120”. We then computed how the LLMs’ actions and hypotheses developed across the campaigns in the different prompt modes, measuring:

\begin{itemize}
\item \textbf{LLM hypothesis-selection behavior:} $P(\mathrm{selected}\mid\mathrm{exploit\text{-}endorsed})$, $P(\mathrm{selected}\mid\mathrm{explore\text{-}endorsed})$, and $P(\mathrm{selected}\mid\mathrm{avoid\text{-}warned})$.
\item \textbf{LLM hypothesis frequency:} fraction of supporting hypotheses per selected sample, per hypothesis category.
\item \textbf{LLM grounding:} fraction of labels referenced by hypotheses already in the validated set.
\end{itemize}

Several limiting biases affected how the extraction model categorized hypotheses: many hypotheses were difficult to categorize or label; many affected a large number of candidates, so many candidates carried multiple overlapping hypotheses of different types; and cross-dimension effects were dropped, as noted above. For a hypothesis such as there seems to be an inverse correlation between temperature and aromatic compounds'', the extractor could extract the hypothesis but label it other’'.

\subsection{Belief estimation from reasoning traces}
\label{sec:methods-belief-stated}

Throughout, we defined a belief'' across labels as the marginalized probability the model assigns to a given label being part of the best’’ candidate, without making immediate assumptions about what that means to the model. We restricted the analysis to marginalized beliefs rather than full conditional beliefs over the combinatorial search space, to obtain a sufficiently dense signal, at the cost of losing cross-dimension interactions. To be explicit: we captured temperature 100 is good, I is good, F is bad'', but not temperature 100 is good when paired with ion I but bad when paired with F’'.

We estimated the model’s belief using an LLM judge. Given a reasoning trace and LLM output for a step, GPT-5-mini is asked to assign a probability $b_i$ of each label $i$ in a dimension being best''. This was repeated for each dimension in the search space at the given step. If the probabilities extracted across a dimension summed to within 5\% of 1, we projected them to the simplex; if the error was larger, or there was a parsing issue, the judge was prompted again with a relevant rule restated at the end of the prompt (e.g.\ probabilities have to sum to 1’'). The judge was told to act as an independent observer, and the prompt was inspired by Wang et al.~\cite{wang2024calibratingverbalizedprobabilitieslarge}. The exact prompt is available in Section~\ref{prompt:extraction}. One dataset (OER) had a search space that made LLM hypotheses extremely broad and difficult to measure with confidence; we therefore dropped OER from the belief analysis.

We ran several tests to validate the judge:

\begin{itemize}
\item \textbf{Judge-model ablation:} the same extraction was performed using qwen3-8B and haiku-4.5, all without reasoning, on all prompt modes for GPT-5.4 on one dataset (direct arylation). Correlations among the three judges were measured using Pearson-$ \rho$ and Spearman-$ \rho$. When computing correlations, we considered only beliefs $b > 0.02$ to reduce noise near 0. Extracted beliefs should be consistent across judges.
\item \textbf{Belief validity against oracle:} we computed Brier, ECE, and AUROC for the stated beliefs against the ground-truth oracle. Because these metrics are binary, the fitness landscape varied across datasets, and we marginalized beliefs per label, we defined a label as ``good/strong’’ if the mean fitness of all candidates using that label was in the top third of all label means. We gave such labels a score of 1, and the remaining two-thirds a score of 0. Beliefs should correlate with strong labels.
\item \textbf{Uncertainty validity:} belief uncertainty was measured by the Gini index of the marginal probability ($1-\sum_i b_i^2$). To test it, we correlated it with the variance of a ridge linear model fit to the history of validated candidates $H_t$, $h(x) = x^\top (H_t^\top H_t + \lambda I)^{-1} x$, with $\lambda = 1$. For comparison with the per-dimension marginalized beliefs, we computed the expected variance over candidates sharing a label, $\bar{h}(v) = \mathbb{E}_{v \in x}[h(x)]$. For very large search spaces such as ALDE ($140\text{k}+$), we evaluated a random subsample of 4000 data points. Belief uncertainty should correlate with linear-model variance.
\item \textbf{Model acts on beliefs:} we correlated belief with the action probability $q(x)$ (described below). A firmer belief in a label should increase its likelihood of selection.
\end{itemize}

\subsection{Belief estimation from actions}
\label{sec:methods-belief-action}

Previous work has shown that LLMs tend not to fully verbalize their beliefs, leading to a discrepancy between reasoning and actions \cite{yamin2026agentssaythinganother}. We therefore also derived a belief directly from the actions taken. Given that selections are sparse (typically less than 10\% of the search space is explored) and multidimensional, we used a smoothed estimate of the marginalized acquisition distribution:

\begin{equation} \label{eq:methods-belief-action}
q_t(x)=  \prod_i^N q_t^i(x_i), \quad q_t^i(x_i)=\frac{n_i}{n_i+b},U[n_i]+\frac{b}{b+n_i},\hat{p}_i(x_i),
\end{equation}

where $N$ is the number of components, $n_i$ is the number of labels in component $i$, $b$ is the batch size, $U$ is a uniform prior, and $\hat{p}_i(\ell) = \sum_j^b \mathbb{1}[x_{ij}=\ell]$ is the empirically estimated marginal, obtained by counting labels in the proposals.

There are two limitations. First, with a large number of labels within a component (as in amino acid domains), the uniform prior may drown out the marginal signal. Second, we marginalized beliefs per component rather than per cell, which simplified comparison with stated beliefs and yielded a denser signal but lost important coupled relationships, e.g., at epistatic protein-binding sites.

\subsection{Belief dynamics}
\label{sec:methods-dynamics}

Here, we tested whether an agent's belief update used all the information already available to it, drawing on the property that a rational agent should not be able to predict the direction of its next belief update from what it already knows, since any such predictable updates should have been incorporated in the previous cycle (the "Martingale property").   

Consider BO steps $t \in {1,\dots,T}$ at which the agent observed a signal $s_t = (x_t, y_t)$, which is added to the history $H_t = (s_1,\dots,s_t)$. Its belief $\pi_t$ is a probability vector over the values of one search-space dimension (prior $\pi_0$). The coordinate $b_t \equiv \pi_t(c,v)$ is the marginal probability of value $v$ of component $c$ belonging to the best candidate. A rational (Bayesian) agent’s belief is a martingale with respect to the observation filtration $\mathcal{F}_t = \sigma(H_t)$ coordinate-wise,
\begin{equation}
\mathbb{E}[\pi_{t+1} \mid \mathcal{F}_{t}] = \pi_{t}
\qquad\Longrightarrow\qquad
\mathbb{E}[\Delta b_t \mid \mathcal{F}_{t}] = 0,
\qquad \Delta b_t = b_{t+1}-b_t,
\label{eq:methods-martingale}
\end{equation}
with the implication of an equivalence when imposed on every coordinate $(c,v)$ jointly. Simply put, the next belief update is not predictable from the information available at time $t$. We quantified departures from this null with two diagnostics.  

\paragraph{Martingale score (He et al.)}
We tested Eq.~\ref{eq:methods-martingale} directly via the slope of the update on the prior belief, $\Delta b_t = \beta_1 b_t + \beta_0 + \epsilon_t$, with $\beta_1 = 0$ under rational updating, $\beta_1 > 0$ entrenchment, and $\beta_1 < 0$ over-reaction~\cite{he_martingale_2025}.

\paragraph{Belief movement (Augenblick–Rabin).}
Define the belief movement $m$ and uncertainty reduction $r$ in Gini form over a window $[t_a, t_b]$:
\begin{equation}
m = \sum_{\tau=t_a+1}^{t_b}\lVert \pi_{\tau} - \pi_{\tau-1}\rVert^2,
\qquad
u_t = 1 - \sum_v \pi_{t,v}^2,
\qquad
r = u_{t_a} - u_{t_b}.
\label{eq:methods-movement}
\end{equation}
Conditioning on $\mathcal{F}_{\tau-1}$ and applying
Eq.~\ref{eq:methods-martingale} we get
\begin{equation}
\mathbb{E}\left[\lVert\pi_\tau-\pi_{\tau-1}\rVert^2 \mid
\mathcal{F}_{\tau-1}\right]
= \mathbb{E}\left[\lVert\pi_\tau\rVert^2\mid\mathcal{F}_{\tau-1}\right]
- \lVert\pi_{\tau-1}\rVert^2
= u_{\tau-1} - \mathbb{E}[u_\tau\mid\mathcal{F}_{\tau-1}],
\label{eq:methods-ar-identity}
\end{equation}
It then follows that $\mathbb{E}[m] = \mathbb{E}[r]$, with no assumption on the signal distribution~\cite{AugenblickRabin2021BeliefMovement}. We considered the per-step excess-movement statistic $Z = \bar{m} - \bar{r}$. $\mathbb{E}[Z] = 0$ under the null, $Z > 0$ over-reaction, $Z < 0$ entrenchment.

\paragraph{Measurement-noise correction.}
\label{sec:methods-correction}
Both diagnostics are computed from noisy belief estimates, whether from an LLM judge or from actions, introducing errors-in-variables biases that neither source corrects for. Augenblick and Rabin derive the resulting $2\sigma^2_\varepsilon$ inflation of excess movement but treat it as a calibration argument rather than a correction \cite{AugenblickRabin2021BeliefMovement}. He. et.al \cite{he_martingale_2025} does not mention any bias.

We modelled the observed belief as $b_t = \beta_t + \varepsilon_t$ with $\mathbb{E}[\varepsilon_t \mid \beta] = 0$, $\varepsilon_t \perp \varepsilon_{t'}$ ($t \neq t'$), and noise energy $\sigma_\varepsilon^2 = \mathbb{E}\lVert\varepsilon_t\rVert^2$. The corrections below require conditional mean-zero and no autocorrelation. The noise is also only an approximation near the boundary of the belief simplex, where $b_t \in [0,1]$ creates an artificial truncation. We probed the effect empirically on a rational actor.

Noise biases the two diagnostics in the same direction. For the slope, the noise $\varepsilon_t$ enters $\Delta b_t = \Delta\beta_t + \varepsilon_{t+1} - \varepsilon_t$ negatively and $b_t = \beta_t + \varepsilon_t$ positively, so
\begin{equation}
\operatorname*{plim}\hat\beta_1^{\text{naive}}
= \frac{\mathrm{Cov}(\Delta\beta_t,\beta_t)-\sigma^2}
{\mathrm{Var}(\beta_t)+\sigma^2}
\label{eq:methods-naive-bias}
\end{equation}
Hence, for a rational agent with$\mathrm{Cov}(\Delta\beta_t,\beta_t)=0, \ \operatorname*{plim}\hat\beta_1^{\text{naive}}=\frac{-\sigma^2}{\mathrm{Var}(\beta_t)+\sigma^2} < 0$

For the belief movement, each step carries $\mathbb{E}\lVert\varepsilon_\tau - \varepsilon_{\tau-1}\rVert^2 = 2\sigma_\varepsilon^2$, whereas the common $-\sigma_\varepsilon^2$ bias in $u$ cancels in the endpoint difference $r$. Hence, $Z$ is inflated by $2\sigma_\varepsilon^2$ per step. This means both metrics would assign over-reaction ($\beta_1 < 0$, $Z > 0$) to a rational agent.

\emph{Martingale slope.}
The lagged belief $b_{t-1}$ was used as an instrument with the noise $\varepsilon_{t-1}$. This noise shares no index with the noise $\varepsilon_{t+1}$, or $\varepsilon_t$, in $\Delta b_t$:
\begin{equation}
\mathrm{Cov}(\Delta b_t,, b_{t-1})
= \mathrm{Cov}(\Delta\beta_t,, \beta_{t-1}),
\qquad
\mathrm{Cov}(b_t,, b_{t-1})
= \mathrm{Cov}(\beta_t,, \beta_{t-1}).
\label{eq:methods-lag-cov}
\end{equation}
We therefore reported the estimator $\tilde{M} = \mathrm{Cov}(\Delta b_t, b_{t-1}) / \mathrm{Cov}(b_t, b_{t-1})$.

\emph{Belief movement.}
Here the extra noise must be estimated and subtracted. We defined $d_\tau = b_\tau - b_{\tau-1} = \delta_\tau + (\varepsilon_\tau - \varepsilon_{\tau-1})$ with latent increment $\delta_\tau = \beta_\tau - \beta_{\tau-1}$. As with the Martingale score, consecutive moves share exactly one noise draw, entering with opposite signs.
\begin{equation}
X_\tau = -2 d_\tau^{\top} d_{\tau-1},
\qquad
\mathbb{E}[X_\tau]
= -2\bigl(\underbrace{\mathbb{E}[\delta_\tau^{\top}\delta_{\tau-1}]}_{
=0\ \text{under null}}
-\sigma\varepsilon^2)
= 2\sigma_\varepsilon^2,
\label{eq:methods-lag2}
\end{equation}
We estimated the per-step bias as $\hat c = \max(0, \bar{X})$. The floor guarantees the correction never moves $\hat Z$ further from zero than the uncorrected $Z$. We reported $\hat Z = Z - \hat c$ and $\hat m = \max(0, m - \hat c, n_{\text{steps}})$.

\paragraph{Rational-agent (null) validation.}
We validated the corrections against a Bayes-optimal actor. Over a discrete space $\mathcal{X}={1,\dots,V}^{C}$ we draw a truth $x^\star\sim\mathrm{Unif}(\mathcal{X})$ per run and observe $y = s(x,x^\star) + \nu$, $\nu\sim\mathcal{N}(0,\tau^2)$, at each $x$, where $s$ simply counts agreeing components. The actor’s belief $b_t(c,v)=\mathbb{P}(x^\star_c = v \mid \mathcal{F}_t) =\mathbb{E}[\mathbb{1}{x^\star_c = v}\mid\mathcal{F}_t]$ ensures Eq.~\ref{eq:methods-martingale} holds exactly, and it acts on the same simplex domain as our LLM beliefs. We added readout noise $b_t=\beta_t+\varepsilon_t$, $\varepsilon_t\sim\mathcal{N}(0,\sigma^2)$ i.i.d.\ per candidate, at a level so the synthetic $\hat c$ matches the $\hat c$ measured on the judged beliefs. Table~\ref{tab:null-validation} reports the outcome, repeated in an edge-heavy regime to probe the additive-noise approximation. Noise corrections are needed in noise-heavy regimes.

\begin{table}[t]
\centering
\caption{Rational-agent (null) validation. CI computed from bootstrap across $N=300$ artificial campaigns.}
\label{tab:null-validation}
\begin{tabular}{llrcl}
\toprule
Diagnostic & Estimator & Estimate & 95\% CI & Prediction \\
\midrule
\multicolumn{5}{l}{\emph{Martingale score}} \\
& naive $M$ & $-0.0352$ & $[-0.0406,,-0.0301]$ & $-0.0397$ (Eq.~\ref{eq:methods-naive-bias}) \\
& $\tilde{M}$ & $+0.0021$ & $[-0.0029,,+0.0068]$ & $0$ \\
& $\tilde{M}$, edge-heavy & $+0.0003$ & $[-0.0011,,+0.0016]$ & $0$ \\
\midrule
\multicolumn{5}{l}{\emph{Belief movement}} \\
& naive $Z$ & $+0.0241$ & $[+0.0211,,+0.0271]$ & $+0.0250$ ($2\sigma_\varepsilon^2$) \\
& $\hat Z$ & $-0.0013$ & $[-0.0041,,+0.0015]$ & $0$ \\
& $\hat Z$, edge-heavy & $-0.0005$ & $[-0.0029,,+0.0020]$ & $0$ \\
\bottomrule
\end{tabular}
\end{table}

\subsection{Decision decomposition}
\label{sec:methods-decomposition}

To characterize the LLM’s decisions, we extended the two-armed-bandit decomposition of Gershman \cite{gershman2018humanAlgorithms}, which separates greedy, UCB, and Thompson behavior through
\begin{equation}
P(x=1 \mid \mathbf{w}) = \phi\left(w_1 \mathbf{V} + w_2 \mathbf{U} + w_3 \mathbf{V}/\mathrm{TU}\right),
\end{equation}
with $\mathbf{V} = \mu(1)-\mu(2)$, $\mathbf{U} = \sigma(1)-\sigma(2)$, $\mathrm{TU} = \sqrt{\sigma(1)^2 + \sigma(2)^2}$, and $\phi$ the standard normal CDF.

We can model decisions on the multidimensional search space as a conditional logit \cite{mcfadden1974conditionallogit} using a GP surrogate for $\mu$ and $\sigma$ estimation:
\begin{equation}
\label{eq:methods-action-model}
P(x \mid H_t) = \frac{\exp\left(V \mu(x) + U \sigma(x)\right)}{\sum_{y \ne x} \exp\left( V\mu(y) + U \sigma(y)\right)},
\end{equation}
where $\mu, \sigma$ are z-standardized features for mean and standard deviation, derived from a surrogate GP. $V$ captures value-seeking, and $U$ captures uncertainty-seeking. Note that we adopt the (U/V) naming convention for the parameters here, as defined as functions in the original paper.

We fit Eq.~\ref{eq:methods-action-model} by minimizing the ridge-penalized ($\lambda=1$) conditional
negative log-likelihood against all choices, pooled across datasets and seeds, for the reasoning, non-reasoning, agentic, and statistical models. During optimization, each Newton step was solved with Tikhonov jitter ($10^{-8} I$) and accepted via backtrack binary search until the objective decreased, iterating until a tolerance of $10^{-9}$ was reached, or $50$ iterations were performed. We used a fixed GP surrogate (\texttt{scikit-learn 1.8.0}) to derive $\mu, \sigma$, with an RBF kernel, lengthscale 2, constant output 1, and white noise 0.1. Because our total budget was much smaller than the size of the search space across all domains, the total uncertainty (TU) was essentially constant throughout the campaign; we therefore dropped it and accepted that we could not guarantee we could distinguish a Thompson acquisition function from a UCB acquisition function. We retrained the GP at every step to obtain the mean and uncertainties. To estimate the denominator, we used negative sampling, drawing $N = 64$ negative candidates uniformly from all candidates not sampled in the batch. We treated every choice in a batch as a unique selection rather than a coupled batch selection, but removed all selected candidates from the negative pool. Note that the fixed GP was used across datasets and was distinct from all dataset-specific models.

We obtained confidence bands via a nonparametric cluster bootstrap of whole campaigns (model$\times$dataset$\times$seed), with replacement and refitting on each of $B = 10{,}000$ replicates. Feature standardization was shared across all actors, making the resulting $(V, U)$ coordinates directly comparable across language models and statistical policies.

\subsection{Exploration metrics}
\label{sec:methods-exploration}

\paragraph{Search-space coding.}
Each search space is a discrete grid over $N$ components, where component $i$ takes one of $n_i$ labels; write $n_{\mathrm{tot}}=\sum_i n_i$. A candidate is the label tuple $x=(x_1,\dots,x_N)$, reusing the notation of Section~\ref{sec:methods-belief-action}. We characterized exploration of each batch by three simple metrics: how far it steps (novelty), how surprising it is (rarity), and how much of the space it opens (coverage). For each, we reported an excess over a random decision.

\paragraph{Novelty.}
The min-Hamming distance from a proposal to the evaluated history, normalized by the number of components,
\begin{equation} \label{eq:methods-novelty}
\mathrm{Novelty}=\frac{1}{N}\min_{h\in H_t}\sum_{i=1}^{N}\mathbb{1}[x_i\neq h_i]\in[0,1],
\end{equation}
averaged over the batch. It is $0$ for a re-proposal of an evaluated point and $1$ when every component differs from every point tried so far.

\paragraph{Rarity.}
We define label counts in the history as $c^i_t(\ell)=\sum_{x\in H_t}\mathbb{1}[x_i=\ell]$ and empirical marginal $\hat{p}^i_t(\ell)=c^i_t(\ell)/m_t$, giving a predictive metric
\begin{equation} \label{eq:methods-rarity-identity}
q_t^i(\ell)=\frac{c^i_t(\ell)+1}{m_t+n_i}
=\frac{n_i}{m_t+n_i},U[n_i]+\frac{m_t}{m_t+n_i},\hat{p}^i_t(\ell),
\end{equation}
which is exactly the action-belief in Eq.~\ref{eq:methods-belief-action}, with the count window widened from one batch to the full history. We define
\begin{equation} \label{eq:methods-rarity}
\mathrm{Rarity}=\frac{1}{N}\sum_{i=1}^{N}\frac{-\log q_t^i(x_i)}{\log (m_t+n_i)}\in(0,1],
\end{equation}
so that, up to normalization, rarity is the surprisal $-\log q_t(x)$ of the pick under the agent’s revealed belief. The counts exclude the batch being scored. The maximum attainable surprisal is $\log(m_t+n_i)$.

\paragraph{Coverage gain.}
Coverage gain is simply the increase in the fraction of all labels the agent has touched. With $S_i(t)={x_i : x\in H_t\cup B_t}$ the distinct labels of component $i$ seen through step $t$,
\begin{equation} \label{eq:methods-coverage}
\mathrm{Cov}(t)=\frac{1}{n_{\mathrm{tot}}}\sum_{i=1}^{N}\bigl|S_i(t)\bigr|,
\qquad \mathrm{CoverageGain}=\mathrm{cov}(t)-\mathrm{cov}(t-1)\ \ge 0,
\end{equation}

\paragraph{Excess over random decision}
The history confounded all three quantities. With growing $H_t$, novelty and coverage gain fell mechanically, and rarity drifted as the prior weight in Eq.~\ref{eq:methods-rarity-identity} decayed. We therefore reported each metric as an excess over a uniform draw from the remaining pool $P_t=\mathcal{X}\setminus H_t$, given the same history, so $0$ is chance behavior, $>0$ explores more than random, and $<0$ explores less.

\subsection{Exploration intent}
\label{sec:methods-exploration-intent}

Exploration intent was extracted from the reasoning trace by a GPT-5-mini judge for belief estimation, returning an explore fraction [0,1], an explicit / explore flag, and an explore/exploit/mixed stance (one call per cycle; traces <150 chars skipped). We validated the judge in two ways on edbo/direct-arylation: cross-judge agreement with claude-haiku-4-5 and qwen3-8b, and correlation with the hypothesis extractor’s explore share.

\subsection{Context ablation}
\label{sec:methods-prompts}
The agent could access context in two ways. The \emph{history} method was the validated-candidates section table in the prompt. The \emph{tool} provided additional information through statistical ranking and filters. When we removed the history table but kept the tool intact, the model would repeatedly apply the same priors through the tool’s include/exclude arguments, allowing the agent to restrict the ranked candidate set based on component values. This created an artificial failure since the agent could not see what filters had been imposed in previous cycles. For no-history runs, we also turned off the tool’s include/exclude argument, leaving the model with only the statistical filters.

We ablate the two coupled actions (removing history and turning off the tool filters) in a $2\times2$ test. We ran each cell on GPT-5.4 and Sonnet-4.6 across 7 datasets in default and blind prompt modes, with 10 seeds per cell.

\subsection{Prompt ablation}
\label{sec:methods-prompts-ablation}
As a separate ablation, we defined a \emph{nudged} prompt: a one-sentence exploration instruction inserted after the overview block each cycle (``You have to consider every candidate in the search space. Consider both how candidates can increase fitness and reduce uncertainty in the
search space for following cycles.‘’). We ran 10 seeds in reasoning mode across all 5 models on the 7 main datasets, in both default and blind modes.

\subsection{Statistical analysis}
\label{sec:methods-stats}

Statistics in the main text were pooled across all 7 datasets, 5 models, 3 prompt modes, and 15 seeds unless otherwise stated. All errors were bootstrapped with 95\% confidence intervals with 10,000 resamples unless stated otherwise. Bootstrapping standard errors ensures the resampling unit is honest. We drew samples first from the datasets, then from the model, then from prompt mode, and finally from the seed, with replacement. Specifically, we assessed significance in the main benchmark using a two-sided, hierarchical, BH-corrected bootstrap with 100,000 replicates; the resampling unit is the dataset. We compared each LLM capability and prompt mode against the mean normalized AUC of the statistical acquisition functions (TS, UCB, greedy, EI), pooling the 7 benchmark datasets and 5 LLM models (perovskites excluded).

\textbf{BH-correction:} We controlled the false discovery rate across families of $M$ simultaneous tests (e.g., metric $\times$ prompt mode $\times$ capability) using the Benjamini–Hochberg correction~\citep{benjamini1995fdr}. Ordering the family’s $p$-values as $p_{(1)} \le p_{(2)} \le \dots \le p_{(M)}$, the BH-adjusted $p$-value ($p_{BH}$) of the hypothesis at rank $i$ is
\begin{equation}
p_{BH,i} = \min_{,i \le j \le M}
\min\Bigl(1, \frac{M}{j},p_{(j)}\Bigr),
\label{eq:methods-bh}
\end{equation}
computed by a single descending pass that enforces $p_{BH,i} \le q_{BH,i+1}$. 

\subsection{Use of Generative AI}
\label{sec:methods-ai}
We used Claude Code and Cursor (Auto model) for all coding. The authors manually reviewed all scripts used in the final analysis. Claude Opus 4.8 detected noise bias in the belief movement and Martingale score papers during a bug search; however, the proposed solution was insufficient to correct all biases fully.

\subsection{Data availability}
\label{sec:methods-data}
All code used to analyze and generate data, along with all prompts and all LLM reasoning traces and outputs are available under an \texttt{MIT} license on Zenodo: \href{https://zenodo.org/records/22984415}{10.5281/zenodo.22984415}

\bibliographystyle{naturemag}
\bibliography{references}

@article{shields2021edbo,
  author  = {Shields, Benjamin J. and Stevens, Jason and Li, Jun and Parasram, Marvin and Damani, Farhan and Martinez Alvarado, Jesus I. and Janey, Jacob M. and Adams, Ryan P. and Doyle, Abigail G.},
  title   = {Bayesian reaction optimization as a tool for chemical synthesis},
  journal = {Nature},
  volume  = {590},
  pages   = {89--96},
  year    = {2021},
  doi     = {10.1038/s41586-021-03213-y},
  url     = {https://doi.org/10.1038/s41586-021-03213-y}
}

@article{Teijlingen2021tripeptides,
author = {van Teijlingen, Alexander and Tuttle, Tell},
title = {Beyond Tripeptides Two-Step Active Machine Learning for Very Large Data sets},
journal = {Journal of Chemical Theory and Computation},
volume = {17},
number = {5},
pages = {3221-3232},
year = {2021},
doi = {10.1021/acs.jctc.1c00159},
url = {https://doi.org/10.1021/acs.jctc.1c00159}
}

@misc{fei2026agentsfailautoresearchendtoend,
      title={How Do Agents Fail on AutoResearch: End-to-End Diagnostic Evaluation on 100 Real-World Frontier Research Tasks}, 
      author={Yanlin Fei and Nazhou Liu and Xinmiao Yu and Shaolong Chen and Lei Li and Rahul Thapa and Madalina Ciobanu and Qingqing Mao and Ritankar Das},
      year={2026},
      eprint={2608.14905},
      archivePrefix={arXiv},
      primaryClass={cs.CL},
      url={https://arxiv.org/abs/2608.14905}, 
}

@article{hase2021gryffin,
   title={Gryffin: An algorithm for Bayesian optimization of categorical variables informed by expert knowledge},
   volume={8},
   ISSN={1931-9401},
   url={http://dx.doi.org/10.1063/5.0048164},
   DOI={10.1063/5.0048164},
   number={3},
   journal={Applied Physics Reviews},
   publisher={AIP Publishing},
   author={Häse, Florian and Aldeghi, Matteo and Hickman, Riley J. and Roch, Loïc M. and Aspuru-Guzik, Alán},
   year={2021},
   month=7 }

@article{lopez2017nfa,
title = {Design Principles and Top Non-Fullerene Acceptor Candidates for Organic Photovoltaics},
journal = {Joule},
author = {Steven A. Lopez and Benjamin Sanchez-Lengeling and Julio {de Goes Soares} and Alán Aspuru-Guzik},
volume = {1},
number = {4},
pages = {857-870},
year = {2017},
issn = {2542-4351},
doi = {https://doi.org/10.1016/j.joule.2017.10.006},
url = {https://www.sciencedirect.com/science/article/pii/S2542435117301307}}

@article{tran_active_2018,
	title = {Active learning across intermetallics to guide discovery of electrocatalysts for {CO2} reduction and {H2} evolution},
	volume = {1},
	copyright = {2018 The Author(s), under exclusive licence to Springer Nature Limited},
	issn = {2520-1158},
	url = {https://www.nature.com/articles/s41929-018-0142-1},
	doi = {10.1038/s41929-018-0142-1},
	language = {en},
	number = {9},
	urldate = {2026-07-03},
	journal = {Nature Catalysis},
	publisher = {Nature Publishing Group},
	author = {Tran, Kevin and Ulissi, Zachary W.},
	month = sep,
	year = {2018},
	pages = {696--703},
}

@misc{akke2025bayesian,
  title        = {Bayesian Optimization for Biochemical Discovery with {LLM}s},
  author       = {Akke, Mattias and Yang, Soojung and Ruza, Jurgis and Song, Jinyeop and Pan, Elton and G{\'o}mez-Bombarelli, Rafael},
  year         = {2025},
  month        = nov,
  howpublished = {ChemRxiv preprint},
  doi          = {10.26434/chemrxiv-2025-w1wsh},
  url          = {https://doi.org/10.26434/chemrxiv-2025-w1wsh},
  note         = {Preprint, not peer-reviewed}
}

@article{rohr2020oer,
author ="Rohr, Brian and Stein, Helge S. and Guevarra, Dan and Wang, Yu and Haber, Joel A. and Aykol, Muratahan and Suram, Santosh K. and Gregoire, John M.",
title  ="Benchmarking the acceleration of materials discovery by sequential learning",
journal  ="Chem. Sci.",
year  ="2020",
volume  ="11",
issue  ="10",
pages  ="2696-2706",
publisher  ="The Royal Society of Chemistry",
doi  ="10.1039/C9SC05999G",
url  ="http://dx.doi.org/10.1039/C9SC05999G"
}

@article{yang2025alde,
  author  = {Yang, Jason and Lal, Ravi G. and Bowden, James C. and Astudillo, Raul and Hameedi, Mikhail A. and Kaur, Sukhvinder and Hill, Matthew and Yue, Yisong and Arnold, Frances H.},
  title   = {Active learning-assisted directed evolution},
  journal = {Nature Communications},
  volume  = {16},
  pages   = {714},
  year    = {2025},
  doi     = {10.1038/s41467-025-55987-8},
  url     = {https://doi.org/10.1038/s41467-025-55987-8}
}

@misc{han2026chembomas,
  author       = {Han, Dong and AI, Zhehong and Cai, Pengxiang and Ye, Zihao and Lu, Shanya and Sun, Shuzhou and Chen, Jianpeng and Gao, Ben and Ge, Lingli and Wang, Weida and Zhou, Xiangxin and Liu, Xihui and Su, Mao and Ouyang, Wanli and Bai, Lei and Zhou, Dongzhan and Xu, Tao and Li, Yuqiang and Zhang, Shufei},
  title        = {{ChemBOMAS}: Accelerated {Bayesian} Optimization for Scientific Discovery in Chemistry with {LLM}-Enhanced Multi-Agent System},
  howpublished = {Preprint},
  year         = {2026},
  url          = {https://openreview.net/forum?id=XEkQu1ZWGN}
}

@article{AugenblickRabin2021BeliefMovement,
  author    = {Augenblick, Ned and Rabin, Matthew},
  title     = {Belief Movement, Uncertainty Reduction, and Rational Updating},
  journal   = {The Quarterly Journal of Economics},
  volume    = {136},
  number    = {2},
  pages     = {933--985},
  year      = {2021},
  month     = {May},
  doi       = {10.1093/qje/qjaa043},
  url       = {https://doi.org/10.1093/qje/qjaa043}
}

@misc{wang2024calibratingverbalizedprobabilitieslarge,
      title={Calibrating Verbalized Probabilities for Large Language Models}, 
      author={Wang, Cheng and Szarvas, Gyuri and Balazs, Georges and Danchenko, Pavel and Ernst, Patrick},
      year={2024},
      month = {10},
      doi = {10.48550/arXiv.2410.06707},
      eprint={2410.06707},
      archivePrefix={arXiv},
      url={https://arxiv.org/abs/2410.06707}, 
}

@misc{yamin2026agentssaythinganother,
      title={When Agents Say One Thing and Do Another: Validating Elicited Beliefs from LLMs}, 
      author={Yamin,Khurram and Tang, Jingjing and Cortes-Gomez, Santiago and Sharma, Amit and Horvitz, Eric and Wilder, Bryan},
      year={2026},
      month = {5},
      eprint={2602.06286},
      archivePrefix={arXiv},
      url={https://arxiv.org/abs/2602.06286}, 
}

@incollection{mcfadden1974conditionallogit,
  author    = {McFadden, Daniel},
  title     = {Conditional Logit Analysis of Qualitative Choice Behavior},
  booktitle = {Frontiers in Econometrics},
  editor    = {Zarembka, Paul},
  publisher = {Academic Press},
  address   = {New York},
  pages     = {105--142},
  year      = {1974}
}

@misc{falck2024incontextlearninglargelanguage,
      title={Is In-Context Learning in Large Language Models Bayesian? A Martingale Perspective}, 
      author={Fabian Falck and Ziyu Wang and Chris Holmes},
      year={2024},
      eprint={2406.00793},
      archivePrefix={arXiv},
      primaryClass={stat.ML},
      url={https://arxiv.org/abs/2406.00793}, 
}

@article{Augenblick2023more,
    author = {Augenblick, Ned and Lazarus, Eben and Thaler, Michael},
    title = {Overinference from Weak Signals and Underinference from Strong Signals},
    journal = {The Quarterly Journal of Economics},
    volume = {140},
    number = {1},
    pages = {335-401},
    year = {2025},
    month = {02},
    issn = {0033-5533},
    doi = {10.1093/qje/qjae032},
    url = {https://doi.org/10.1093/qje/qjae032},
    eprint = {https://academic.oup.com/qje/article-pdf/140/1/335/61032634/qjae032.pdf},
}

@article{gershman2018humanAlgorithms,
title = {Deconstructing the human algorithms for exploration},
journal = {Cognition},
volume = {173},
pages = {34-42},
year = {2018},
month = {4},
issn = {0010-0277},
doi = {https://doi.org/10.1016/j.cognition.2017.12.014},
url = {https://www.sciencedirect.com/science/article/pii/S0010027717303359},
author = {Gershman, Samuel J.}
}

@misc{openai2026gpt54,
  title        = {{GPT-5.4} Thinking System Card},
  author       = {{OpenAI}},
  year         = {2026},
  month        = mar,
  url = {https://deploymentsafety.openai.com/gpt-5-4-thinking},
}

@misc{anthropic2026sonnet46,
  title        = {System Card: Claude Sonnet 4.6},
  author       = {{Anthropic}},
  year         = {2026},
  month        = feb,
  url = {https://www-cdn.anthropic.com/78073f739564e986ff3e28522761a7a0b4484f84.pdf},
}

@misc{moonshot2026kimi25,
  title        = {{Kimi K2.5:} Visual Agentic Intelligence},
  author       = {{Moonshot AI}},
  year         = {2026},
  month        = jan,
  url          = {https://www.kimi.com/blog/kimi-k2-5},
}

@misc{qwen2026qwen35,
  title        = {{Qwen3.5}: Towards Native Multimodal Agents},
  author       = {{Qwen Team}},
  year         = {2026},
  month        = feb,
  url          = {https://qwen.ai/blog?id=qwen3.5},
}

@misc{zhipu2026glm47,
  title        = {{GLM-4.7}},
  author       = {{Z.AI}},
  year         = {2025},
  month        = dec,
  url          = {https://z.ai/blog/glm-4.7},
}

@article{benjamini1995fdr,
 ISSN = {00359246},
 url = {http://www.jstor.org/stable/2346101},
 author = {Benjamini, Yoav and Hochberg, Yosef},
 journal = {Journal of the Royal Statistical Society. Series B (Methodological)},
 number = {1},
 pages = {289--300},
 publisher = {[Royal Statistical Society, Oxford University Press]},
 title = {Controlling the False Discovery Rate: A Practical and Powerful Approach to Multiple Testing},
 urldate = {2026-07-19},
 volume = {57},
 year = {1995}
}

@misc{kim2026sciencescalingagentsystems,
      title={Towards a Science of Scaling Agent Systems}, 
      author={Yubin Kim and Ken Gu and Chanwoo Park and Chunjong Park and Samuel Schmidgall and A. Ali Heydari and Yao Yan and Zhihan Zhang and Yuchen Zhuang and Yun Liu and Mark Malhotra and Paul Pu Liang and Hae Won Park and Yuzhe Yang and Xuhai Xu and Yilun Du and Shwetak Patel and Tim Althoff and Daniel McDuff and Xin Liu},
      year={2026},
      eprint={2512.08296},
      archivePrefix={arXiv},
      primaryClass={cs.AI},
      url={https://arxiv.org/abs/2512.08296}, 
}

@misc{englander_agents_2026,
	title = {Agents {Explore} but {Agents} {Ignore}: {LLMs} {Lack} {Environmental} {Curiosity}},
	shorttitle = {Agents {Explore} but {Agents} {Ignore}},
	url = {https://arxiv.org/abs/2604.17609v1},
	language = {en},
	urldate = {2026-07-03},
	journal = {arXiv.org},
	author = {Engländer, Leon and Althammer, Sophia and Üstün, Ahmet and Gallé, Matthias and Sherborne, Tom},
	month = apr,
	year = {2026},
    doi = {10.48550/arXiv.2604.17609}
}

@misc{zhang_comparing_2025,
	title = {Comparing {Exploration}-{Exploitation} {Strategies} of {LLMs} and {Humans}: {Insights} from {Standard} {Multi}-armed {Bandit} {Experiments}},
	shorttitle = {Comparing {Exploration}-{Exploitation} {Strategies} of {LLMs} and {Humans}},
	url = {https://arxiv.org/abs/2505.09901v3},
	language = {en},
	urldate = {2026-07-03},
	journal = {arXiv.org},
    doi = {10.48550/arXiv.2505.09901},
	author = {Zhang, Ziyuan and Wang, Darcy and Chen, Ningyuan and Mansur, Rodrigo and Sarhangian, Vahid},
	month = may,
	year = {2025},
}

@article{mcdonald_bayesian_2025,
	title = {Bayesian {Optimization} over {Multiple} {Experimental} {Fidelities} {Accelerates} {Automated} {Discovery} of {Drug} {Molecules}},
	volume = {11},
	issn = {2374-7943},
	url = {https://doi.org/10.1021/acscentsci.4c01991},
	doi = {10.1021/acscentsci.4c01991},
	number = {2},
	urldate = {2026-07-03},
	journal = {ACS Central Science},
	publisher = {American Chemical Society},
	author = {McDonald, Matthew A. and Koscher, Brent A. and Canty, Richard B. and Zhang, Jason and Ning, Angelina and Jensen, Klavs F.},
	month = feb,
	year = {2025},
	pages = {346--356},
}

@inproceedings{
    dang_preferential_2025,
    title={Preferential Multi-Objective Bayesian Optimization for Drug Discovery},
    author={Tai Dang and Long-Hung Pham and Sang T. Truong and Ari Glenn and Wendy Nguyen and Edward A Pham and Jeffrey S. Glenn and Sanmi Koyejo and Thang Luong},
    booktitle={ICLR 2025 Workshop on Generative and Experimental Perspectives for Biomolecular Design},
    year={2025},
    url={https://openreview.net/forum?id=z9mxjhz8Nb}
}

@article{dave_autonomous_2022,
	title = {Autonomous optimization of non-aqueous {Li}-ion battery electrolytes via robotic experimentation and machine learning coupling},
	volume = {13},
	copyright = {2022 The Author(s)},
	issn = {2041-1723},
	url = {https://www.nature.com/articles/s41467-022-32938-1},
	doi = {10.1038/s41467-022-32938-1},
	language = {en},
	number = {1},
	urldate = {2026-07-03},
	journal = {Nature Communications},
	publisher = {Nature Publishing Group},
	author = {Dave, Adarsh and Mitchell, Jared and Burke, Sven and Lin, Hongyi and Whitacre, Jay and Viswanathan, Venkatasubramanian},
	month = sep,
	year = {2022},
	pages = {5454},
}

@article{jenewein_navigating_2024,
	title = {Navigating the unknown with {AI}: multiobjective {Bayesian} optimization of non-noble acidic {OER} catalysts},
	volume = {12},
	issn = {2050-7488},
	shorttitle = {Navigating the unknown with {AI}},
	url = {https://doi.org/10.1039/d3ta06651g},
	doi = {10.1039/d3ta06651g},
	number = {5},
	urldate = {2026-07-03},
	journal = {Journal of Materials Chemistry A},
	author = {Jenewein, Ken J. and Torresi, Luca and Haghmoradi, Navid and Kormányos, Attila and Friederich, Pascal and Cherevko, Serhiy},
	month = feb,
	year = {2024},
	pages = {3072--3083},
}

@article{ghorbani_active_2024,
	title = {An active machine learning approach for optimal design of magnesium alloys using {Bayesian} optimisation},
	volume = {14},
	copyright = {2024 The Author(s)},
	issn = {2045-2322},
	url = {https://www.nature.com/articles/s41598-024-59100-9},
	doi = {10.1038/s41598-024-59100-9},
	language = {en},
	number = {1},
	urldate = {2026-07-03},
	journal = {Scientific Reports},
	publisher = {Nature Publishing Group},
	author = {Ghorbani, M. and Boley, M. and Nakashima, P. N. H. and Birbilis, N.},
	month = apr,
	year = {2024},
	pages = {8299},
}

@article{khan_toward_2023,
	title = {Toward real-world automated antibody design with combinatorial {Bayesian} optimization},
	volume = {3},
	issn = {2667-2375},
	url = {https://www.cell.com/cell-reports-methods/abstract/S2667-2375(22)00276-4},
	doi = {10.1016/j.crmeth.2022.100374},
	language = {English},
	number = {1},
	urldate = {2026-07-03},
	journal = {Cell Reports Methods},
	publisher = {Elsevier},
	author = {Khan, Asif and Cowen-Rivers, Alexander I. and Grosnit, Antoine and Deik, Derrick-Goh-Xin and Robert, Philippe A. and Greiff, Victor and Smorodina, Eva and Rawat, Puneet and Akbar, Rahmad and Dreczkowski, Kamil and Tutunov, Rasul and Bou-Ammar, Dany and Wang, Jun and Storkey, Amos and Bou-Ammar, Haitham},
	month = jan,
	year = {2023},
}

@inproceedings{stanton_accelerating_2022,
	title = {Accelerating {Bayesian} {Optimization} for {Biological} {Sequence} {Design} with {Denoising} {Autoencoders}},
	issn = {2640-3498},
	url = {https://proceedings.mlr.press/v162/stanton22a.html},
	language = {en},
	urldate = {2026-07-03},
	booktitle = {Proceedings of the 39th {International} {Conference} on {Machine} {Learning}},
	publisher = {PMLR},
	author = {Stanton, Samuel and Maddox, Wesley and Gruver, Nate and Maffettone, Phillip and Delaney, Emily and Greenside, Peyton and Wilson, Andrew Gordon},
	month = jun,
	year = {2022},
	pages = {20459--20478},
}

@inproceedings{yang_large_2024,
     author = {Yang, Chengrun and Wang, Xuezhi and Lu, Yifeng and Liu, Hanxiao and Le, Quoc V and Zhou, Denny and Chen, Xinyun},
     booktitle = {International Conference on Learning Representations},
     editor = {B. Kim and Y. Yue and S. Chaudhuri and K. Fragkiadaki and M. Khan and Y. Sun},
     pages = {12028--12068},
     title = {Large Language Models as Optimizers},
     url = {https://proceedings.iclr.cc/paper_files/paper/2024/file/3339f19c5fcee3ad74502947a32be9e6-Paper-Conference.pdf},
     volume = {2024},
     year = {2024}
}

@article{schwaller2024chemistryllm,
    author ={Jablonka, Kevin M. and Schwaller, Philippe and  Ortega-Guerrero, Andres and Smit, Berend},
    title = {Leveraging large language models for predictive chemistry},
    journal = {Nature Machine Intelligence},
    pages = {161–169},
    volume = {6},
    year = {2024},
    month = {2},
    doi = {https://doi.org/10.1038/s42256-023-00788-1},
    url = {https://www.nature.com/articles/s42256-023-00788-1}
}

@article{rankovic2025largelanguagemodelsuncertaintycalibrated,
  author    = {Rankovi{\'c}, Bojana and Griffiths, Ryan-Rhys and Schwaller, Philippe},
  title     = {Large Language Models as Uncertainty-Calibrated Optimizers for Experimental Discovery},
  journal   = {Nature Machine Intelligence},
  year      = {2026},
  volume    = {8},
  number    = {9},
  pages     = {1466--1477},
  doi       = {10.1038/s42256-026-01283-z},
  eprint    = {2504.06265},
  archivePrefix = {arXiv}
}

@article{reinhart_large_2024,
	title = {Large language models design sequence-defined macromolecules via evolutionary optimization},
	volume = {10},
    author = {Reinhart, Wesley F. and Statt, Antonia},
	issn = {2057-3960},
	url = {https://www.nature.com/articles/s41524-024-01449-6},
	doi = {10.1038/s41524-024-01449-6},
	language = {en},
	number = {1},
	urldate = {2026-07-03},
	journal = {npj Computational Materials},
	month = {11},
	year = {2024},
	pages = {262},
}

@misc{noauthor_introducing_nodate,
    author = {Anthropic},
	title = {Introducing {Claude} 3.5 {Sonnet}},
	url = {https://www.anthropic.com/news/claude-3-5-sonnet},
	language = {en},
    year = {2024},
    month = {6},
	urldate = {2026-07-03},
}

@inproceedings{liu_large_2024,
     author = {Liu, Tennison and Astorga, Nicol\'{a}s and Seedat, Nabeel and van der Schaar, Mihaela},
     booktitle = {International Conference on Learning Representations},
     editor = {B. Kim and Y. Yue and S. Chaudhuri and K. Fragkiadaki and M. Khan and Y. Sun},
     pages = {31252--31284},
     title = {Large Language Models to Enhance Bayesian Optimization},
     url = {https://proceedings.iclr.cc/paper_files/paper/2024/file/84b8d9fcb4e262fcd429544697e1e720-Paper-Conference.pdf},
     volume = {2024},
     year = {2024}
}

@article{lu_generative_2025,
	title = {Generative {Design} of {Functional} {Metal} {Complexes} {Utilizing} the {Internal} {Knowledge} and {Reasoning} {Capability} of {Large} {Language} {Models}},
	volume = {147},
	issn = {0002-7863},
	url = {https://doi.org/10.1021/jacs.5c02097},
	doi = {10.1021/jacs.5c02097},
	number = {36},
	urldate = {2026-07-03},
	journal = {Journal of the American Chemical Society},
	publisher = {American Chemical Society},
	author = {Lu, Jieyu and Song, Zhangde and Zhao, Qiyuan and Du, Yuanqi and Cao, Yirui and Jia, Haojun and Duan, Chenru},
	month = sep,
	year = {2025},
	pages = {32377--32388},
}

@InProceedings{kristiadi_sober_2024,
  title = 	 {A Sober Look at {LLM}s for Material Discovery: Are They Actually Good for {B}ayesian Optimization Over Molecules?},
  author =       {Kristiadi, Agustinus and Strieth-Kalthoff, Felix and Skreta, Marta and Poupart, Pascal and Aspuru-Guzik, Alan and Pleiss, Geoff},
  booktitle = 	 {Proceedings of the 41st International Conference on Machine Learning},
  pages = 	 {25603--25622},
  year = 	 {2024},
  editor = 	 {Salakhutdinov, Ruslan and Kolter, Zico and Heller, Katherine and Weller, Adrian and Oliver, Nuria and Scarlett, Jonathan and Berkenkamp, Felix},
  volume = 	 {235},
  series = 	 {Proceedings of Machine Learning Research},
  month = 	 {21--27 Jul},
  publisher =    {PMLR},
  url = 	 {https://proceedings.mlr.press/v235/kristiadi24a.html}
}

@misc{wang_molecular_2024,
  title  = {Molecular Active Learning: How Can LLMs Help?},
  author = {Wang, Yuanqing and Xu, Yuzhi and Martiniani, Stefano and Karaletsos, Theofanis and Wilson, Andrew Gordon and Cho, Kyunghyun},
  note   = {Withdrawn submission (ICLR 2025)},
  year   = {2025},
  url    = {https://openreview.net/forum?id=kYg04pmX7i}
}

@article{sun_synllama_2025,
    author = {Sun, Kunyang and Bagni, Dorian and Cavanagh, Joseph M. and Wang, Yingze and Sawyer, Jacob M. and Zhou, Bo and Gritsevskiy, Andrew and Zhang, Oufan and Head-Gordon, Teresa},
    title = {SynLlama: Generating Synthesizable Molecules and Their Analogs with Large Language Models},
    journal = {ACS Central Science},
    volume = {11},
    number = {11},
    pages = {2108-2120},
    year = {2025},
    doi = {10.1021/acscentsci.5c01285},
    url = { 
            https://doi.org/10.1021/acscentsci.5c01285
    }
}

@misc{chaves_tx-llm_2024,
	title = {Tx-{LLM}: {A} {Large} {Language} {Model} for {Therapeutics}},
	shorttitle = {Tx-{LLM}},
	url = {http://arxiv.org/abs/2406.06316},
	doi = {10.48550/arXiv.2406.06316},
	urldate = {2026-07-03},
	publisher = {arXiv},
	author = {Chaves, Juan Manuel Zambrano and Wang, Eric and Tu, Tao and Vaishnav, Eeshit Dhaval and Lee, Byron and Mahdavi, S. Sara and Semturs, Christopher and Fleet, David and Natarajan, Vivek and Azizi, Shekoofeh},
	month = jun,
	year = {2024},
	note = {arXiv:2406.06316 [cs.CL]},
}

@article{thomas_test-time_2025,
    author = {Thomas, Morgan and Bou, Albert and De Fabritiis, Gianni},
    title = {Test-Time Training Scaling Laws for Chemical Exploration in Drug Design},
    journal = {Journal of Chemical Information and Modeling},
    volume = {65},
    number = {24},
    pages = {13178-13186},
    year = {2025},
    doi = {10.1021/acs.jcim.5c02316},
        note ={PMID: 41363014},
    url = { 
            https://doi.org/10.1021/acs.jcim.5c02316
    }
}

@article{
    sumers_cognitive_2024,
    title={Cognitive Architectures for Language Agents},
    author={Theodore Sumers and Shunyu Yao and Karthik R Narasimhan and Thomas L. Griffiths},
    journal={Transactions on Machine Learning Research},
    issn={2835-8856},
    year={2024},
    url={https://openreview.net/forum?id=1i6ZCvflQJ},
    note={Survey Certification, Featured Certification}
}

@misc{ghafarollahi_sparks_2025,
	title = {Sparks: {Multi}-{Agent} {Artificial} {Intelligence} {Model} {Discovers} {Protein} {Design} {Principles}},
	shorttitle = {Sparks},
	url = {http://arxiv.org/abs/2504.19017},
	doi = {10.48550/arXiv.2504.19017},
	urldate = {2026-07-03},
	publisher = {arXiv},
	author = {Ghafarollahi, Alireza and Buehler, Markus J.},
	month = apr,
	year = {2025},
	note = {arXiv:2504.19017 [cs.AI]},
}

@misc{ghareeb_robin_2025,
	title = {Robin: {A} multi-agent system for automating scientific discovery},
	shorttitle = {Robin},
	url = {http://arxiv.org/abs/2505.13400},
	doi = {10.48550/arXiv.2505.13400},
	urldate = {2026-07-03},
	publisher = {arXiv},
	author = {Ghareeb, Ali Essam and Chang, Benjamin and Mitchener, Ludovico and Yiu, Angela and Szostkiewicz, Caralyn J. and Laurent, Jon M. and Razzak, Muhammed T. and White, Andrew D. and Hinks, Michaela M. and Rodriques, Samuel G.},
	month = may,
	year = {2025},
	note = {arXiv:2505.13400 [cs.AI]},
}

@article{ghafarollahi_automating_2025,
	title = {Automating alloy design and discovery with physics-aware multimodal multiagent {AI}},
	volume = {122},
	url = {https://www.pnas.org/doi/10.1073/pnas.2414074122},
	doi = {10.1073/pnas.2414074122},
	number = {4},
	urldate = {2026-07-03},
	journal = {Proceedings of the National Academy of Sciences},
	publisher = {Proceedings of the National Academy of Sciences},
	author = {Ghafarollahi, Alireza and Buehler, Markus J.},
	month = jan,
	year = {2025},
	pages = {e2414074122},
}

@inproceedings{gupta_llms_2025,
    title = {LLMs for Bayesian Optimization in Scientific Domains: Are We There Yet?},
    author = {Gupta, Rushil  and Hartford, Jason  and Liu, Bang},
    editor = {Christodoulopoulos, Christos  and Chakraborty, Tanmoy  and Rose, Carolyn  and Peng, Violet},
    booktitle = {Findings of the Association for Computational Linguistics: EMNLP 2025},
    month = {11},
    year = {2025},
    publisher = {Association for Computational Linguistics},
    url = {https://aclanthology.org/2025.findings-emnlp.838/},
    doi = {10.18653/v1/2025.findings-emnlp.838},
    pages = {15482--15510},
    ISBN = {979-8-89176-335-7},
}

@inproceedings{he_martingale_2025,
    title={Martingale Score: An Unsupervised Metric for Bayesian Rationality in LLM Reasoning},
    author={He, Zhonghao and Qiu, Tianyi and Shirado, Hirokazu and Sap, Maarten},
    booktitle={The Thirty-ninth Annual Conference on Neural Information Processing Systems},
    year={2025},
    url={https://openreview.net/forum?id=BfO6od6JD6}
}

@inproceedings{krishnamurthy_can_2024,
 author = {Krishnamurthy, Akshay and Harris, Keegan and Foster, Dylan J. and Zhang, Cyril and Slivkins, Aleksandrs},
 booktitle = {Advances in Neural Information Processing Systems},
 doi = {10.52202/079017-3818},
 editor = {A. Globerson and L. Mackey and D. Belgrave and A. Fan and U. Paquet and J. Tomczak and C. Zhang},
 pages = {120124--120158},
 publisher = {Curran Associates, Inc.},
 title = {Can large language models explore in-context?},
 url = {https://proceedings.neurips.cc/paper_files/paper/2024/file/d951f73c521d069fefbb73396df01424-Paper-Conference.pdf},
 volume = {37},
 year = {2024}
}

@InProceedings{nie_evolve_2025,
  title = 	 {{EVOL}v{E}: Evaluating and Optimizing {LLM}s For In-Context Exploration},
  author =       {Nie, Allen and Su, Yi and Chang, Bo and Lee, Jonathan and Chi, Ed H. and Le, Quoc V and Chen, Minmin},
  booktitle = 	 {Proceedings of the 42nd International Conference on Machine Learning},
  pages = 	 {46346--46376},
  year = 	 {2025},
  editor = 	 {Singh, Aarti and Fazel, Maryam and Hsu, Daniel and Lacoste-Julien, Simon and Berkenkamp, Felix and Maharaj, Tegan and Wagstaff, Kiri and Zhu, Jerry},
  volume = 	 {267},
  series = 	 {Proceedings of Machine Learning Research},
  month = 	 {13--19 Jul},
  publisher =    {PMLR},
  url = 	 {https://proceedings.mlr.press/v267/nie25b.html}
}

@misc{schmied_llms_2025,
	title = {{LLMs} are {Greedy} {Agents}: {Effects} of {RL} {Fine}-tuning on {Decision}-{Making} {Abilities}},
	shorttitle = {{LLMs} are {Greedy} {Agents}},
	url = {http://arxiv.org/abs/2504.16078},
	doi = {10.48550/arXiv.2504.16078},
	urldate = {2026-07-03},
	publisher = {arXiv},
	author = {Schmied, Thomas and Bornschein, Jörg and Grau-Moya, Jordi and Wulfmeier, Markus and Pascanu, Razvan},
	month = apr,
	year = {2025},
	note = {arXiv:2504.16078 [cs.LG]},
}

@misc{bini_behavioral_2026,
	title = {Behavioral {Economics} of {AI}: {LLM} {Biases} and {Corrections}},
	shorttitle = {Behavioral {Economics} of {AI}},
	url = {http://arxiv.org/abs/2602.09362},
	doi = {10.48550/arXiv.2602.09362},
	urldate = {2026-07-03},
	publisher = {arXiv},
	author = {Bini, Pietro and Cong, Lin William and Huang, Xing and Jin, Lawrence J.},
	month = feb,
	year = {2026},
	note = {arXiv:2602.09362 [econ.GN]},
}

\newpage
\appendix

\renewcommand{\thefigure}{SI-\arabic{figure}}
\setcounter{figure}{0}
\renewcommand{\thetable}{SI-\arabic{table}}
\setcounter{table}{0}
\renewcommand{\thesection}{SI-\arabic{section}}
\setcounter{section}{0}
\section{Validations}
\label{sec:belief-validation}

Here we present the validation of the beliefs derived using an LLM judge.

\subsection{Hypothesis analysis}
\label{sec:hypothesis}
We extracted hypotheses stated in the reasoning traces of the reasoning models using an LLM judge. We grouped the hypotheses into three tags: Exploit, Explore, and Avoid. There was also a fourth "other" tag that the judge could assign if the hypothesis was vague or did not fit squarely in either of the first options. For each hypothesis, the judge also extracted all elements that the hypothesis argued for/against. An exploit hypothesis could be: "F is a strong candidate", an explore could be "F is untested but could perhaps go well with W", an avoid would be "F is a generally poor candidate", and finally other-tagged hypotheses are of multiple valencies: "F is good when paired with W but not F and has an inverse correlation with G concentration". On average, each reasoning trace contained $\sim7$ hypotheses of clear valence and tag, of which $\sim 45\%$ were exploitative, $\sim40\%$ exploratory, and $\sim15\%$ avoiding (Table~\ref{tab:hypotheses-frac}). The fractions were constant across prompt modes, with limited differences between models. The probability of a model selecting a value supported by an exploitative hypothesis was greater than the probability of it selecting one supported by an exploratory hypothesis, and significantly smaller than the probability of selecting one it intended to avoid (Table~\ref{tab:hypothesis-select}). Again, the explore/exploit split is consistent across all models and prompt modes. 92\% (CI$_{95\%}[0.85, 0.96], N=11,081$) of all traces invoked exploration. Further, 94\% (CI$_{95\%}[0.90, 0.96]$) of all 26,920 exploration hypotheses referenced at least one value that was selected in the batch, and 78.2\% (CI$_{95\%}[0.763, 0.800]$) referenced at least one value selected that was itself not referenced by any exploit-hypothesis. The models certainly intend to, and act to, explore. 

This analysis comes with several limitations, the most significant of which is the marginalization of all values. The judge can list only values supported by the hypothesis with the same valence. We therefore miss coupling between values and more complicated interactions. We also increase noise significantly. It is common for a value to be registered as supported by both an exploit and an avoid hypothesis as the judge attempts to consolidate cross interactions. It is therefore possible that the selection frequencies around 10\% of values registered as "avoid" are an artificial floor. Relative comparisons still hold.

\begin{table}[ht]\centering
\caption{Fraction of hypotheses of different tags by model and prompt mode (exploit, explore, avoid). hyps/cycle: mean number of hypotheses per cycle. 95\% CI only. $N$ marks the total number of hypotheses extracted per model-prompt mode combination. We use only hypotheses from the first 5 cycles to highlight early-stage hypothesis dynamics.}
\label{tab:hypotheses-frac}
\small
\begin{tabular}{l l c c c c r}
\toprule
Model & Mode & exploit & explore & avoid & hyps/cycle & $N$ \\
\midrule
\multirow{3}{*}{\texttt{gpt-5-4-2026-03-05}} & default & [0.45, 0.49] & [0.33, 0.37] & [0.17, 0.20] & [6.7, 7.7] & 2{,}648 \\
 & alias & [0.44, 0.47] & [0.35, 0.38] & [0.16, 0.19] & [7.1, 7.8] & 2{,}764 \\
 & blind & [0.43, 0.46] & [0.36, 0.40] & [0.16, 0.19] & [6.9, 7.5] & 2{,}510 \\
\midrule
\multirow{3}{*}{\texttt{claude-sonnet-4-6}} & default & [0.46, 0.49] & [0.35, 0.38] & [0.15, 0.18] & [8.0, 8.8] & 3{,}188 \\
 & alias & [0.45, 0.49] & [0.38, 0.41] & [0.12, 0.15] & [7.5, 8.2] & 2{,}979 \\
 & blind & [0.43, 0.47] & [0.38, 0.42] & [0.14, 0.17] & [6.3, 7.0] & 2{,}399 \\
\midrule
\multirow{3}{*}{\texttt{moonshotai-kimi-k2-5}} & default & [0.41, 0.45] & [0.41, 0.46] & [0.13, 0.15] & [7.1, 8.0] & 2{,}494 \\
 & alias & [0.39, 0.43] & [0.41, 0.46] & [0.14, 0.17] & [7.4, 8.3] & 2{,}580 \\
 & blind & [0.40, 0.43] & [0.39, 0.44] & [0.15, 0.19] & [6.5, 7.3] & 1{,}922 \\
\midrule
\multirow{3}{*}{\texttt{qwen-qwen3-5-397b-a17b}} & default & [0.42, 0.45] & [0.41, 0.44] & [0.13, 0.15] & [7.2, 8.0] & 2{,}318 \\
 & alias & [0.41, 0.44] & [0.41, 0.44] & [0.14, 0.16] & [7.4, 8.1] & 2{,}362 \\
 & blind & [0.45, 0.48] & [0.34, 0.38] & [0.16, 0.19] & [6.8, 7.6] & 1{,}654 \\
\midrule
\multirow{3}{*}{\texttt{glm-4-7}} & default & [0.41, 0.46] & [0.40, 0.45] & [0.13, 0.16] & [6.6, 7.6] & 2{,}404 \\
 & alias & [0.38, 0.41] & [0.43, 0.48] & [0.14, 0.16] & [7.0, 7.8] & 2{,}538 \\
 & blind & [0.40, 0.45] & [0.40, 0.46] & [0.13, 0.17] & [5.7, 6.5] & 2{,}061 \\
\bottomrule
\end{tabular}
\end{table}

\begin{table}[ht]\centering
\caption{Fraction of values backed by hypotheses which are later selected in the batch. The hypotheses are divided into three tags: exploitative, exploratory, and avoiding. Relative differences between the three tags highlight the model's adherence to its own claims. 95\% CI shown, bootstrapped from all extracted hypotheses ($N$). We use only hypotheses from the first 5 cycles to highlight early-stage hypothesis dynamics.}
\label{tab:hypothesis-select}
\small
\begin{tabular}{l l c c c r}
\toprule
Model & Tag & default & alias & blind & $N$ \\
\midrule
\texttt{gpt-5-4-2026-03-05} & \texttt{exploit} & [0.81, 0.86] & [0.79, 0.84] & [0.82, 0.88] & 3{,}585 \\
 & \texttt{explore} & [0.68, 0.74] & [0.69, 0.75] & [0.68, 0.74] & 2{,}901 \\
 & \texttt{avoid} & [0.12, 0.18] & [0.15, 0.21] & [0.15, 0.22] & 1{,}436 \\
\midrule
\texttt{claude-sonnet-4-6} & \texttt{exploit} & [0.81, 0.86] & [0.86, 0.90] & [0.80, 0.87] & 3{,}989 \\
 & \texttt{explore} & [0.75, 0.81] & [0.74, 0.80] & [0.66, 0.73] & 3{,}281 \\
 & \texttt{avoid} & [0.12, 0.17] & [0.15, 0.23] & [0.10, 0.19] & 1{,}296 \\
\midrule
\texttt{moonshotai-kimi-k2-5} & \texttt{exploit} & [0.82, 0.88] & [0.78, 0.87] & [0.88, 0.93] & 2{,}924 \\
 & \texttt{explore} & [0.67, 0.74] & [0.68, 0.78] & [0.65, 0.72] & 3{,}002 \\
 & \texttt{avoid} & [0.11, 0.19] & [0.14, 0.25] & [0.11, 0.19] & 1{,}070 \\
\midrule
\texttt{qwen-qwen3-5-397b-a17b} & \texttt{exploit} & [0.82, 0.88] & [0.78, 0.85] & [0.87, 0.93] & 2{,}762 \\
 & \texttt{explore} & [0.71, 0.80] & [0.71, 0.79] & [0.69, 0.78] & 2{,}599 \\
 & \texttt{avoid} & [0.08, 0.15] & [0.16, 0.26] & [0.12, 0.23] & 973 \\
\midrule
\texttt{glm-4-7} & \texttt{exploit} & [0.82, 0.87] & [0.83, 0.89] & [0.86, 0.91] & 2{,}883 \\
 & \texttt{explore} & [0.82, 0.88] & [0.80, 0.87] & [0.78, 0.85] & 3{,}076 \\
 & \texttt{avoid} & [0.11, 0.19] & [0.11, 0.21] & [0.10, 0.19] & 1{,}044 \\
\bottomrule
\end{tabular}
\end{table}

\subsection{Belief judge ablation}
\label{sec:belief-ablation}

Table~\ref{tab:cross-judge} shows cross-model judge alignment on the direct arylation dataset, GPT-5.4, pooled across prompt modes. We dropped points with less than 0.02 belief weight, as large search spaces introduce significant noise in Spearman around 0. Agreement is excellent ($>0.8$ on both overall correlation and which-is-best). 

\begin{table}[ht]\centering
\caption{Cross-model judge agreement on the verbalized beliefs for GPT-5.4 on direct arylation across prompt modes, $n=7800$ points. Each gpt-5-mini-vs-judge pair is reported in two ways: the ranked agreement (Spearman $\rho$) and the overall agreement (Pearson $r$).}
\small
\begin{tabular}{llc}
\toprule
Judge pair & Agreement & value \\
\midrule
gpt-5-mini vs claude-haiku-4-5 & Spearman $\rho$ & 0.886 \\
gpt-5-mini vs claude-haiku-4-5 & Pearson $r$ & 0.941 \\
gpt-5-mini vs qwen3-8b & Spearman $\rho$ & 0.827 \\
gpt-5-mini vs qwen3-8b & Pearson $r$ & 0.894 \\
\bottomrule
\end{tabular}\label{tab:cross-judge}
\end{table}

\subsection{Belief quality}
\label{sec:belief-quality}

To further validate the extracted beliefs, we correlate them with the model's actions at the respective steps, compute Brier, ECE, and AUROC scores against the oracle (the full true fitness landscape), and compute the Gini score. Only a modest correlation is expected, as the LLM is exploring only a small subset of the full landscape and thus cannot be expected to predict overall fitness. Table~\ref{tab:belief-quality} reports belief quality. All models tested perform similarly. There is strong agreement between stated beliefs and the model's actions ($\rho>0.5$), but the correlation with the oracle is modest (AUROC $>0.6$). Poor or no calibration with uncertainty ($\rho(\mathrm{Gini}, h_{\mathrm{data}}) \sim 0.1$) may reflect poor calibration of the judge, artifacts from marginalizing uncertainty, artifacts from the uncertainty estimation (ridge regression), or be yet another indicator of the LLM's inability to identify high-information areas of the search space discussed in the main text. Finally, ECE and Brier scores $>0.25$ indicate clear overconfidence in the model or judge. The same ECE and Brier values appear for all judges tested. However, we cannot determine whether this is a judge-wide issue in understanding hypotheses or the model overstating beliefs (likely both).

\begin{table}[ht]\centering 
\caption{Belief quality (AUROC/Brier/ECE vs. oracle) and Spearman correlations, per judged agent model. Bootstrap 95\% CI shown (N=10,000). AUROC significant if CI excludes $0.5$, correlations if CI excludes $0$.}
\label{tab:belief-quality}
\small
\begin{tabular}{lccccc}
\toprule
Agent model & AUROC & Brier & ECE & $\rho(\mathrm{Gini},h_{\mathrm{data}})$ & $\rho(b,P_{\mathrm{sel}})$ \\
\midrule
gpt-5 & $[0.586,\,0.607]$ & $[0.301,\,0.307]$ & $[0.284,\,0.292]$ & $[0.005,\,0.222]$ & $[0.730,\,0.775]$  \\
sonnet-4-6 & $[0.602,\,0.623]$ & $[0.302,\,0.308]$ & $[0.289,\,0.296]$ & $[0.009,\,0.214]$ & $[0.694,\,0.744]$  \\
kimi-k2 & $[0.611,\,0.626]$ & $[0.299,\,0.304]$ & $[0.284,\,0.291]$ & $[-0.013,\,0.194]$ & $[0.587,\,0.641]$ \\
qwen3.5-397b & $[0.609,\,0.625]$ & $[0.301,\,0.306]$ & $[0.285,\,0.292]$ & $[-0.008,\,0.201]$ & $[0.625,\,0.676]$  \\
glm-4-7 & $[0.616,\,0.631]$ & $[0.298,\,0.304]$ & $[0.286,\,0.293]$ & $[0.013,\,0.219]$ & $[0.699,\,0.741]$  \\
\bottomrule
\end{tabular}
\end{table}

We find a close correlation between the action belief and the stated beliefs (Pearson-$\rho=0.76$ CI$_{95\%}[+0.754, +0.764]$, Spearman-$\rho=0.55$ CI$_{95\%}[+0.529, +0.562], \ N= 13{,}484$ reasoning traces, 6 datasets, 5 models, 3 prompt modes). This strong correlation may be due in part to the judge's ability to see the final selections in the reasoning traces. However, the entropy is not collapsed onto only the selected candidates. For example, ``I should likely choose 0.1 or 0.153'' is realized as $b_{0.153} = 0.45$ even though 0.153 was never selected in that batch. In fact, in 35\% (CI$_{95\%}[+0.34, +0.36],\ N= 13{,}485$ batches) of all batches, a highest-belief value was never selected. Further, selections for which no hypothesis was extracted (18.8\% of selections, 1014 campaigns, 107{,}348 labels without hypothesis) consistently receive less probability mass (mean $0.137$, CI [0.130,0.145], $n=20{,}128$) than selections with assigned hypotheses (mean $0.323$, CI [0.317,0.329], $n=87{,}220$) meaning the judge is not naively assigning mass to the selected candidates but is effectively assessing the hypotheses.

\begin{figure}
    \centering
    \includegraphics[width=0.75\linewidth]{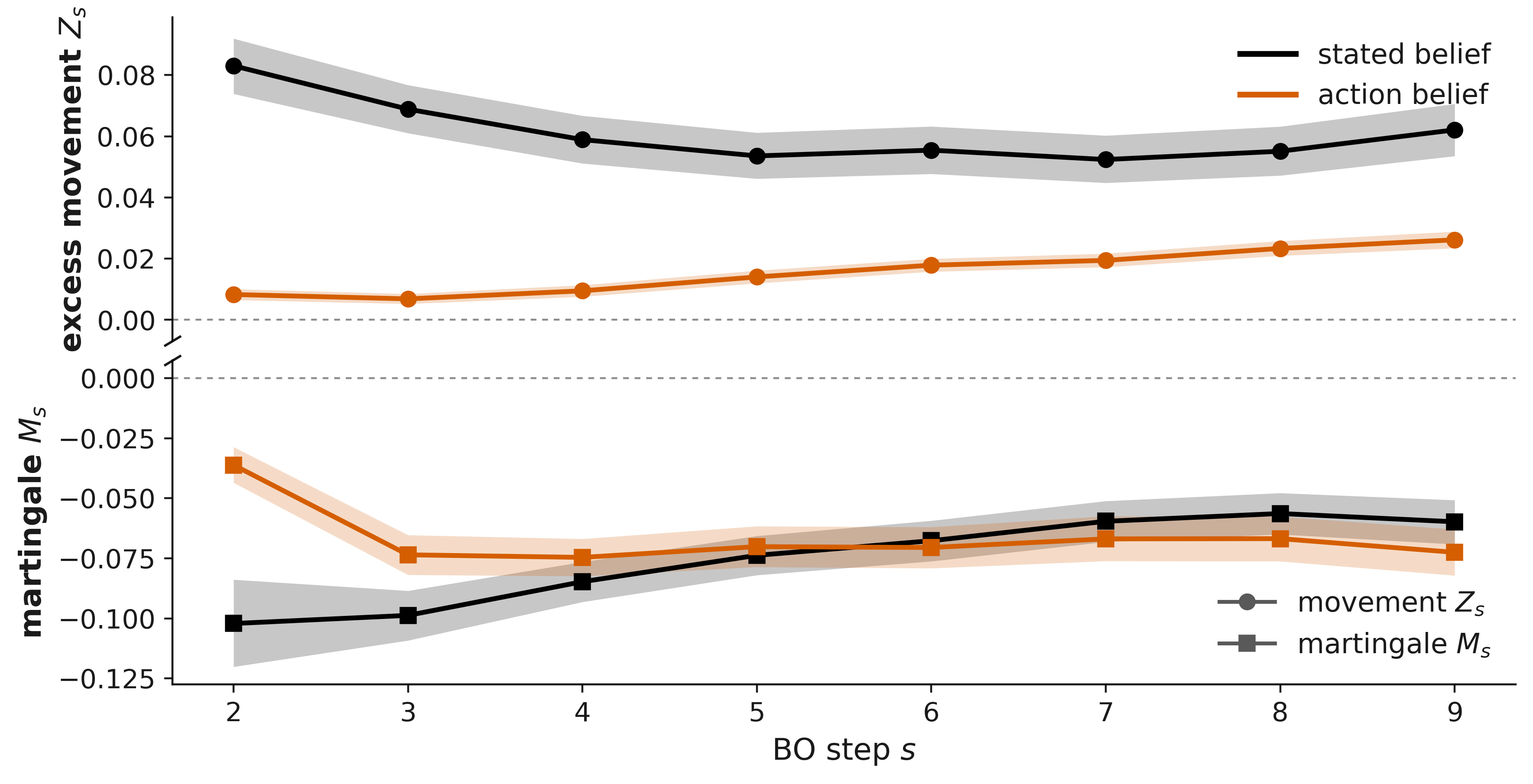}
    \caption{Martingale score (lower) and belief movement (upper) graph from the main paper, pooled across prompt modes (stated belief). Here, we have also derived the same metrics using beliefs estimated directly from actions (action beliefs). Action beliefs and stated beliefs yield the same measured overreaction. }
    \label{fig:si-martingale-action}
\end{figure}

Measuring Martingale score and belief movement using the action beliefs confirms the exact findings of using stated beliefs (Figure \ref{fig:si-martingale-action}).

\subsection{Exploration judge ablation}
\label{sec:exploration-ablation}
We validated the exploration intent judge in the same way as we did the belief judge. First, we performed a judge ablation against claude-haiku-4.5 and qwen3-8b on one dataset (direct arylation) and one model (GPT-5.4) across all prompt modes (Table~\ref{tab:exploration-ablation}). The verdicts of all three models were well correlated. We then correlated the intentions against the hypotheses extracted using a different prompt method (Section~\ref{sec:methods-hypotheses}). Higher exploration intent was indeed correlated with an increased number of exploratory hypotheses in the reasoning traces but not the number of exploitative hypotheses (Table~\ref{tab:exploration-crosscheck}). 

\begin{table}[ht]\centering
\caption{Cross-judge agreement on \texttt{explore\_fraction}, GPT-5.4 on direct arylation dataset, pooled over prompt modes, $N=447$ reasoning traces.}
\small
\label{tab:exploration-ablation}
\begin{tabular}{lcc}
\toprule
Judge pair & Spearman $\rho$ & Pearson $r$ \\
\midrule
gpt-5-mini vs claude-haiku-4-5 & 0.83 & 0.84 \\
gpt-5-mini vs qwen3-8b & 0.71 & 0.71 \\
claude-haiku-4-5 vs qwen3-8b & 0.75 & 0.72 \\
\bottomrule
\end{tabular}
\end{table}

\begin{table}[ht]\centering
\caption{Exploration judge (gpt-5-mini) against independently extracted hypotheses, which tags each stated hypothesis as \texttt{explore}/\texttt{exploit}/\texttt{avoid}. $N=448$ reasoning traces. Greater intent to explore should correlate with the number of exploratory hypotheses, but not exploitative ones.}
\label{tab:exploration-crosscheck}
\small
\begin{tabular}{lc}
\toprule
Target & Spearman $\rho$ [95\% CI] \\
\midrule
$n_\text{explore}$ & $+0.24$ $[+0.14, +0.33]$ \\
$n_\text{exploit}$  & $-0.19$ $[-0.27, -0.10]$  \\
$n_\text{total}$ & $-0.04$ $[-0.13, +0.05]$\\
\bottomrule
\end{tabular}
\end{table}
\section{Further results and failure modes}
\label{si:benchmark}
 
In this section, we delve deeper into the benchmark results to identify and characterize variance in the model and dataset. In particular, we describe a series of detrimental failure modes that the population-level analysis averages over. The shared failure mode (overreaction, stickiness) is treated in the main text and is
not revisited here.
 
\subsection{Per-dataset and per-model performance}
\label{si:per-dataset}
 
\begin{figure}[htbp]
  \centering
  \begin{subfigure}[b]{0.45\textwidth}\centering
    \includegraphics[width=\textwidth]{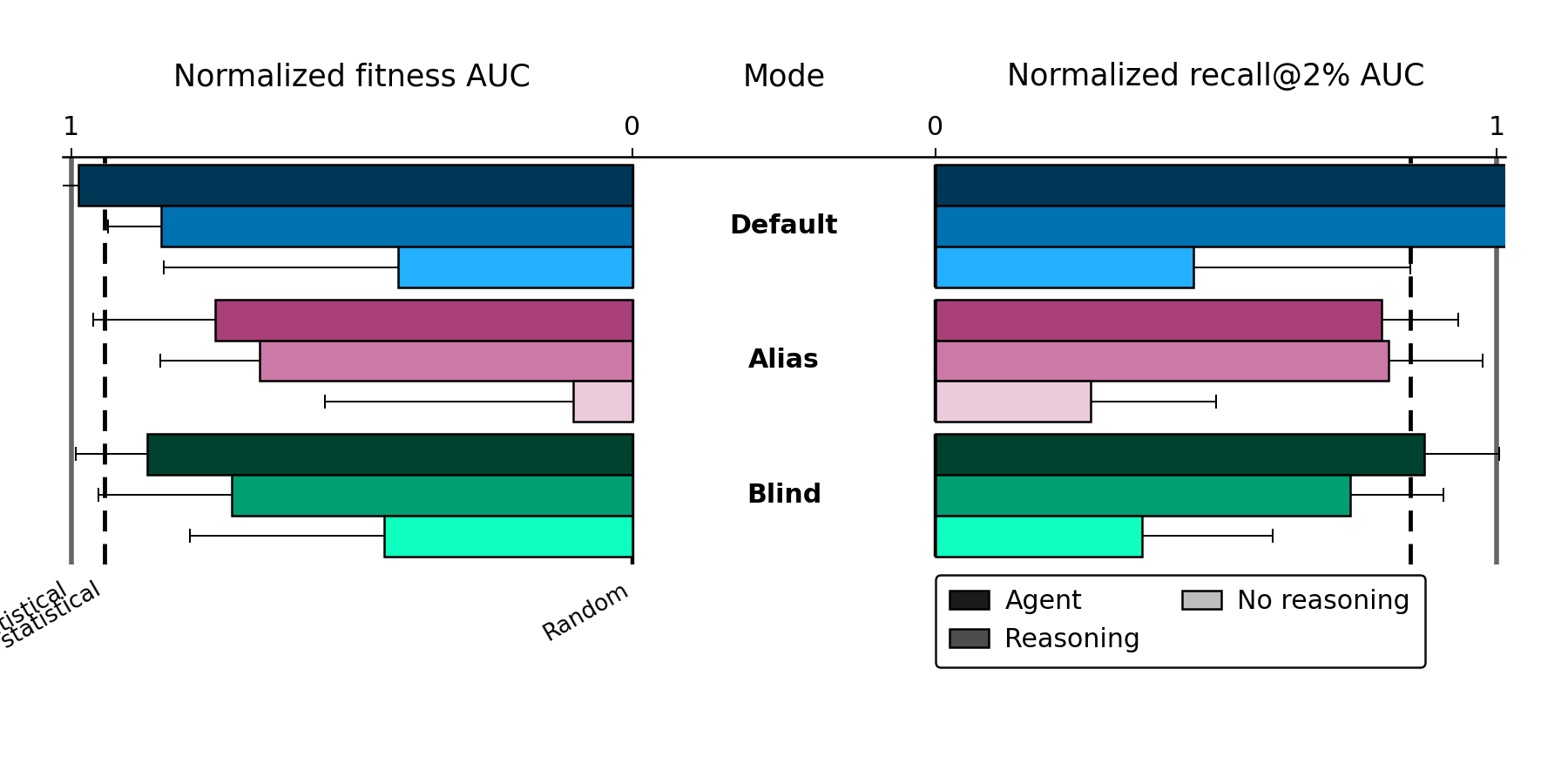}\caption{ALDE, GB1}\end{subfigure}\hfill
  \begin{subfigure}[b]{0.45\textwidth}\centering
    \includegraphics[width=\textwidth]{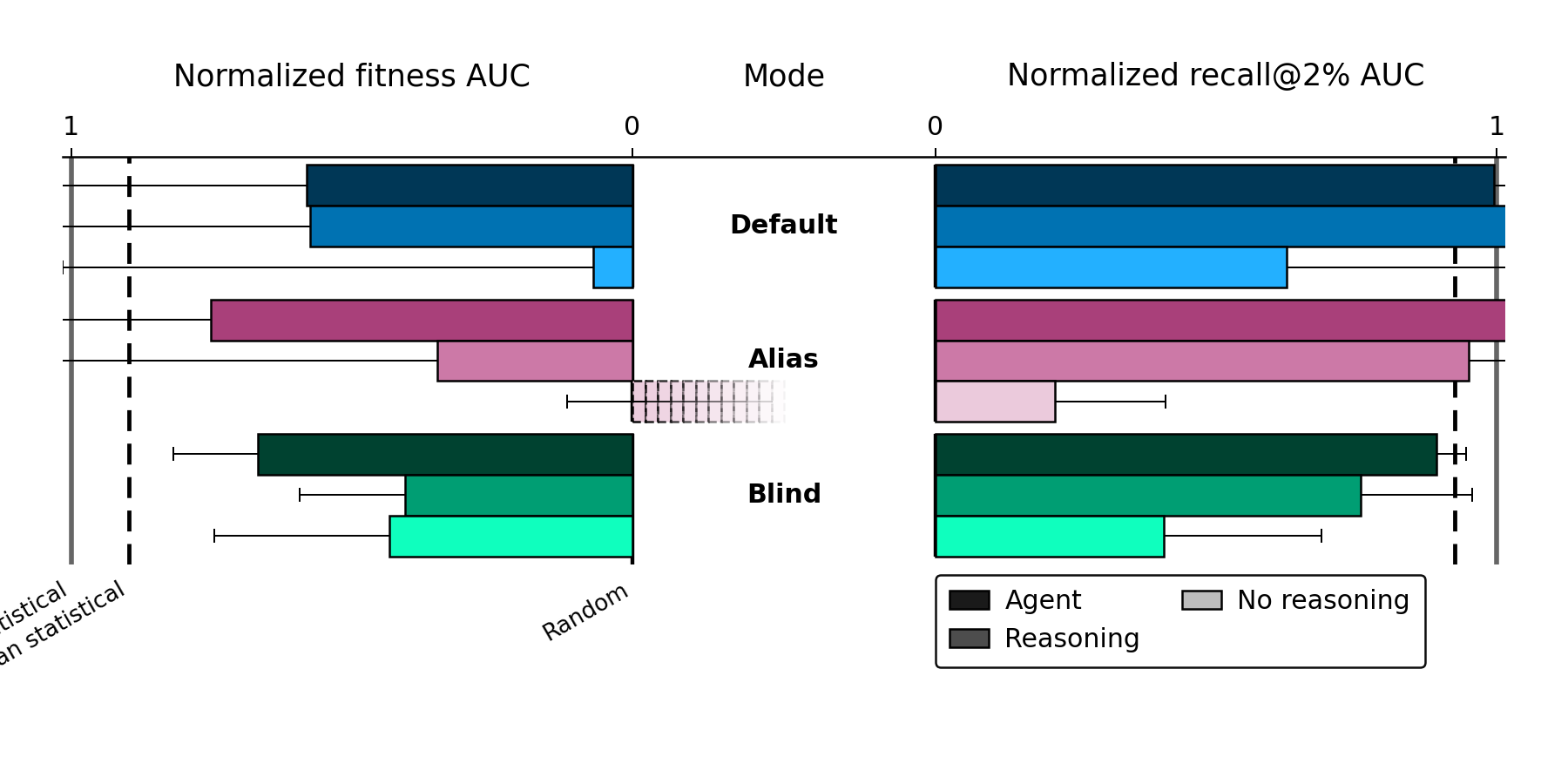}\caption{ALDE, TrpB}\end{subfigure}
  \vspace{0.4cm}
  \begin{subfigure}[b]{0.45\textwidth}\centering
    \includegraphics[width=\textwidth]{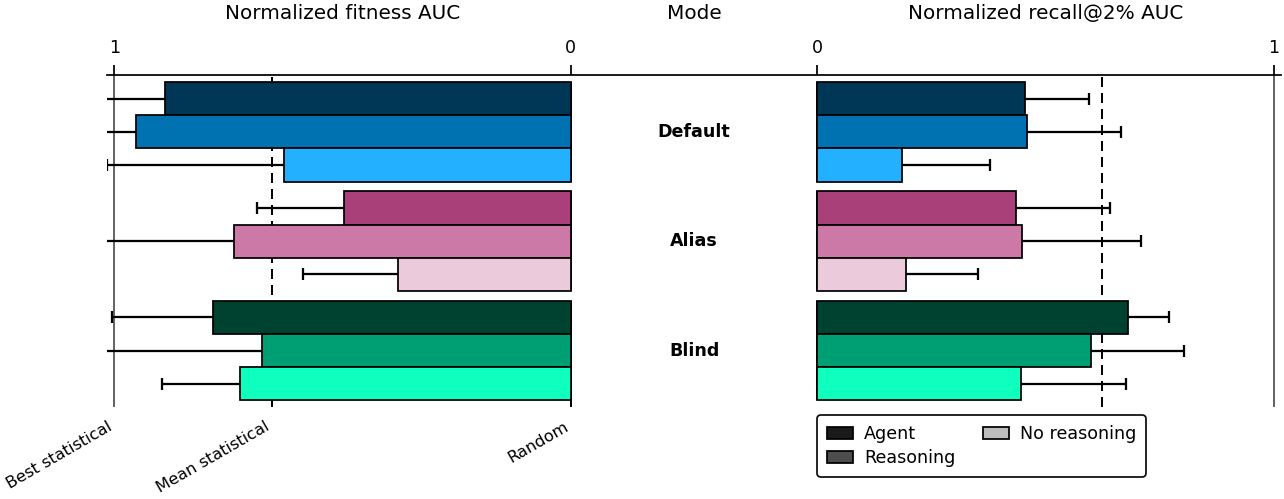}\caption{EDBO, direct arylation}\end{subfigure}\hfill
  \begin{subfigure}[b]{0.45\textwidth}\centering
    \includegraphics[width=\textwidth]{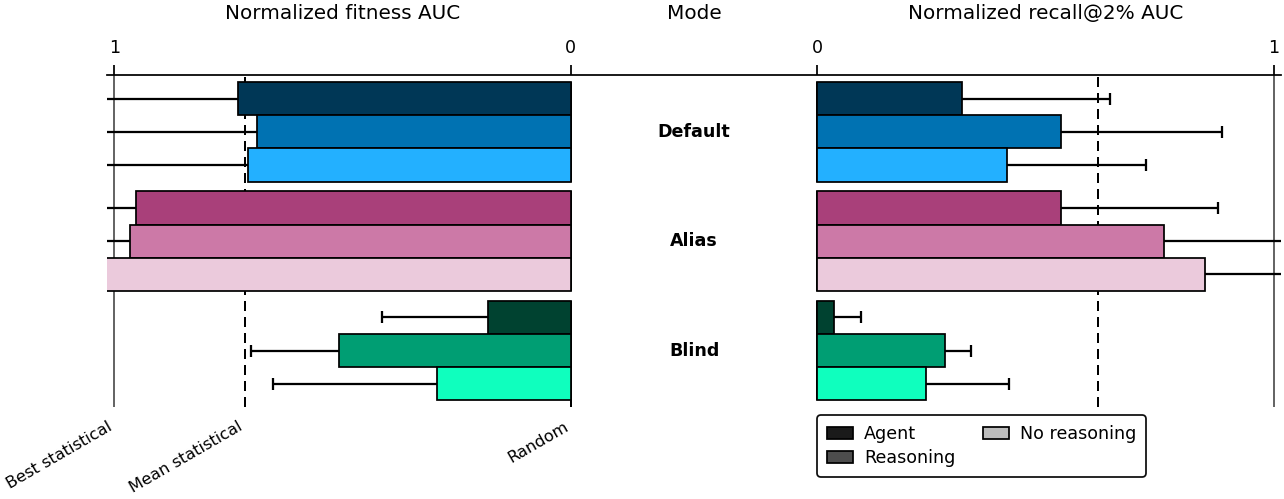}\caption{EDBO, aryl amination}\end{subfigure}
  \vspace{0.4cm}
  \begin{subfigure}[b]{0.45\textwidth}\centering
    \includegraphics[width=\textwidth]{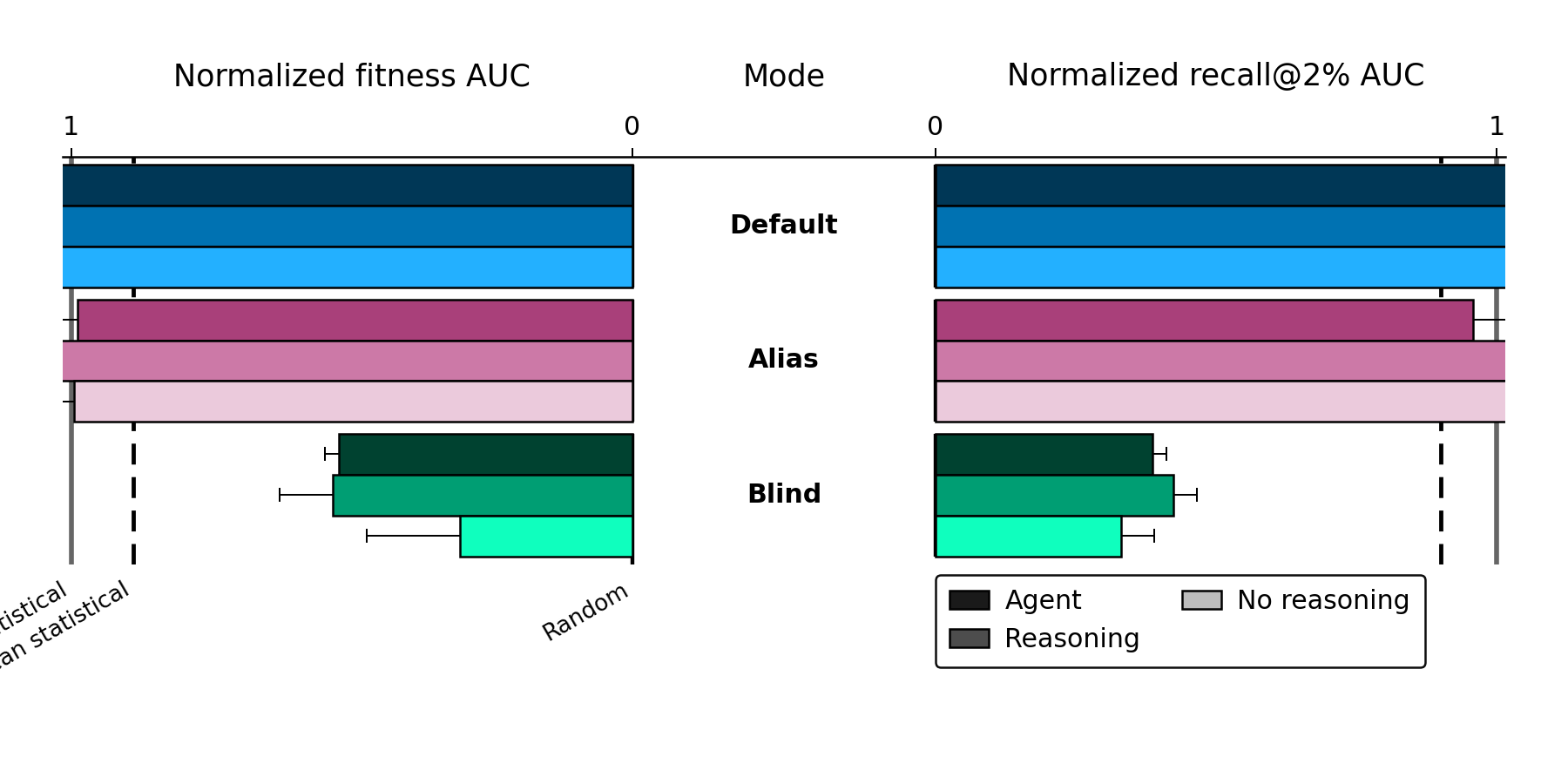}\caption{Tripeptides}\end{subfigure}\hfill
  \begin{subfigure}[b]{0.45\textwidth}\centering
    \includegraphics[width=\textwidth]{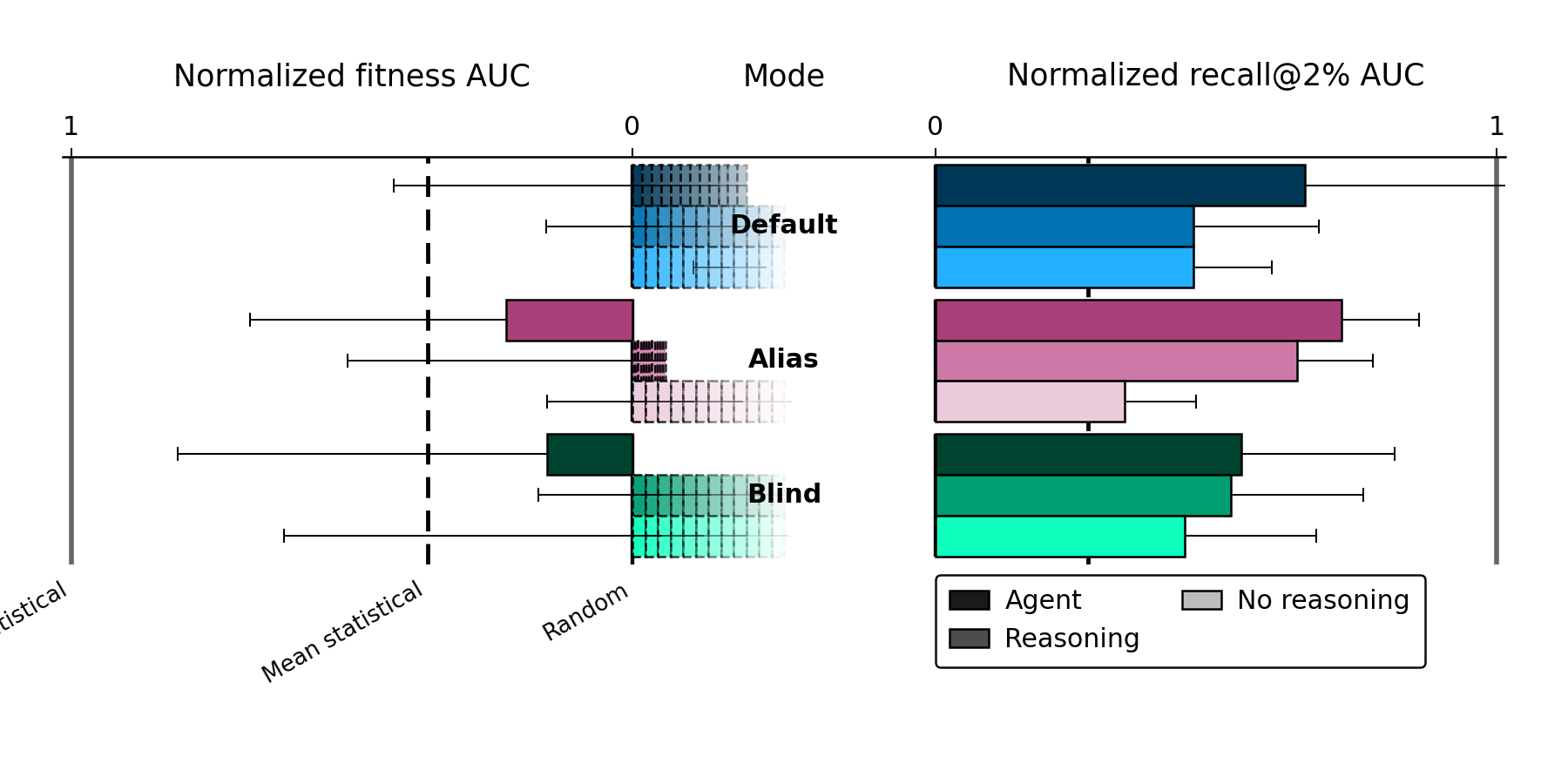}\caption{OER}\end{subfigure}
  \vspace{0.4cm}
  \begin{subfigure}[b]{0.45\textwidth}\centering
    \includegraphics[width=\textwidth]{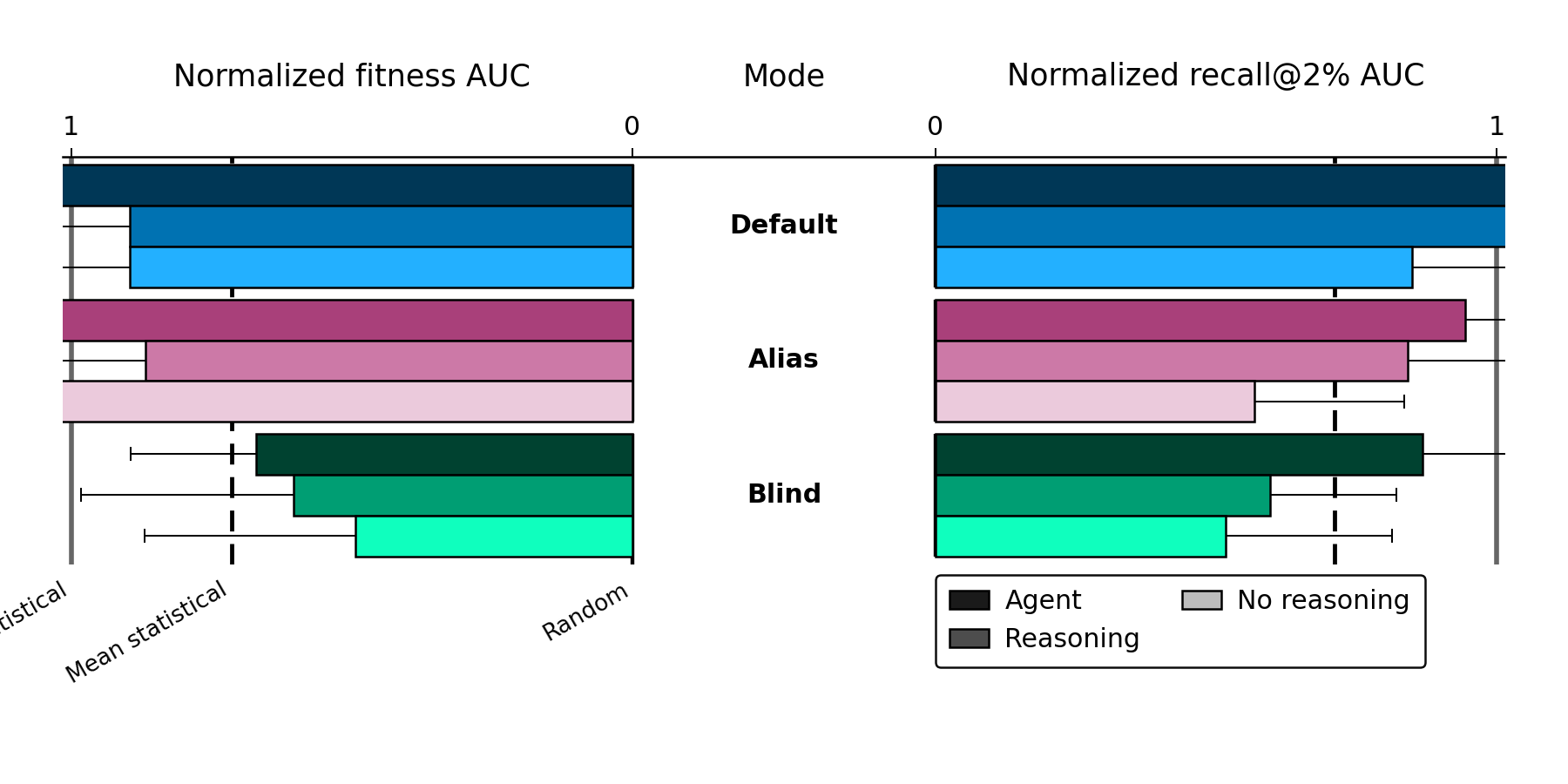}\caption{NFA}\end{subfigure}\hfill
  \begin{subfigure}[b]{0.45\textwidth}\centering
    \includegraphics[width=\textwidth]{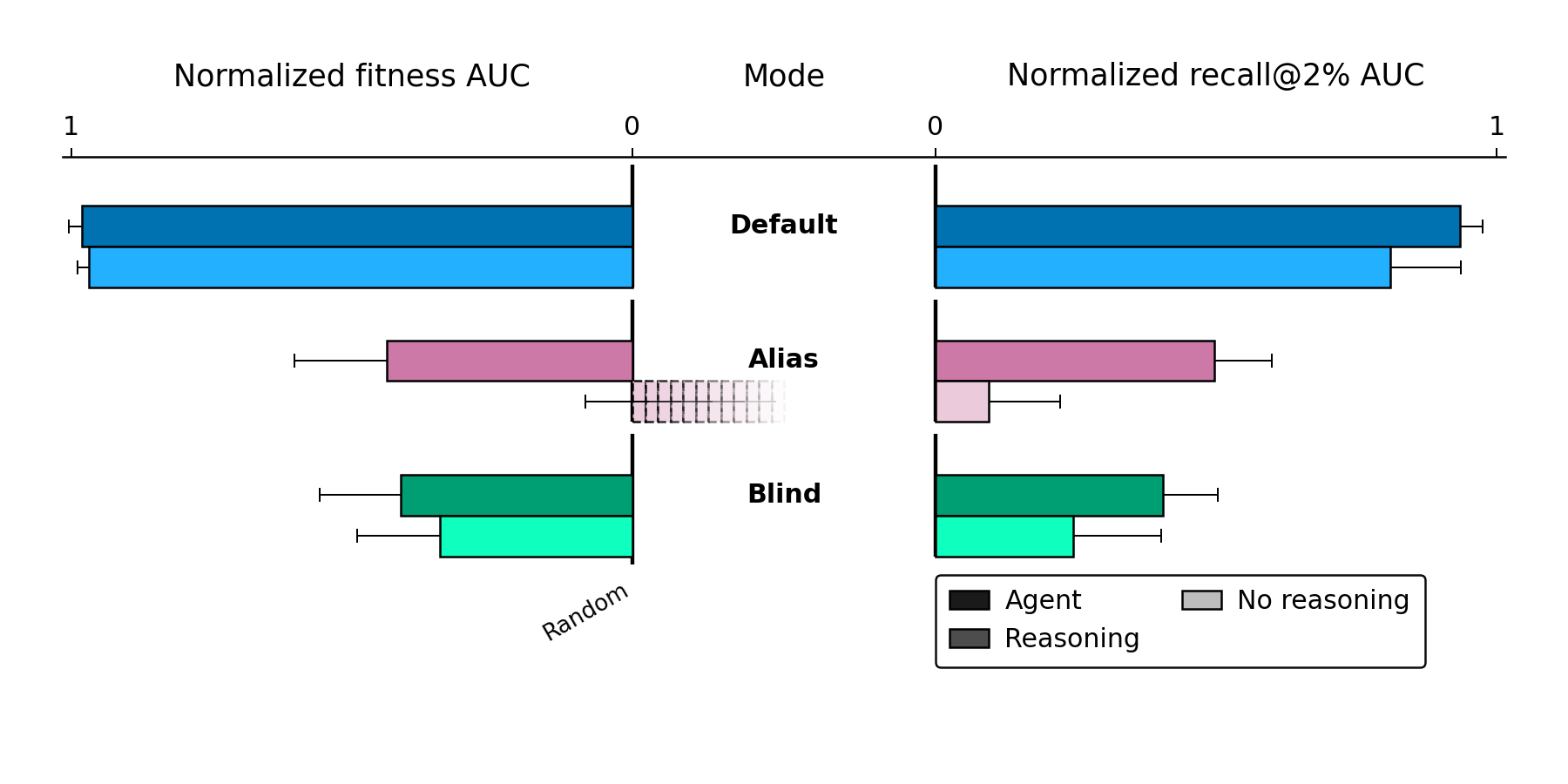}\caption{Perovskites}\end{subfigure}
  \caption{Per-dataset normalized AUC across the five models, three prompt modes,
  Moreover, reasoning ablation (max fitness and recall merged per panel). Performance is
  normalized per dataset against the best statistical model ($y=1$) and random
  selection ($y=0$). Error bars are standard deviation, dominated by between-model
  variance.}
  \label{fig:si-per-dataset}
\end{figure}

\begin{figure}
    \centering
    \includegraphics[width=0.95\linewidth]{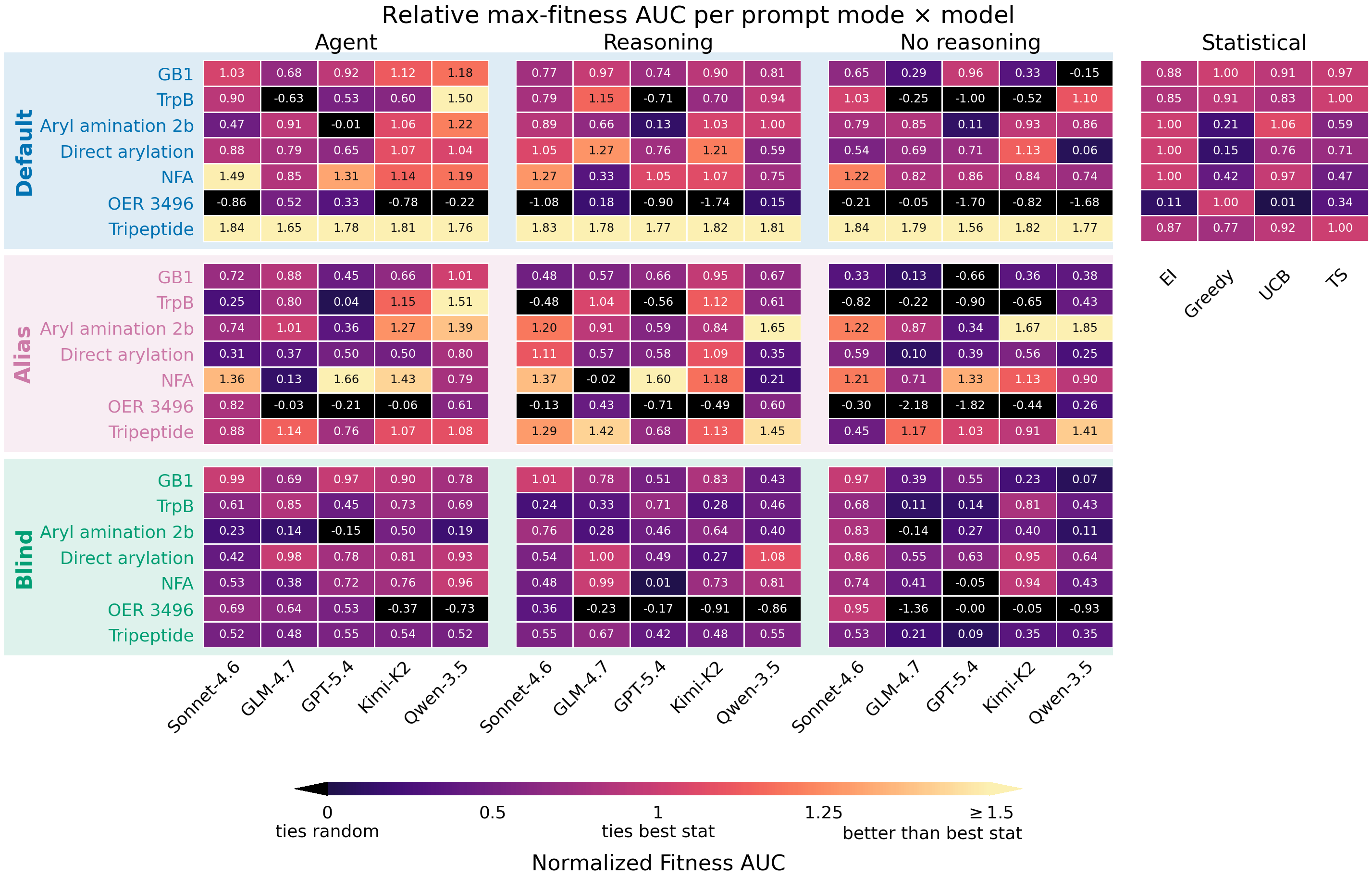}
    \caption{Heatmap with full breakdown of normalized AUC by best fitness for each dataset-model-prompt mode triplet. Yellow means better than the best statistical model; black means worse than random. The table to the right gives statistical models' relative performance as a baseline. Colors do not mark statistical significance. No model performs significantly better or worse than others. Note: The best statistical model is defined by a combination of fitness and recall performance, whereby non-"best" statistical models can be assigned normalized fitness > 1 when performance is close to a tie on one metric.}
    \label{fig:si-heatmap}
\end{figure}

Most variance in performance lies between seeds, then datasets, and then models. Figure~\ref{fig:si-per-dataset} shows the benchmark results per dataset, with error bars marking standard deviation. Figure~\ref{fig:si-heatmap} shows a heatmap over each prompt-mode$\times$LLM-model pair on each dataset, colored by relative AUC of max fitness. No single model is strictly better than any other. Moreover, priors often help but can occasionally lead to worse performance. All models are history sticky, exploring less than even greedy (Table \ref{tab:per-model-behaviour}), although GPT and Sonnet are significantly stickier than the other three models.

We removed the perovskites dataset from the main analysis after running only 10 seeds for reasoning and non-reasoning models. All models consistently suggested one of the top three candidates in the search space (196 candidates total) in the first batch in default mode. Removing the explicit reagent names and explicit target interaction while keeping chemical descriptors (alias) completely removed this behavior, and the models started to reason generally. We consider this a positive control of the detection mechanisms of the default-alias split. 
 
\begin{figure}[htbp]
  \centering
  \begin{subfigure}[b]{0.45\textwidth}\centering
    \includegraphics[width=\textwidth]{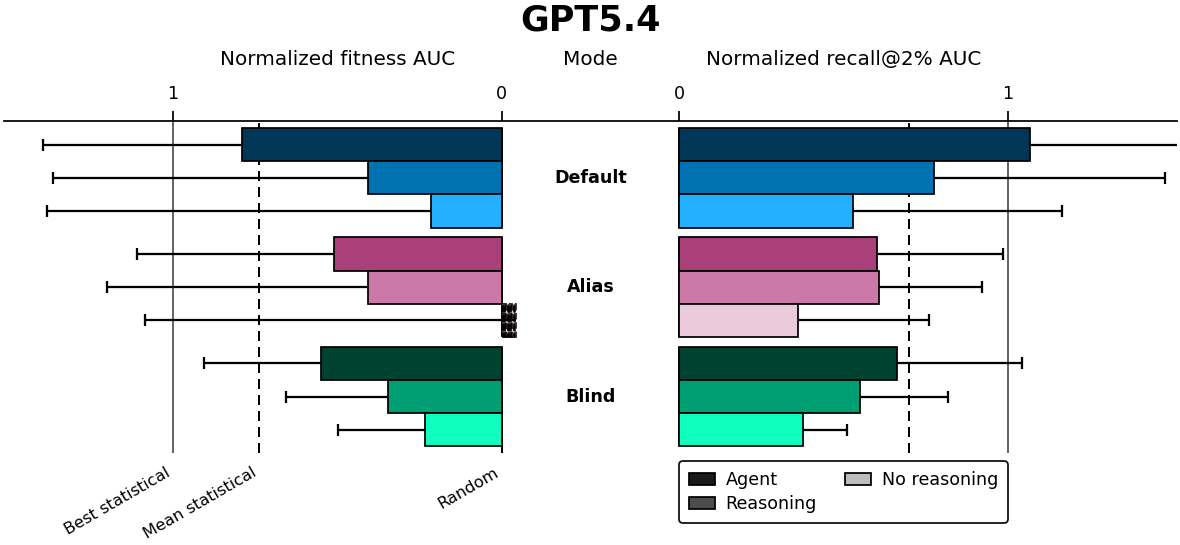}\caption{GPT-5.4}\end{subfigure}\hfill
  \vspace{0.4cm}
  \begin{subfigure}[b]{0.45\textwidth}\centering
    \includegraphics[width=\textwidth]{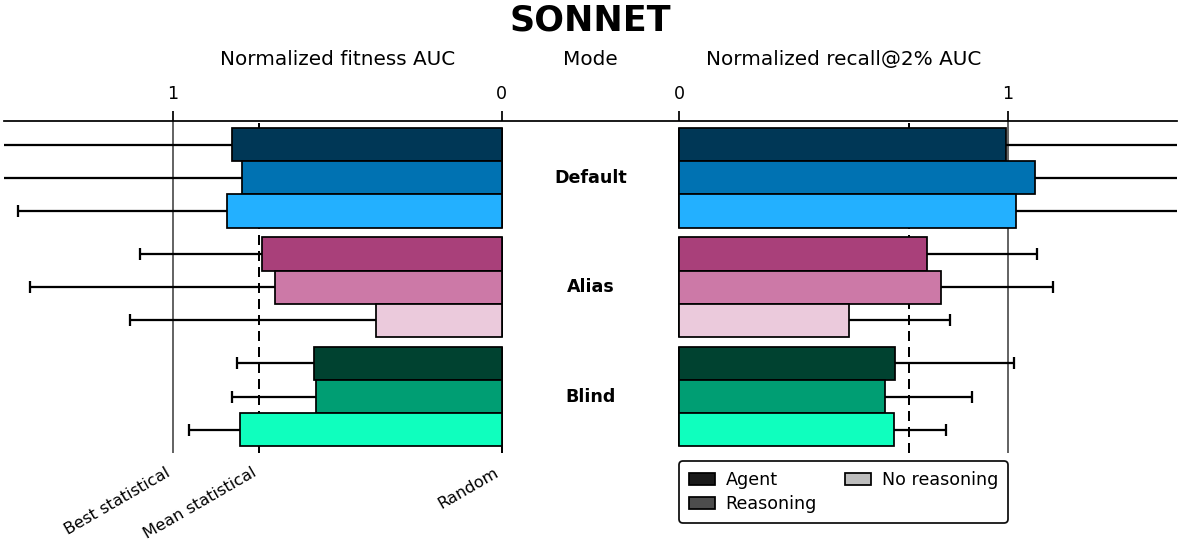}\caption{Sonnet-4.6}\end{subfigure}\hfill
  \begin{subfigure}[b]{0.45\textwidth}\centering
    \includegraphics[width=\textwidth]{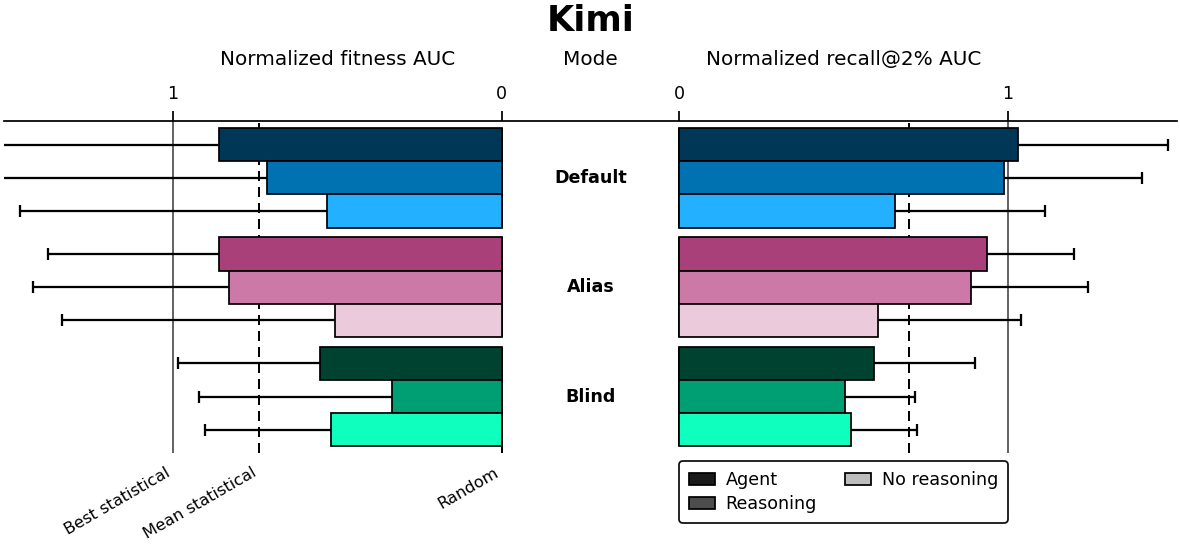}\caption{Kimi-2.5}\end{subfigure}
  \vspace{0.4cm}
  \begin{subfigure}[b]{0.45\textwidth}\centering
    \includegraphics[width=\textwidth]{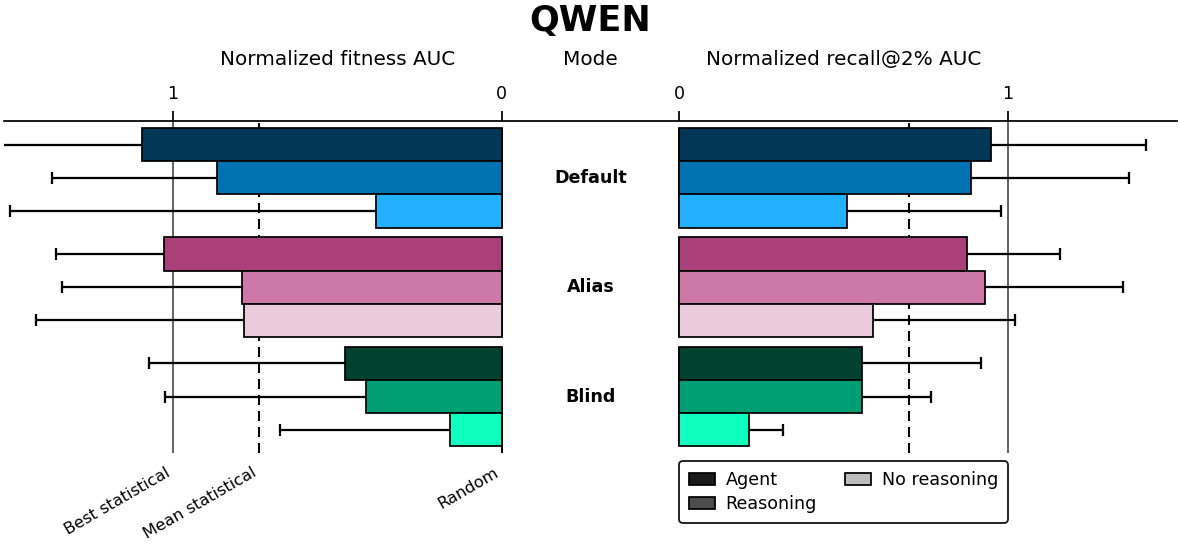}\caption{Qwen-3.5}\end{subfigure}\hfill
  \begin{subfigure}[b]{0.45\textwidth}\centering
    \includegraphics[width=\textwidth]{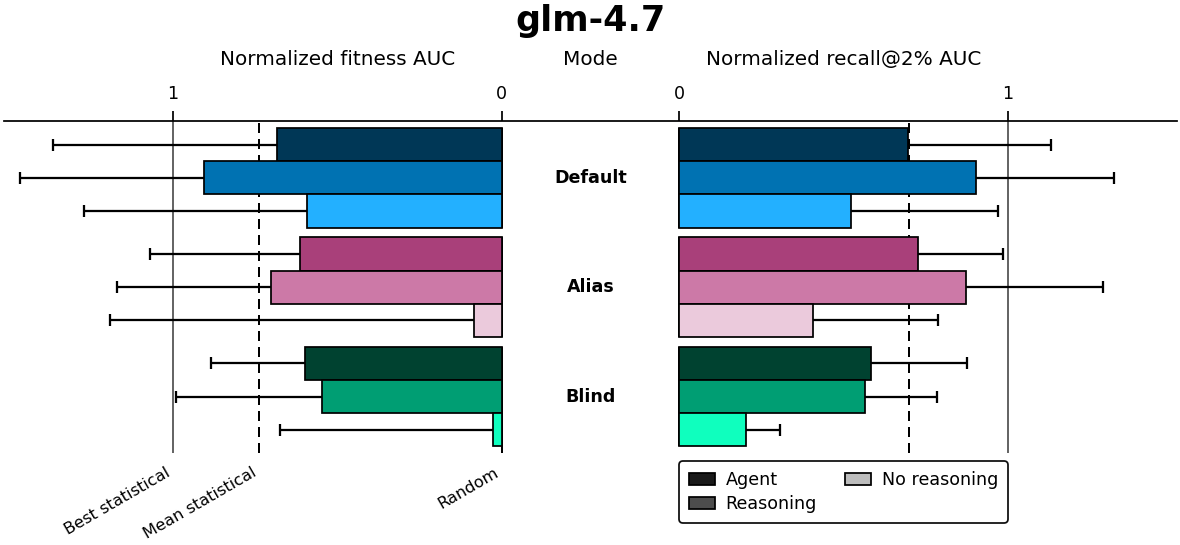}\caption{GLM-4.7}\end{subfigure}
  \caption{Per-model normalized AUC across 7 datasets. No single model dominates: each is strong on a different subset of datasets and modes.}
  \label{fig:si-per-model}
\end{figure}

 \begin{figure}
    \centering
    \includegraphics[width=0.9\linewidth]{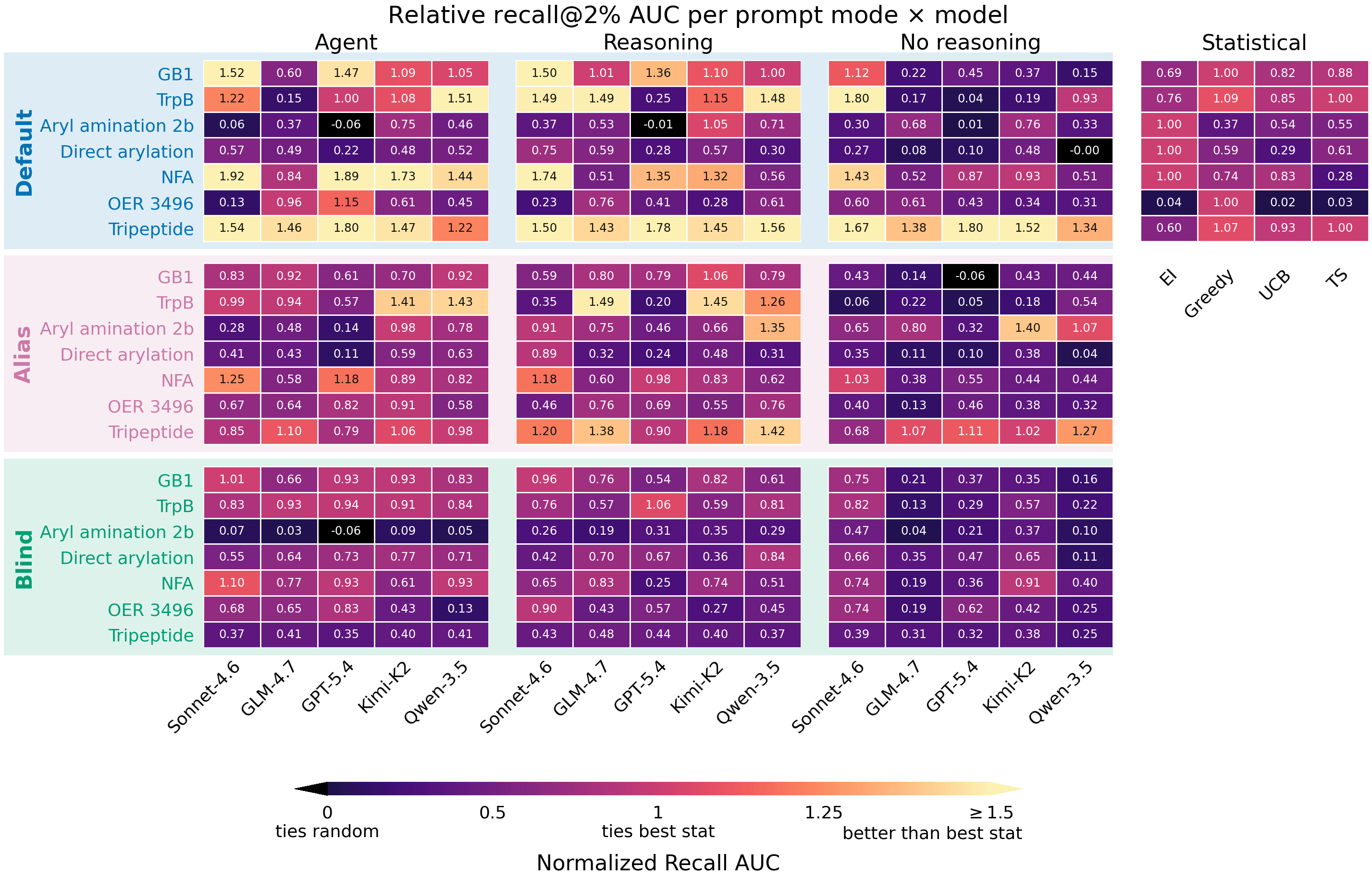}
    \caption{Heatmap with full breakdown of normalized AUC by recall (number of found high-value candidates) for each dataset-model-prompt mode triplet. Yellow means better than the best statistical model; black means worse than random. The table to the right gives statistical models' relative performance as a baseline. Colors do not mark statistical significance. No one model is performing significantly better or worse than others. Best statistical model is defined by a combination of fitness and recall performance whereby non-"best" statistical models can be assigned normalized fitness > 1 when performance is close to a tie on one metric.}
    \label{fig:si-heatmap-recall}
\end{figure}

\begin{table}[ht]\centering
\caption{Per-model behavioral statistics, pooled across 7 datasets, 3 prompt modes, and all seeds (reasoning). Exploration metrics are measured as excess over uniform sampling. Higher $U$, Novelty, Rarity, or Coverage are more exploratory. Brackets are 95\% CIs from a 10,000-resample campaign cluster bootstrap. 14,175 batches over 1,575 campaigns. \textsc{Greedy-GP} and \textsc{UCB-GP} are statistical acquisition functions run on the same datasets for reference. Exploration columns use 9,450 batches over 1,050 campaigns. $V$/$U$ here subsample 10 seeds/dataset.}
\label{tab:per-model-behaviour}
\small\setlength{\tabcolsep}{4pt}
\begin{tabular}{lccccc}
\toprule
Model & $V$ & $U$ & Novelty & Rarity & Coverage \\
\midrule
\texttt{gpt-5.4-2026-03-05} & [+2.18, +2.47] & [-3.39, -2.95] & [-0.320, -0.293] & [-0.272, -0.251] & [-0.111, -0.096] \\
\texttt{claude-sonnet-4-6} & [+2.34, +2.63] & [-3.36, -2.86] & [-0.314, -0.289] & [-0.266, -0.249] & [-0.087, -0.073] \\
\texttt{moonshotai/kimi-k2.5} & [+2.00, +2.24] & [-2.88, -2.48] & [-0.291, -0.266] & [-0.240, -0.222] & [-0.064, -0.054] \\
\texttt{qwen3.5-397b-a17b} & [+1.92, +2.15] & [-3.14, -2.69] & [-0.288, -0.263] & [-0.243, -0.225] & [-0.072, -0.061] \\
\texttt{glm-4.7} & [+1.89, +2.12] & [-2.97, -2.60] & [-0.284, -0.261] & [-0.238, -0.221] & [-0.070, -0.060] \\
\midrule
\textsc{Greedy-GP} & [+1.61, +2.44] & [-2.80, -1.93] & [-0.262, -0.242] & [-0.201, -0.189] & [-0.068, -0.063] \\
\textsc{UCB-GP} & [+1.73, +2.45] & [-0.40, +0.92] & [-0.167, -0.141] & [-0.132, -0.116] & [-0.034, -0.027] \\
\bottomrule
\end{tabular}
\end{table}

\subsection{Characterized failure modes}
\label{si:failures}

\begin{figure}[htbp]
  \centering
  \begin{subfigure}[b]{0.45\textwidth}\centering
    \includegraphics[width=\textwidth]{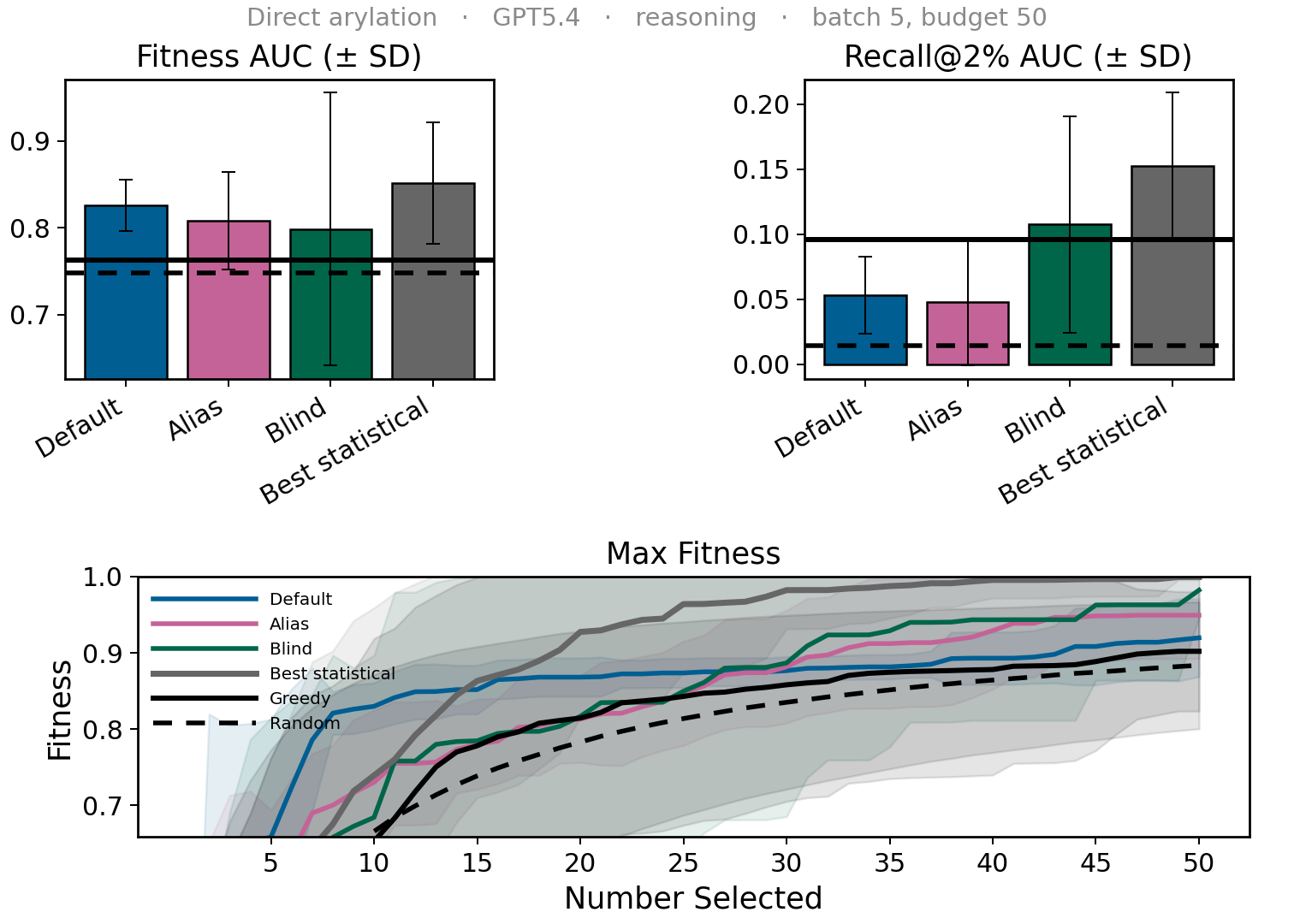}\caption{GPT-5.4}\end{subfigure}\hfill
  \begin{subfigure}[b]{0.45\textwidth}\centering
    \includegraphics[width=\textwidth]{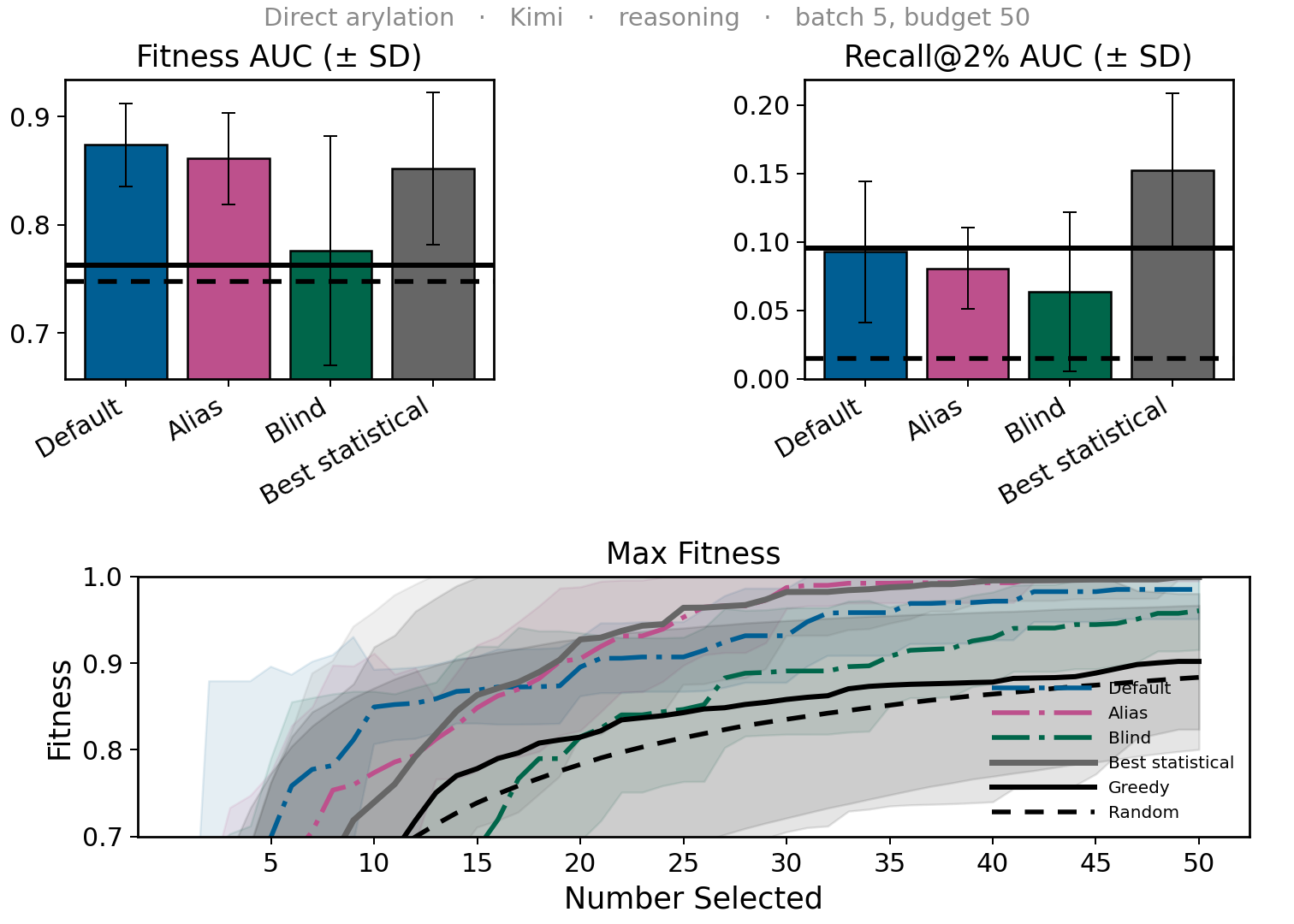}\caption{Kimi-2.5}\end{subfigure}
  \vspace{0.5cm}
  \begin{subfigure}[b]{0.45\textwidth}\centering
    \includegraphics[width=\textwidth]{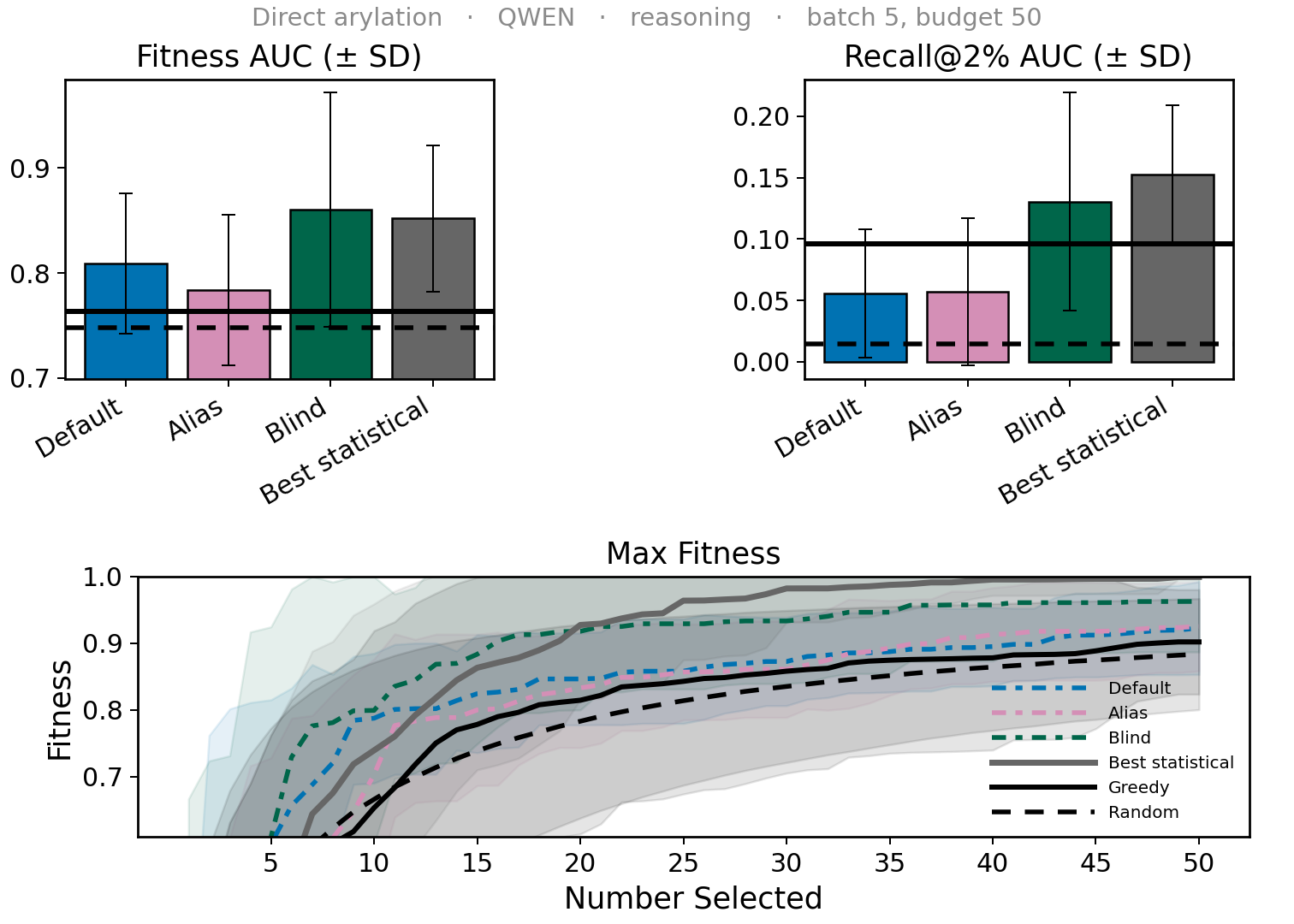}\caption{Qwen-3.5}\end{subfigure}\hfill
  \begin{subfigure}[b]{0.45\textwidth}\centering
    \includegraphics[width=\textwidth]{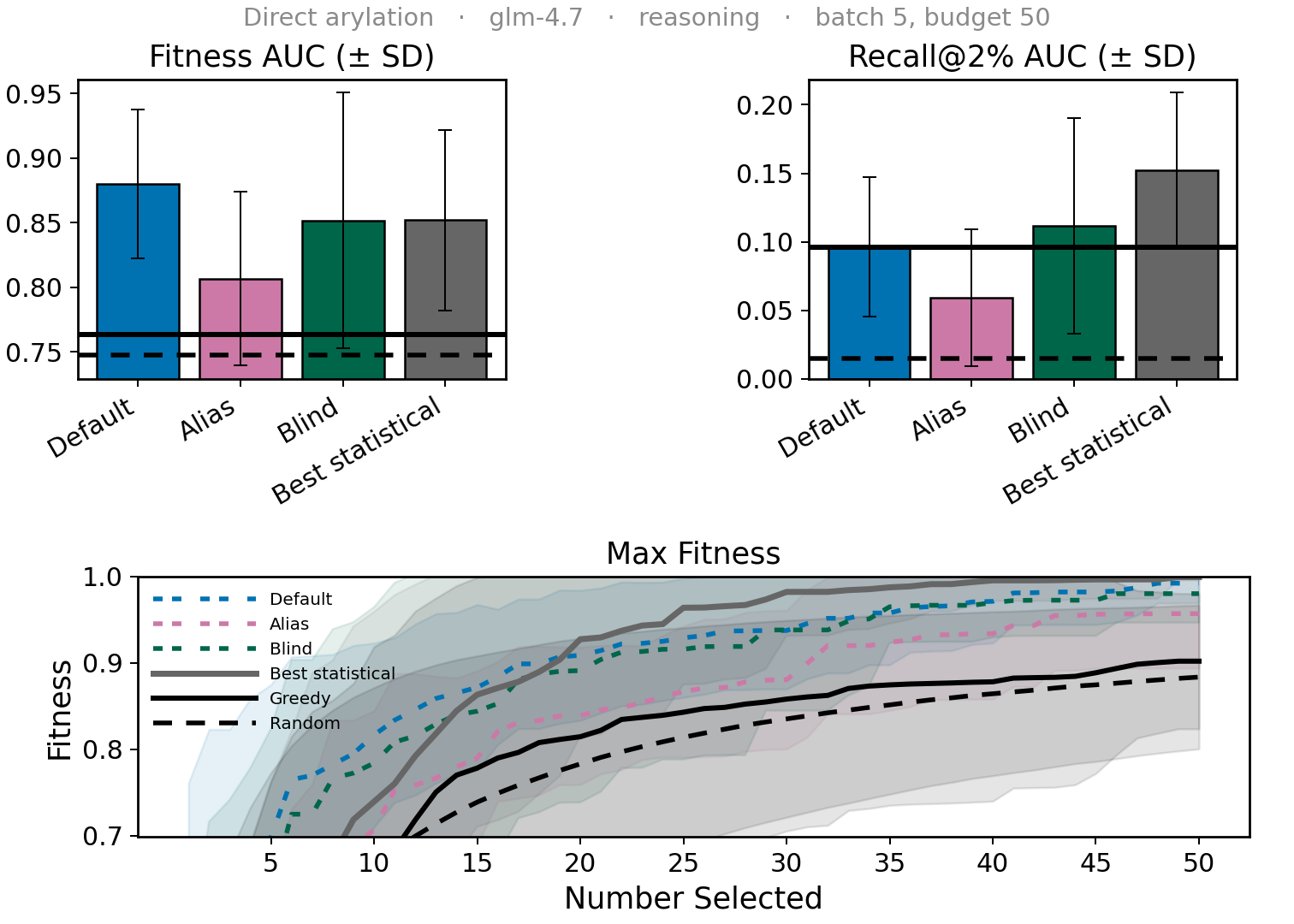}\caption{GLM-4.7}\end{subfigure}
  \caption{Summary statistics on the performance of GPT-5.4, Kimi-2.5, Qwen-3.5, and GLM on the direct arylation task. All models but GLM show some negative correlation between the default prompt mode and performance; the variance is large among models. Error bars are standard deviation. $N=15$ seeds.}
  \label{fig:direct-arylation-models}
\end{figure}

\begin{figure}[htbp]
  \centering
  \begin{subfigure}[b]{0.45\textwidth}\centering
    \includegraphics[width=\textwidth]{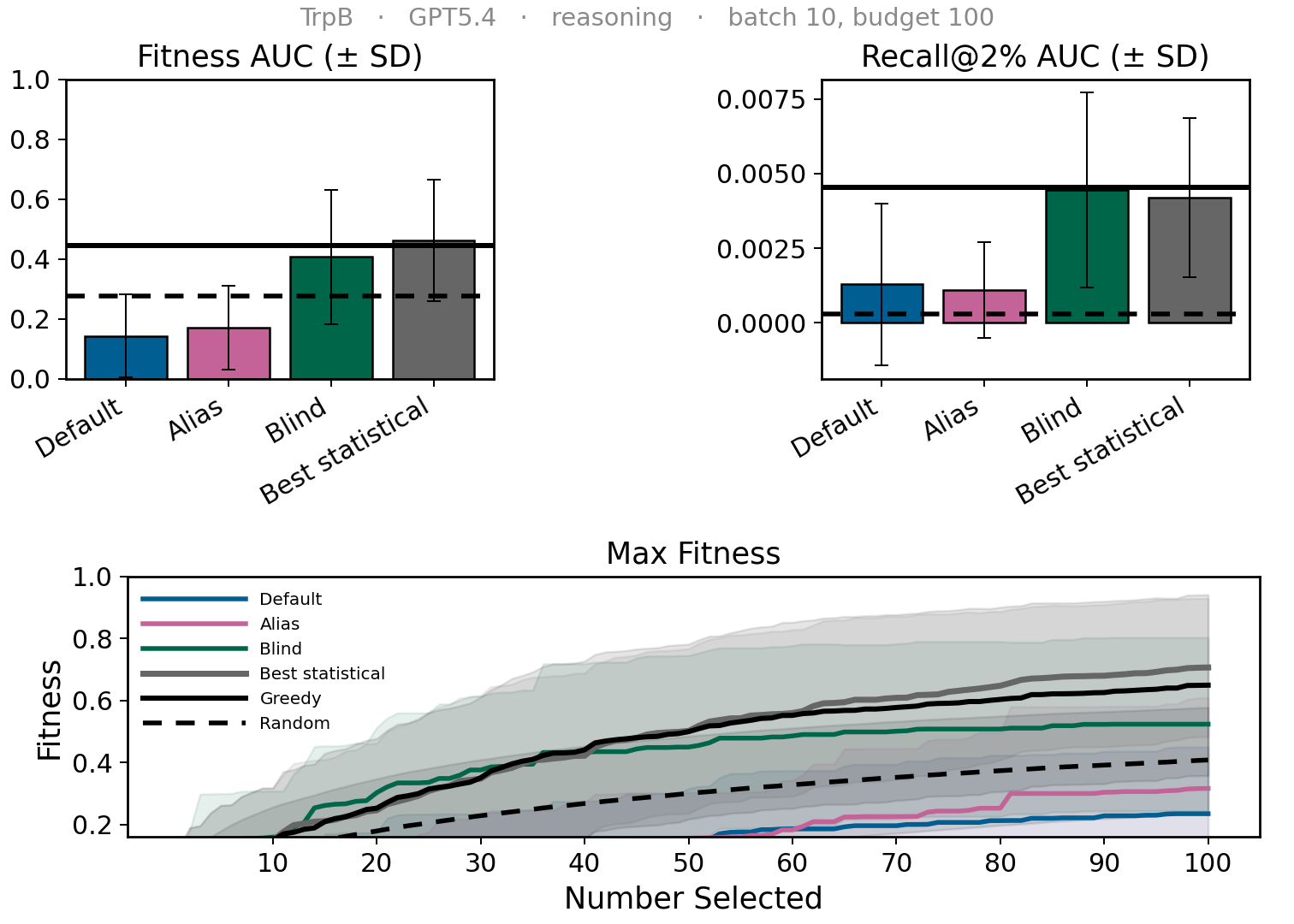}\caption{GPT-5.4}\end{subfigure}\hfill
  \begin{subfigure}[b]{0.45\textwidth}\centering
    \includegraphics[width=\textwidth]{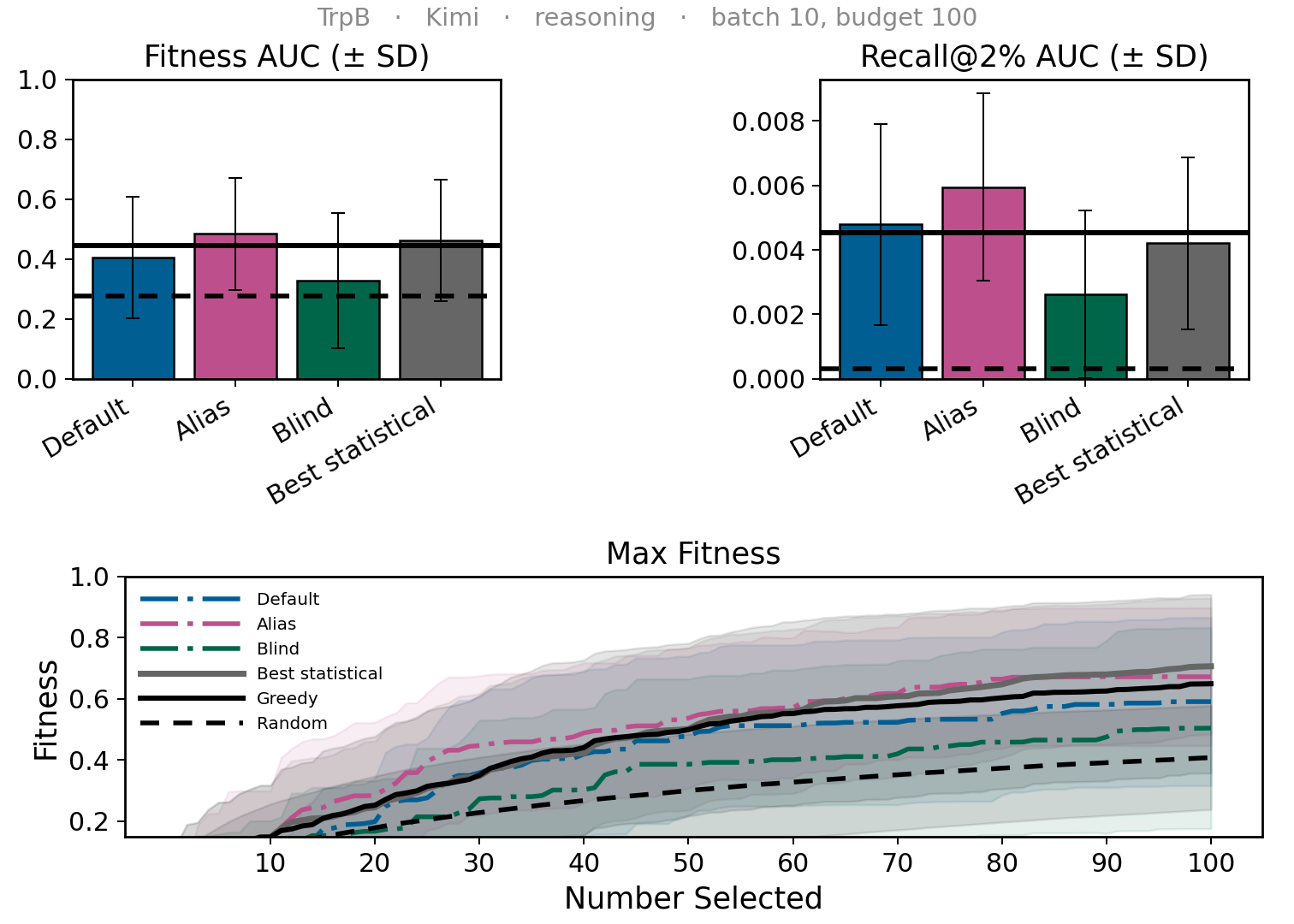}\caption{Kimi-2.5}\end{subfigure}
  \vspace{0.5cm}
  \begin{subfigure}[b]{0.45\textwidth}\centering
    \includegraphics[width=\textwidth]{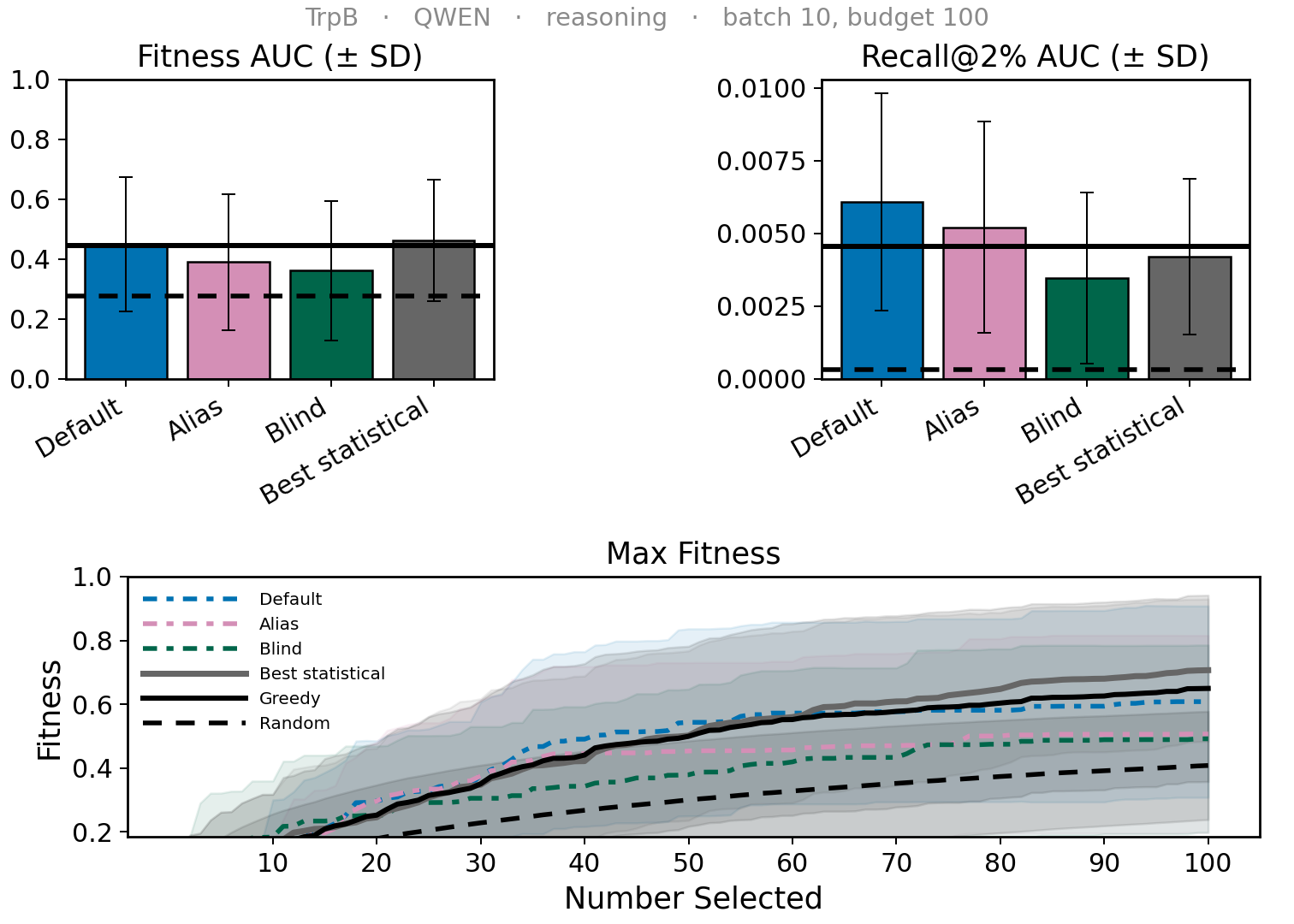}\caption{Qwen-3.5}\end{subfigure}\hfill
  \begin{subfigure}[b]{0.45\textwidth}\centering
    \includegraphics[width=\textwidth]{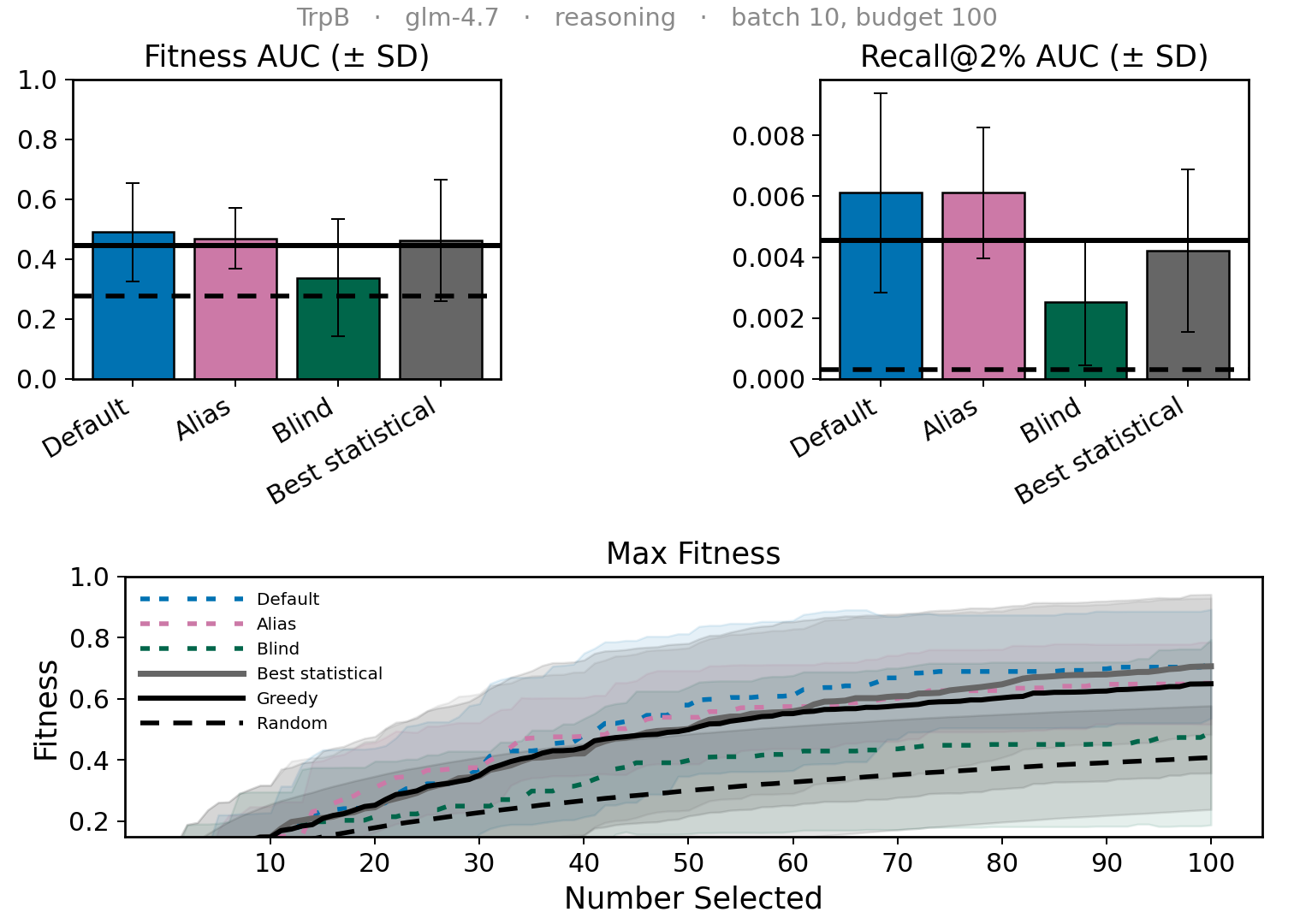}\caption{GLM-4.7}\end{subfigure}
  \caption{Summary statistics on the performance of GPT-5.4, Kimi-2.5, Qwen-3.5, and GLM on the TrpB task. GPT and Qwen show some negative correlation between default prompt mode and performance. 
  Error bars are the standard error. $N=15$ seeds.}
  \label{fig:TrpB-models}
\end{figure}

The investigation raised a series of individually interesting failure modes. Two of the failures below occur only in default prompt mode, not in alias mode, indicating that specifics about the exact dataset (such as target protein names or exact ligand names) are the culprit. Figures~\ref{fig:direct-arylation-models} and \ref{fig:TrpB-models} show the per-model breakdown on those two datasets. A third failure (simple starts on OER) is described separately.

\paragraph{Over-preference for well-cited ligands.}
We observed that for some models, alias or blind mode paradoxically performed better than the default mode on the direct arylation dataset (Figure~\ref{fig:direct-arylation-models}), with the effect most pronounced for GPT-5.4. On the direct arylation dataset, the optimal ligand is ``CgMe-PPh''.  For the general class of reactions (Pd-catalyzed cross-coupling), however, ``X-Phos'' is the most widely cited strong ligand. Searching ``X-Phos palladium '' on Google Scholar yields 2{,}600 hits, compared with only 13 for CgMe-PPh. Similarly, X-Phos appears 435 times in the RedPajama and C4 pretraining corpora, compared with 0 instances of CgMe-PPh. In our setting, GPT-5.4 selects the optimal CgMe-PPh in just $5.6\%$ of default-mode selections compared with $48.5\%$ under blind (15 seeds), while it selects X-Phos in $63.3\%$ of selections in default vs $12.3\%$ in blind. Alias falls in between ($22.3\%$). Qwen-3.5 exposes a similar pattern, sampling CgMe-PPh 12.7\% in default mode vs 33.9\% in blind mode. The remaining three models do not display significant undersampling (Mean difference between blind and default: CI$_{95\%}[-0.09, +0.16]$, 3 models, 15 seeds). Moreover, GPT-5.4 rarely even considers CgMe-PPh. It names X-Phos in $96.7\%$ of cycles but CgMe-PPh in only $22.7\%$. Worse, in $20\%$ of its campaigns it never states the name CgMe-PPh even once. Even when GPT-5.4 or Qwen-3.5 does name CgMe-PPh, they select it far less often than the other models (frequencies $0.41$ and $0.24$ vs $0.64$--$0.95$). GPT-5.4 flags CgMe-PPh as ``unknown, high risk'', but also often ``may be high-value exploration target", yet then declines to use it.

\paragraph{Overcommitment to aromatics in TrpB.}
On ALDE TrpB, GPT-5.4 underperforms in default mode again (Figure~\ref{fig:TrpB-models}). A quick analysis of the reasoning traces showed a strong preference for the amino acid W. However, it is highly underrepresented among the top candidates in the search space (W makes up only $0.2\%$ of residues in the top $1\%$ of candidates; the uniform rate is $5\%$). 

To isolate the probe if the target name (TrpB) triggers memorization, we compare the default prompt against the alias prompt, which keeps the real amino-acid list and the biochemist framing but removes the enzyme's identity (TrpB'' and tryptophan''). We pool all five models (15 seeds each) and apply a one-sided Welch $t$-test to the per-seed fraction of a given amino acid in selected residues. Comparing default and alias, the effect is modest, but multiple models (Kimi, GLM, GPT, Qwen) exhibit an increased preference for aromatics when ``TrpB'' appears in the prompt: on the first cycle, before any experimental feedback, the selected fraction of W rises from $0.048$ to $0.065$ ($p \approx  1\times10^{-4}$). Sonnet is an exception, with a non-positive change. A simmilar effect appears for F ($0.050 \to 0.060$, $p = 0.003$), and but not significantly for Y ($p = 0.17$). The models frequently justify selections using biochemical reasoning (aromatic $\pi$-stacking with the indole substrate). The models partly adapt their priors as the campaign continues. Pooled over all cycles, the W enrichment is no longer significant ($p = 0.38$), yet the models never fully abandon W or F despite accumulating results that argue against aromatic residues.

\begin{figure}[htbp]
  \centering
  \begin{subfigure}[b]{0.45\textwidth}\centering
    \includegraphics[width=\textwidth]{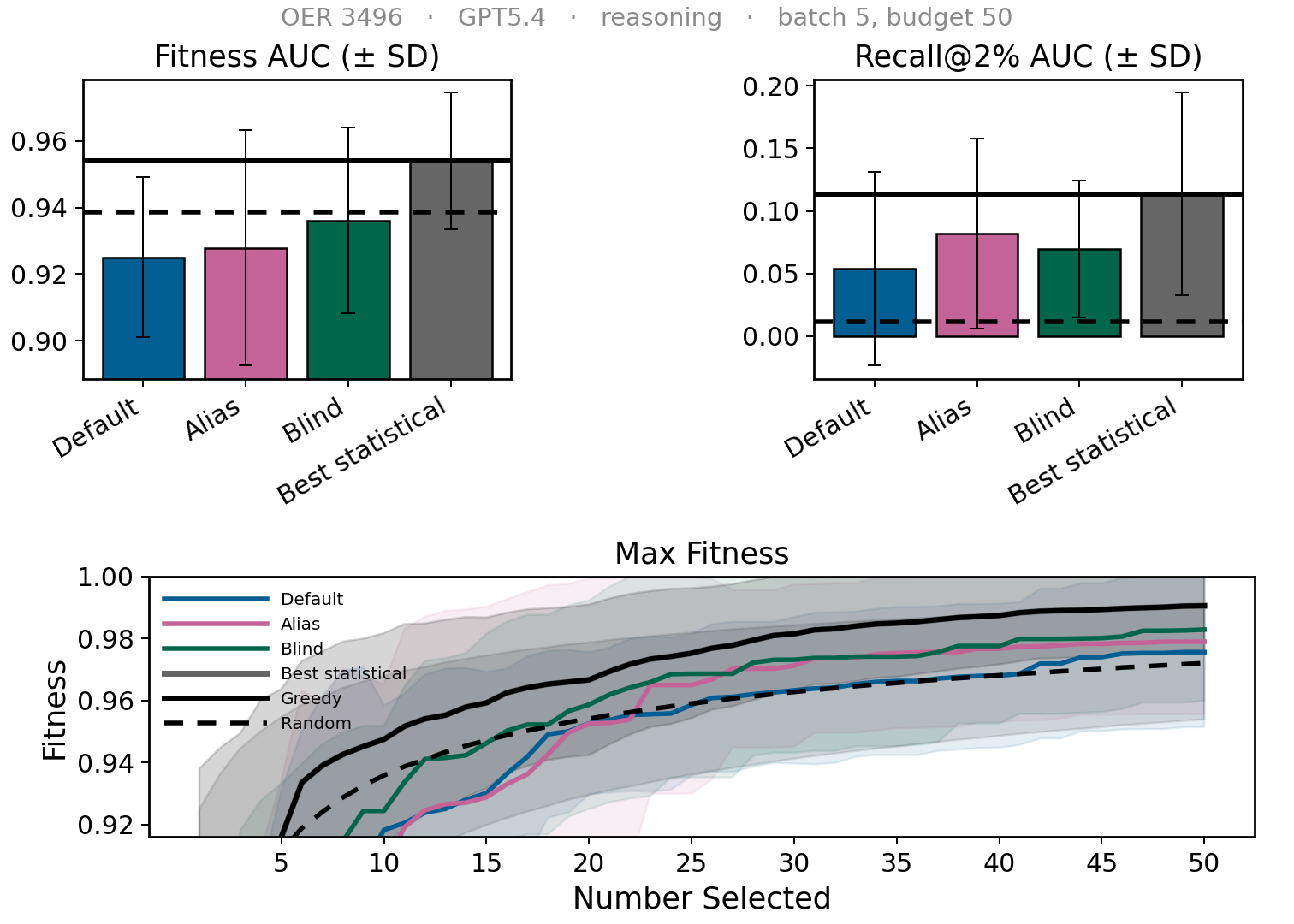}\caption{GPT-5.4}\end{subfigure}\hfill
  \begin{subfigure}[b]{0.45\textwidth}\centering
    \includegraphics[width=\textwidth]{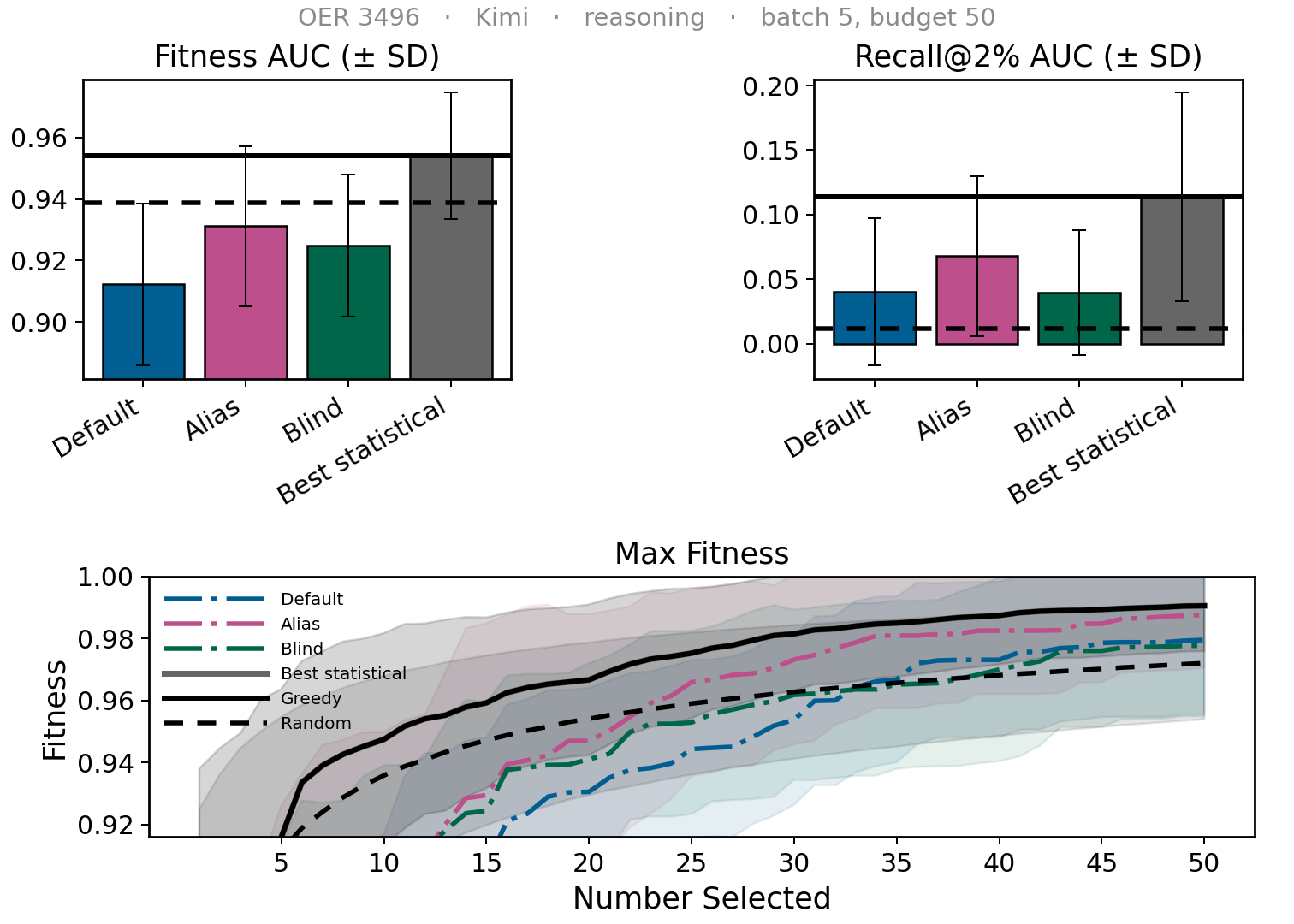}\caption{Kimi-2.5}\end{subfigure}
  \vspace{0.5cm}
  \begin{subfigure}[b]{0.45\textwidth}\centering
    \includegraphics[width=\textwidth]{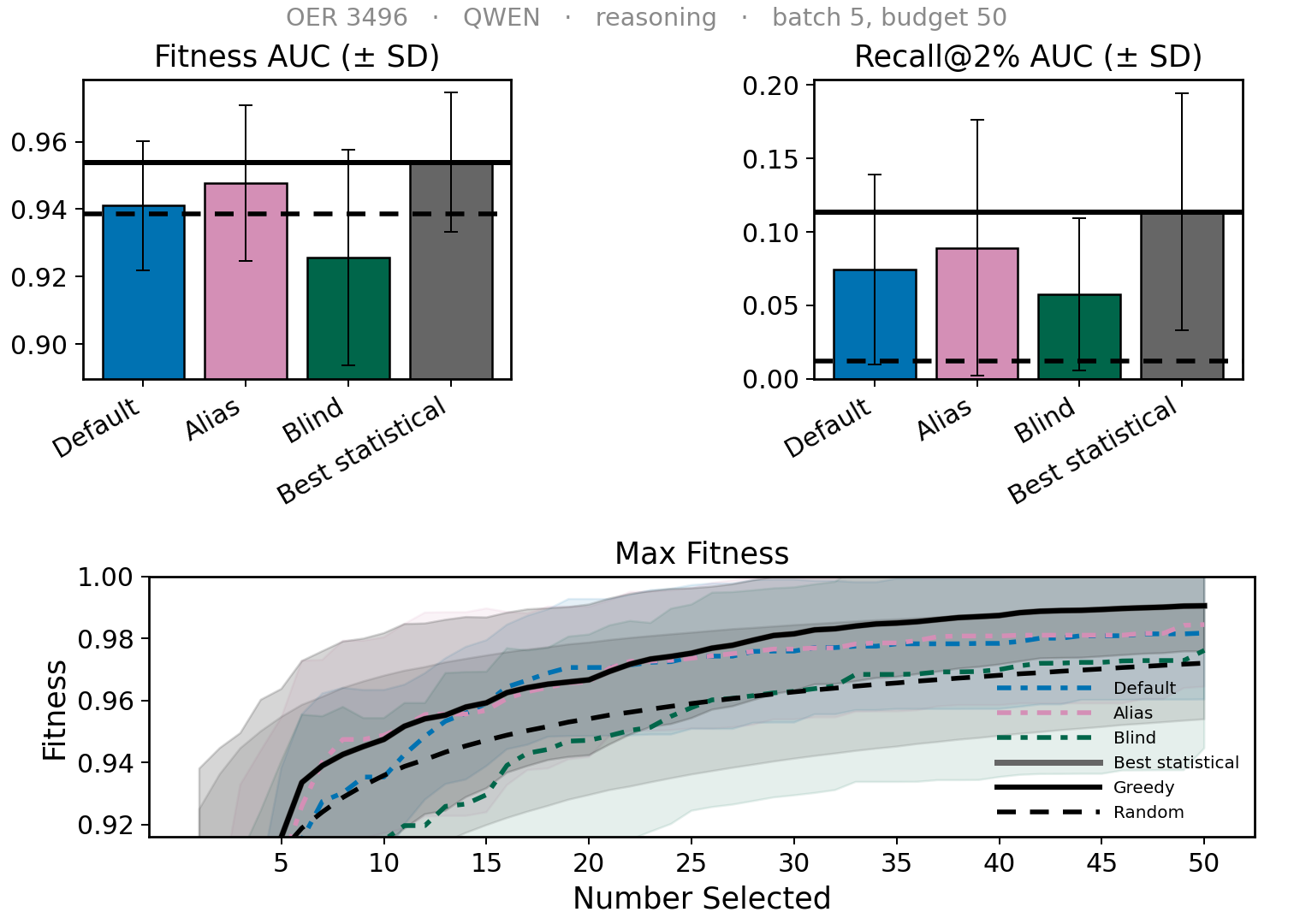}\caption{Qwen-3.5}\end{subfigure}\hfill
  \begin{subfigure}[b]{0.45\textwidth}\centering
    \includegraphics[width=\textwidth]{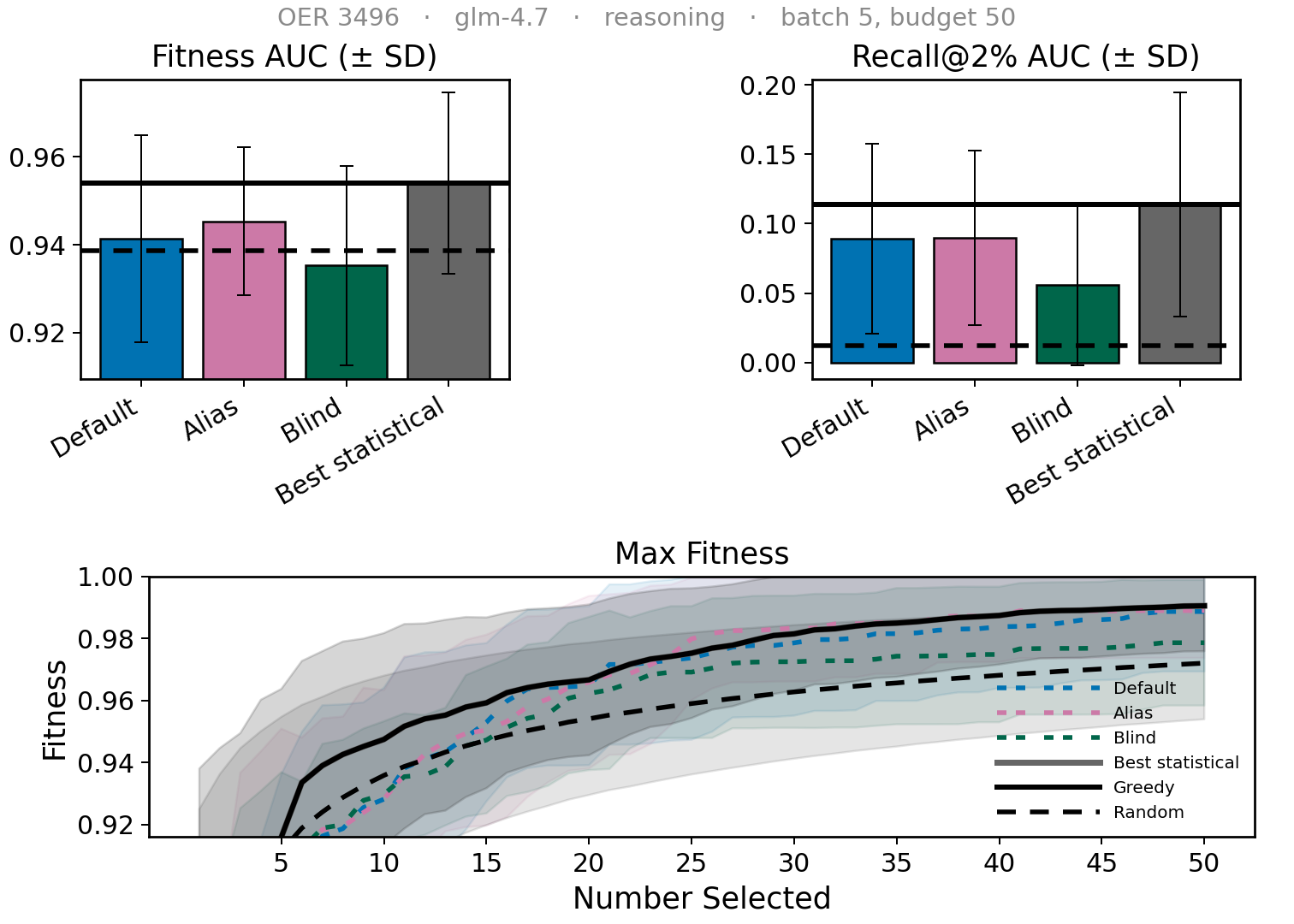}\caption{GLM-4.7}\end{subfigure}
  \caption{Summary statistics on the performance of GPT-5.4, Kimi-2.5, Qwen-3.5, and GLM on the OER task.}
  \label{fig:oer-models}
\end{figure}

\paragraph{Simple starts in OER.}
The failure of many models on the OER dataset stems from a tendency to start the searches with simple choices (selection rate $CI_{95\%}\  [0.18, 0.25]$, 5 models x 3 modes x 3 capabilities, relative to a uniform rate of 0.06) leading to a significant decrease in performance (Figure~\ref{fig:oer-models}). The OER search space comprises six metal ions that should be combined in mixtures at different ratios, with increments of 0.1. All models tend to start simple, with mixtures of only two ions. The strategy would likely be advisable for a human as well. Unfortunately, the OER fitness landscape leans heavily towards mixtures of four different ions. 

\section{Ablations and robustness checks}
\label{si:ablations}

In this section, we report complementary ablation studies that support and validate the claims made in the main text. We find that the behavioral signal observed (history-stickiness) is a structural property rather than an artifact of a particular model, tool configuration, GP surrogate, or prompt. Each subsection is self-contained and points back to the relevant main-text figure.

\subsubsection{Convergence traces}
\label{si:convergence}
 
\begin{figure}[htbp]
  \centering
  \begin{subfigure}[b]{0.45\textwidth}\centering
    \includegraphics[width=\textwidth]{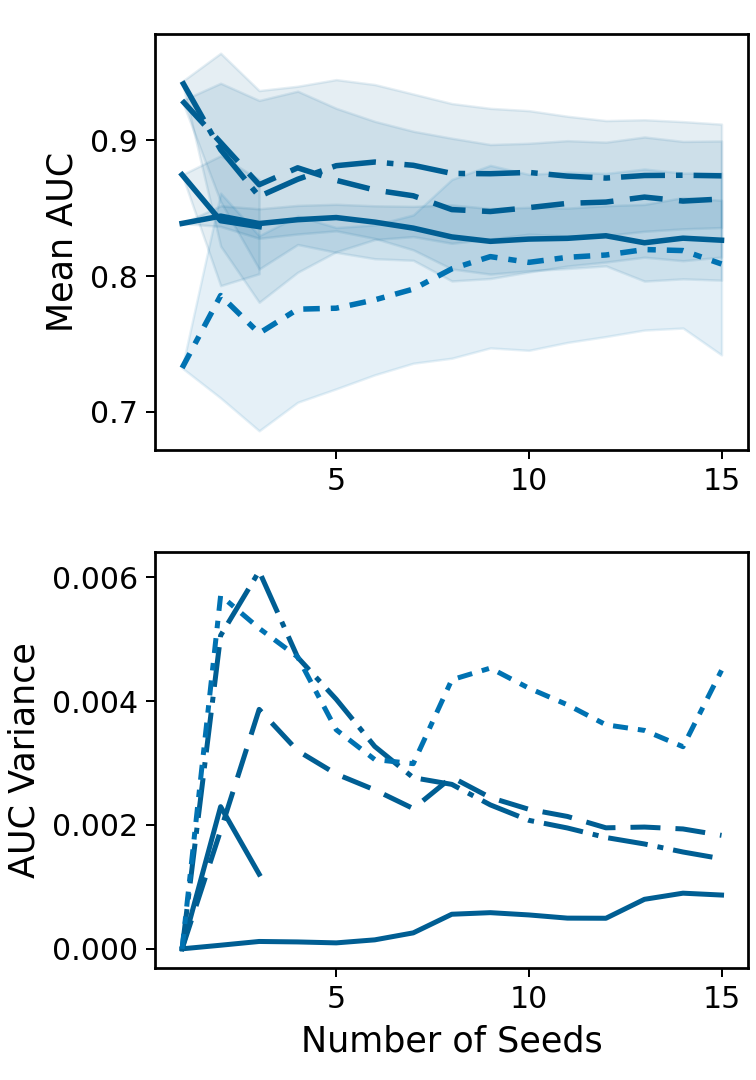}\caption{Default}\end{subfigure}\hfill
  \begin{subfigure}[b]{0.45\textwidth}\centering
    \includegraphics[width=\textwidth]{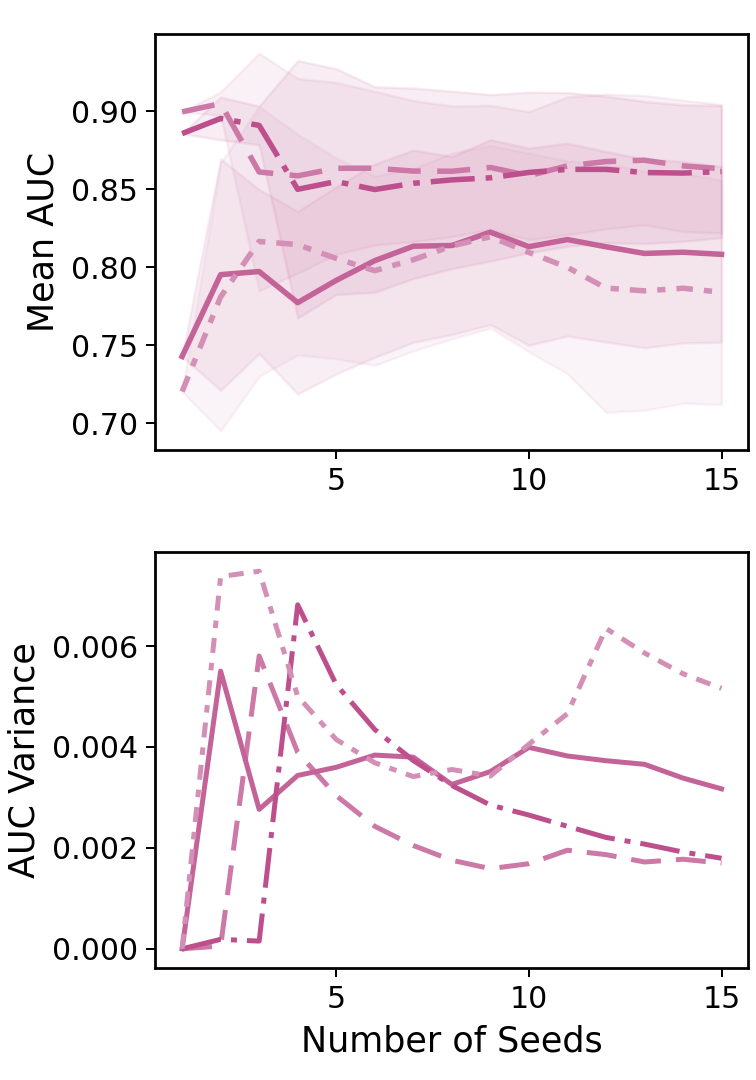}\caption{Alias}\end{subfigure}
  \vspace{0.5cm}
  \begin{subfigure}[b]{0.45\textwidth}\centering
    \includegraphics[width=\textwidth]{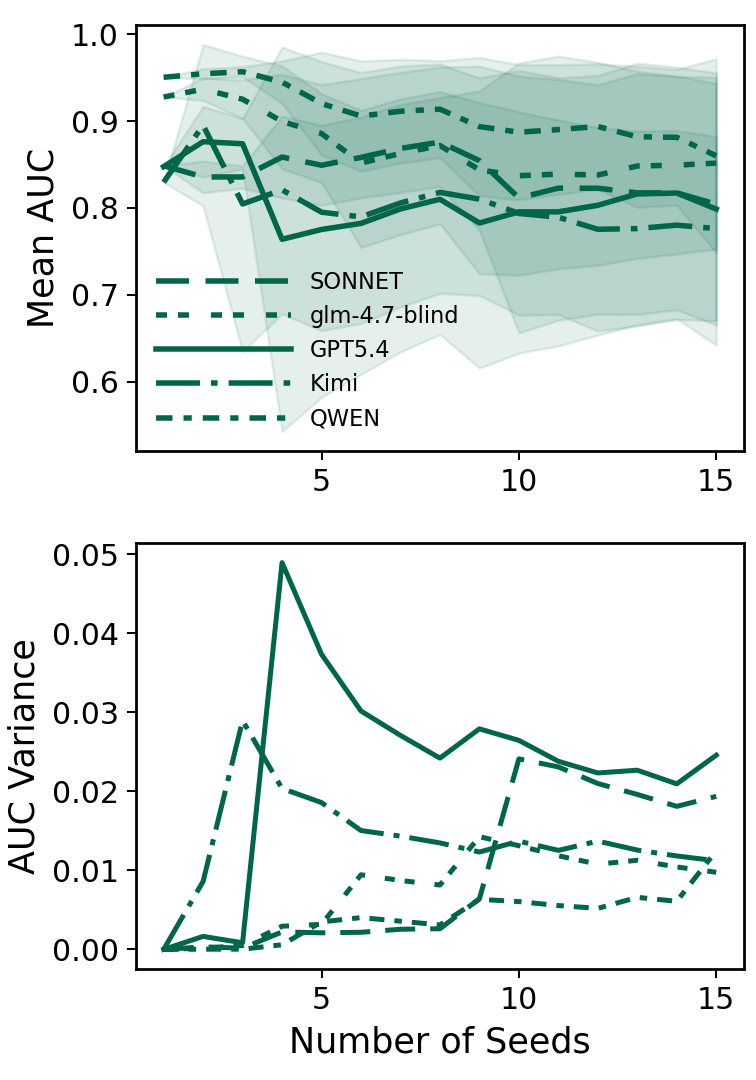}\caption{Blind}\end{subfigure}\hfill
  \begin{subfigure}[b]{0.45\textwidth}\centering
    \includegraphics[width=\textwidth]{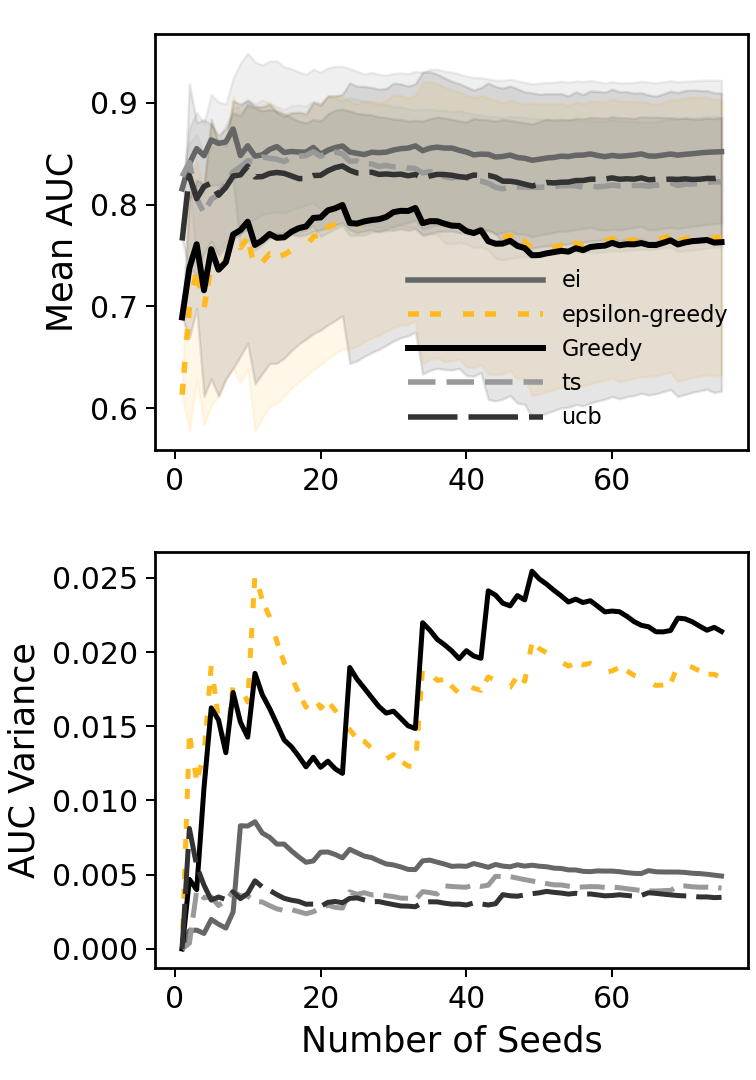}\caption{Statistical}\end{subfigure}
  \caption{Convergence of LLMs and statistical models on the EDBO direct
  arylation dataset (best-yet fitness). LLM panels use 15 seeds; statistical
  models use 75 seeds for tighter estimates, since these normalize the AUC across
  datasets and their variance propagates to all downstream metrics.}
  \label{fig:convergence-direct}
\end{figure}

We inspect the full best-fitness convergence to check that the experimental budget (15 seeds) is enough for an honest aggregate ranking. Figure~\ref{fig:convergence-direct} shows the direct arylation traces split by prompt mode, with the statistical baselines in a separate panel for reference. After $\sim10$ seeds, variance starts to converge and stabilizes by 15 seeds. 
 
\subsection{Context vs.\ tool: a history x filter 2x2}
\label{si:icl-filter}
The main text shows that only removing in-context data (no-history) fully frees the agent from sticking to the validated dataset (main-text Figure~\ref{fig:mechanism-decomp}). In our no-history implementation, we removed two things at once — the in-context validated table and the tool's include/exclude arguments — so we could not uniquely identify the two contributions. Here, we cross them in a $2\times2$ table (Section~\ref{sec:methods-prompts}) and report each cell's performance across several metrics. 

The standard agent (history + filter) and the no-history agent (no-history, no-filter) were run against  7 datasets, 5 models, 2 prompt modes (blind, default), and 15 seeds, matching the main benchmark. The two off-diagonal cells that isolate each contribution (history no-filter and no-history filter) were run on 7 datasets, 2 models (Sonnet-4.6 and GPT-5.4), 2 prompt modes (blind, default), and 10 seeds. We made the 2×2 comparison on the shared GPT-Sonnet, 10-seed subset.

\begin{table}[H]\centering
\caption{history$\times$filter $2\times2$, reported in $(V,U)$ coordinates, normalized AUC, DE-fraction, and the exploration triple (novelty, rarity, coverage gain). Standard agent: 95\% bootstrap CI of the respective metric. Other configurations: $\Delta$ vs.\ Standard (cell $-$ Standard), bootstrap difference on joint dataset and model. Exploration metrics$^{\dagger}$ are excess over the pooled null ($0$ = chance). 10{,}000 resamples; point estimates omitted. Pooled over 7 datasets, 2 models (Sonnet-4.6 and GPT-5.4), and 10 seeds. \textbf{Bold} $\Delta$ CI excludes 0. Arrows mark increased \emph{exploration}.}
\label{tab:icl-filter}
\tiny\setlength{\tabcolsep}{4pt}
\begin{tabular}{llccccccc}
\toprule
Mode & Configuration & $V$ & $U\uparrow$ & norm.\ AUC & DE-frac. $\downarrow$& Novelty$^{\dagger}\uparrow$ & Rarity$^{\dagger}\uparrow$ & Cov.\ gain$^{\dagger}\uparrow$ \\
\midrule
\multirow{4}{*}{Default} & Standard agent & $[+2.29, +2.92]$ & $[-2.48, -1.86]$ & $[0.66, 0.86]$ & $[0.31, 0.54]$ & $[-0.36, -0.23]$ & $[-0.29, -0.20]$ & $[-0.11, -0.06]$ \\
 & No-history, no filter & $\mathbf{[-0.59, -0.15]}$ & $\mathbf{[+0.38, +1.16]}$ & $[-0.06, +0.02]$ & $\mathbf{[-0.16, -0.00]}$ & $\mathbf{[+0.04, +0.06]}$ & $\mathbf{[+0.04, +0.07]}$ & $\mathbf{[+0.01, +0.03]}$ \\
 & history, no filter & $[-0.42, +0.02]$ & $[-0.11, +0.19]$ & $[-0.04, +0.02]$ & $[-0.01, +0.05]$ & $[-0.00, +0.01]$ & $\mathbf{[+0.00, +0.02]}$ & $[-0.00, +0.01]$ \\
 & No-history, full filter & $\mathbf{[-0.76, -0.36]}$ & $\mathbf{[+0.19, +0.79]}$ & $\mathbf{[-0.09, -0.00]}$ & $\mathbf{[-0.18, -0.03]}$ & $\mathbf{[+0.03, +0.06]}$ & $\mathbf{[+0.04, +0.06]}$ & $\mathbf{[+0.00, +0.03]}$ \\
\midrule
\multirow{4}{*}{Blind} & Standard agent & $[+2.23, +2.98]$ & $[-2.09, -1.59]$ & $[0.60, 0.82]$ & $[0.32, 0.58]$ & $[-0.32, -0.20]$ & $[-0.24, -0.16]$ & $[-0.07, -0.03]$ \\
 & No-history, no filter & $\mathbf{[-0.86, -0.31]}$ & $\mathbf{[+0.48, +1.40]}$ & $[-0.02, +0.05]$ & $\mathbf{[-0.21, -0.03]}$ & $\mathbf{[+0.03, +0.06]}$ & $\mathbf{[+0.03, +0.06]}$ & $\mathbf{[+0.00, +0.02]}$ \\
 & history, no filter & $[-0.21, +0.28]$ & $\mathbf{[+0.04, +0.48]}$ & $[-0.03, +0.06]$ & $[-0.11, +0.03]$ & $\mathbf{[+0.00, +0.03]}$ & $\mathbf{[+0.00, +0.03]}$ & $[-0.01, +0.01]$ \\
 & No-history, full filter & $\mathbf{[-0.83, -0.30]}$ & $\mathbf{[+0.43, +1.27]}$ & $[-0.02, +0.05]$ & $\mathbf{[-0.23, -0.05]}$ & $\mathbf{[+0.03, +0.07]}$ & $\mathbf{[+0.03, +0.06]}$ & $[-0.00, +0.02]$ \\
\bottomrule
\end{tabular}
\end{table}

\subsection{Do no-history agents perform relatively better where local search performs relatively worse?}
\begin{figure}
    \centering
    \includegraphics[width=0.8\linewidth]{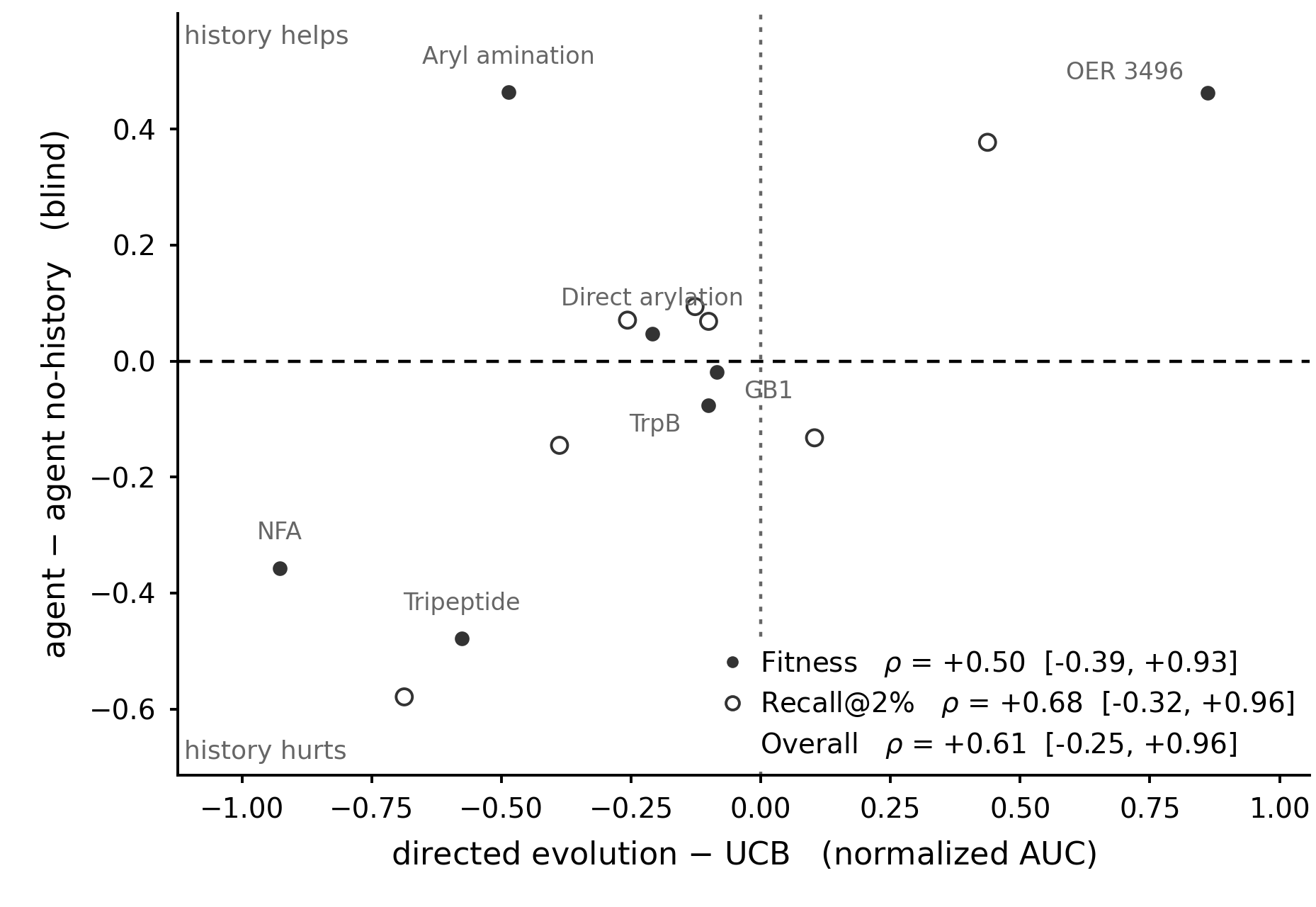}
    \caption{Correlation between relative improvement of including the history in the agent prompt and the relative performance of a local optimizer (DE) over an uncertainty-driven optimizer (UCB). Measured over two metrics: normalized AUC of fitness and recall.}
    \label{fig:si-DE-agent}
\end{figure}

To answer this question, we computed the difference in normalized AUC between the agent and the no-history agent for fitness and recall, respectively. We then correlated this difference with a diagnostic of the relative performance of local optimization (DE) against uncertainty-guided optimization (UCB): the DE-UCB difference. On 7 datasets, the correlation is positive, suggesting that history harms where uncertainty helps, but the 95\% CI is wide and includes 0. More datasets are needed (Figure \ref{fig:si-DE-agent}).

\subsection{Surrogate sensitivity of the (V,U) decomposition}
\label{si:surrogate}

The $(V,U)$ channels are read off a single fixed GP surrogate (RBF kernel, lengthscale $\ell=2$, white noise $0.1$) shared across datasets. Table~\ref{tab:surrogate} sweeps the GP hyperparameters and reports the rank correlation of the fitted $U$- and $V$-orderings against this reference. The $U$-channel ordering is essentially invariant ($\rho=1.00$ everywhere except at the shortest lengthscale $\ell=0.5$, where $\rho=0.96$). The $V$-channel is robust for $\ell\ge 2$ and noise $\le 0.5$ and softens only at short length scales.

\begin{table}[H]\centering
\caption{ $(V,U)$ decomposition sensitivity analysis to the hyperparameters of the GP surrogate. Spearman-$\rho$ of the policy ordering on each channel against the fixed GP ($\ell=2$, noise $=0.1$, $c=1$; RBF), with a paired-bootstrap 95\% CI (2{,}000 resamples)}
\label{tab:surrogate}
\small\setlength{\tabcolsep}{4pt}
\begin{tabular}{lll}
\toprule
GP hyperparameters & $\rho(U)$ & $\rho(V)$ \\
\midrule
\textbf{Reference} ($\ell=2$, noise $=0.1$, $c=1$) & $1.00$ & $1.00$\\
\midrule
\multicolumn{3}{l}{\emph{Lengthscale $\ell$}} \\
\quad $\ell=0.5$ & $0.96$ {\scriptsize $[0.64, 1.00]$} & $0.79$ {\scriptsize $[0.61, 0.96]$}  \\
\quad $\ell=1$ & $1.00$ {\scriptsize $[0.89, 1.00]$} & $0.86$ {\scriptsize $[0.61, 0.96]$}  \\
\quad $\ell=4$ & $1.00$ {\scriptsize $[0.89, 1.00]$} & $1.00$ {\scriptsize $[0.86, 1.00]$}  \\
\quad $\ell=8$ & $1.00$ {\scriptsize $[0.89, 1.00]$} & $1.00$ {\scriptsize $[0.86, 1.00]$} \\
\midrule
\multicolumn{3}{l}{\emph{Noise}} \\
\quad noise $=0.01$ & $1.00$ {\scriptsize $[0.93, 1.00]$} & $0.93$ {\scriptsize $[0.82, 1.00]$}  \\
\quad noise $=0.5$ & $1.00$ {\scriptsize $[0.96, 1.00]$} & $1.00$ {\scriptsize $[0.86, 1.00]$}  \\
\quad noise $=1.0$ & $1.00$ {\scriptsize $[0.89, 1.00]$} & $1.00$ {\scriptsize $[0.86, 1.00]$}  \\
\midrule
\multicolumn{3}{l}{\emph{Constant $c$}} \\
\quad $c=0.25$ & $1.00$ {\scriptsize $[0.96, 1.00]$} & $1.00$ {\scriptsize $[0.86, 1.00]$}  \\
\quad $c=4$ & $1.00$ {\scriptsize $[0.96, 1.00]$} & $0.93$ {\scriptsize $[0.86, 1.00]$}  \\
\bottomrule
\end{tabular}
\end{table}

\subsection{A capability-nudge prompt does not remove the stickiness}
\label{si:nudge}

To test whether the stickiness is prompt-dependent rather than structural, we added a prompt variant that explicitly instructs the model to look for under-explored, high-uncertainty regions of the search space. If the bias is prompt-dependent, the nudge should shift the model toward higher $U$ and higher novelty. We ran all 5 reasoning models for 10 seeds with the added nudge across all 7 datasets, in both blind and default mode (Details in Sec~\ref{sec:methods}). Table~\ref{tab:nudge} shows the effect of adding an exploration nudge, pooled across models and datasets for each of the exploration and performance metrics used in this report. We find that nudging the model to consider the entire search space incrementally alleviates anti-correlation with uncertainty ($p_{BH}<0.05$, bootstrap, 7 datasets, 5 models, 10 seeds), but does not significantly change any of the exploration metrics. The bias is not addressed by a direct prompt-level instruction to explore, consistent with a structural rather than instructional origin. However, we test only a single intervention, which is not sufficient for a stronger claim.

\begin{table}[H]\centering
\caption{Capability-nudge ablation (all five reasoning LLMs), on the same battery as Table~\ref{tab:icl-filter}: $(V,U)$ coordinates, normalized AUC, DE-fraction, and the exploration triple (novelty, rarity, coverage gain). Nudged: $\Delta$ vs.\ Baseline (nudged $-$ baseline), bootstrap difference on shared dataset and model. Exploration metrics$^{\dagger}$ are excess over the pool null ($<0$ implies less exploration than random). All CIs are based on 10{,}000 resamples across 7 datasets, 5 models, and 10 seeds each. \textbf{Bold} $\Delta$ CI separated from 0.}
\label{tab:nudge}
\small\setlength{\tabcolsep}{4pt}
\tiny{\begin{tabular}{llccccccc}
\toprule
Mode & Prompt & $V$ & $U$ & norm.\ AUC & DE-frac. & Novelty$^{\dagger}$ & Rarity$^{\dagger}$ & Cov.\ gain$^{\dagger}$ \\
\midrule
\multirow{2}{*}{Default} & Baseline & $[+1.94, +2.14]$ & $[-3.18, -2.74]$ & $[0.66, 0.82]$ & $[0.43, 0.58]$ & $[-0.34, -0.26]$ & $[-0.28, -0.23]$ & $[-0.11, -0.07]$ \\
 & Nudged & $[-0.06, +0.21]$ & $[-0.05, +0.31]$ & $[-0.01, +0.05]$ & $[-0.02, +0.02]$ & $[-0.00, +0.01]$ & $[-0.00, +0.01]$ & $\mathbf{[+0.00, +0.01]}$ \\
\midrule
\multirow{2}{*}{Blind} & Baseline & $[+2.11, +2.41]$ & $[-2.94, -2.60]$ & $[0.63, 0.77]$ & $[0.44, 0.60]$ & $[-0.32, -0.24]$ & $[-0.26, -0.21]$ & $[-0.08, -0.04]$ \\
 & Nudged & $[-0.11, +0.26]$ & $\mathbf{[+0.04, +0.44]}$ & $[-0.01, +0.04]$ & $[-0.01, +0.04]$ & $[-0.00, +0.01]$ & $[-0.00, +0.01]$ & $[-0.00, +0.01]$ \\
\bottomrule
\end{tabular}}
\end{table}



\newcommand{\promptfragdir}{sections/all-prompts}

\newcommand{\promptinput}[1]{\lstinputlisting{\promptfragdir/#1}}

\section{LLM prompt templates}

\label{sec:prompt-appendix}

\subsection{Shared schema}
\label{prompt:schema}
\promptcaption{Sent on every API and agent turn; agent campaigns append a line requiring a tool call or \textless{}batch\textgreater{} block.}
\promptinput{system.txt}

\promptcaption{User prompt schema}
\promptinput{schema.txt}

\subsection{Default mode user prompts}

\promptcaption{Cycle-1 user message: domain background, search space, and output format with canonical labels.}

\subsubsection{EDBO direct-arylation}
\label{prompt:default:edbo:direct-arylation}
\promptcaption{Default prompt mode.}
\promptinput{edbo_default_direct_arylation.txt}

\subsubsection{EDBO aryl-amination-2b}
\label{prompt:default:edbo:aryl-amination-2b}
\promptcaption{Default prompt mode.}
\promptinput{edbo_default_aryl_amination_2b.txt}

\subsubsection{ALDE GB1}
\label{prompt:default:alde:GB1}
\promptcaption{Default prompt mode.}
\promptinput{alde_default_GB1.txt}

\subsubsection{ALDE TrpB}
\label{prompt:default:alde:TrpB}
\promptcaption{Default prompt mode.}
\promptinput{alde_default_TrpB.txt}

\subsubsection{GRYFFIN perovskites}
\label{prompt:default:gryffin:perovskites}
\promptcaption{Default prompt mode.}
\promptinput{gryffin_default_perovskites.txt}

\subsection{Alias mode user prompts}

\promptcaption{Codenames for reagent values; component names unchanged. Seed fixes the alias draw.}

\subsubsection{EDBO direct-arylation}
\label{prompt:alias:edbo:direct-arylation}
\promptcaption{Alias prompt mode.}
\promptinput{edbo_alias_direct_arylation.txt}

\subsubsection{EDBO aryl-amination-2b}
\label{prompt:alias:edbo:aryl-amination-2b}
\promptcaption{Alias prompt mode.}
\promptinput{edbo_alias_aryl_amination_2b.txt}

\subsubsection{ALDE GB1}
\label{prompt:alias:alde:GB1}
\promptcaption{Alias prompt mode.}
\promptinput{alde_alias_GB1.txt}

\subsubsection{ALDE TrpB}
\label{prompt:alias:alde:TrpB}
\promptcaption{Alias prompt mode.}
\promptinput{alde_alias_TrpB.txt}

\subsubsection{GRYFFIN perovskites}
\label{prompt:alias:gryffin:perovskites}
\promptcaption{Alias prompt mode.}
\promptinput{gryffin_alias_perovskites.txt}

\subsection{Blind mode user prompt (EDBO direct-arylation)}
\label{prompt:blind}
\promptcaption{Generic background, codenames for components and values, blind validated-section title.}
\promptinput{edbo_blind_direct_arylation.txt}

\subsection{Agent tool (filter-candidates)}
\label{prompt:tool}
\promptcaption{Tool schema bound on the first agent turn; identical across datasets.}
\promptinput{tool.txt}

\subsection{Parse-retry feedback (EDBO direct-arylation)}
\label{prompt:feedback}
\promptcaption{Appended after a failed parse; example valid/invalid tuples from the dataset oracle.}
\promptinput{edbo_feedback_direct_arylation.txt}

\subsection{Non-reasoning user prompt (EDBO direct-arylation)}
\label{prompt:none}
\promptcaption{--reasoning none: no strategic-approach block; stricter output requirements and suffix.}
\promptinput{edbo_none_direct_arylation.txt}

\subsection{Fresh agent response}
\label{prompt:agent-response}
\promptcaption{Thinking example from qwen agent on the first step of a campaign}
\promptinput{agent-thinking-response-qwen.txt}

\subsection{Mid campaign agent response}
\label{prompt:mid-campaign-response}
\promptcaption{Thinking example from qwen agent at a later step of a campaign. Note that tool payloads and calls occured before this thinking session.}
\promptinput{agent-thinking-response-qwen2.txt}

\subsection{Extraction prompts}
\label{prompt:extraction}
\promptcaption{Prompt used to extract hypotheses from reasoning}
\promptinput{hypothesis_extraction.txt}

\promptcaption{Prompt used to extract beliefs from reasoning}
\promptinput{belief_extraction.txt}

\end{document}